\documentclass[11pt]{article}

\usepackage[final]{acl}

\usepackage[markup=underlined]{changes}

\usepackage[utf8]{inputenc}   
\usepackage[T1]{fontenc}    	
\usepackage{url}              		 
\usepackage{booktabs}       	
\usepackage{amsfonts}       	
\usepackage{nicefrac}       	  
\usepackage{microtype}      	

\usepackage{times}
\usepackage{latexsym}
\usepackage{inconsolata}

\usepackage{graphicx}
\usepackage{color}
\usepackage{rotating}
\usepackage{tabularx}
\usepackage{pdflscape}
\usepackage{amsmath,amsthm,amssymb}
\usepackage{algorithm,algorithmic}
\usepackage{bbm,dsfont}
\usepackage{subfig}
\usepackage{caption}
\usepackage{wrapfig}
\usepackage{cancel}
\usepackage{enumerate,cases}
\usepackage{thmtools,thm-restate}
\usepackage{mathtools}

\usepackage{multirow}
\usepackage{makecell}
\usepackage{setspace}
\usepackage{pifont}
\usepackage{array}
\usepackage{enumitem}
\usepackage{lineno}     
\usepackage{bm}
\usepackage{diagbox}

\allowdisplaybreaks

\usepackage{silence}
\usepackage{hyperref}		 
\hypersetup{
	colorlinks	= true,		 
	urlcolor     = blue,	 
	linkcolor	 = purple, 	 
	citecolor    = violet    
}

\usepackage[capitalize]{cleveref}
\crefformat{figure}{Fig.~#2#1#3}       
\crefformat{table}{Tab.~#2#1#3}        
\crefformat{equation}{Eq.~(#2#1#3)}    
\crefformat{section}{Sec.~#2#1#3}      
\crefformat{appendix}{App.~#2#1#3}     
\crefformat{algorithm}{Alg.~#2#1#3}    

\usepackage{silence}
\renewcommand{\phi}{\varphi}
\renewcommand{\epsilon}{\varepsilon}

\newcommand{\lp}{\left(}
\newcommand{\rp}{\right)}

\newcommand{\step}[1]{{\large \textcircled{\small #1}}}

\newcommand{\sH}[1]{\text{H}\lp #1 \rp}

\renewcommand{\phi}{\varphi}
\renewcommand{\epsilon}{\varepsilon}

\newcommand{\para}[1]{\paragraph{#1\hspace{-1mm}}}

\newcommand{\squishlisttwo}{
 \begin{list}{$\bullet$}
  { \setlength{\itemsep}{5pt}
     \setlength{\parsep}{0pt}
    \setlength{\topsep}{0pt}
    \setlength{\partopsep}{0pt}
    \setlength{\leftmargin}{1em}
    \setlength{\labelwidth}{1.5em}
    \setlength{\labelsep}{0.5em} } }

\newcommand{\squishend}{
  \end{list}  }

\newcommand{\cA}{\mathcal{A}}

\newcommand{\cC}{\mathcal{C}}

\newcommand{\cE}{\mathcal{E}}

\newcommand{\cS}{\mathcal{S}}

\newcommand{\cV}{\mathcal{V}}

\newcommand{\cX}{\mathcal{X}}
\newcommand{\cY}{\mathcal{Y}}

\theoremstyle{plain}

 \newtheorem{remark}{Remark}
 
 \newtheorem{assu}{Assumption}

\title{
    Prompting the Unknown: \\
    Understanding Response Uncertainty in Large Language Models
}

\author{
    \textbf{Ze Yu Zhang\textsuperscript{$^\star$1}},
    \textbf{Arun Verma\textsuperscript{$^\star$2}},
    \textbf{Finale Doshi-Velez\textsuperscript{3}},
    \textbf{Bryan Kian Hsiang Low\textsuperscript{1,2}}
    \\
    {\small
        \textsuperscript{1}Department of Computer Science, National University of Singapore, Republic of Singapore
    }
    ~\\
    {\small
        \textsuperscript{2}Singapore-MIT Alliance for Research and Technology Centre \qquad
        \textsuperscript{3}Harvard University
    }
    ~\\
    {\small
        \texttt{zhan1130@comp.nus.edu.sg} \qquad \texttt{arun.verma@smart.mit.edu}
    }
    ~\\{\small
        \texttt{finale@seas.harvard.edu}\qquad \texttt{lowkh@comp.nus.edu.sg}
    }
}

\begin{document}
    \makeatletter
    \def\blankfootnote{\xdef\@thefnmark{}\@footnotetext}
    \makeatother
    
    \maketitle

    \begin{abstract}    
        Large language models (LLMs) are widely used in decision-making across diverse domains. Ensuring the generation of safe and reliable responses is critical for the effective deployment of LLM-based applications, particularly in high-stakes domains such as healthcare and finance. Most of these applications typically use carefully crafted prompts to guide response generation; however, the relationship between prompts and the reliability of LLM-generated responses is not yet fully understood. To address this gap, we propose a novel prompt-response concept model that explains the relationship between the amount of task-relevant information (informativeness) provided in the prompt and the LLM-generated response uncertainty by identifying four sources of response uncertainty: prompt underspecification, model quality, task variability, and semantic redundancy. We prove that response uncertainty decreases as prompt informativeness or model quality increases, mirroring the behavior of epistemic uncertainty in probabilistic models. Our experimental results on real-world datasets further validate our proposed model and corroborate the theoretical results.
        \blankfootnote{\textbf{$^\star$}Equal contribution and corresponding authors.}
    \end{abstract}



    \section{Introduction}
    \label{sec:introduction}

Large language models (LLMs) have demonstrated impressive performance across a variety of tasks~\citep{ArXiv23_google2023palm, ArXiv23_openai2023gpt, ArXiv23_zhao2023survey}. 
This success has led to their widespread adoption and significant involvement in decision-making applications across various domains, such as healthcare~\citep{karabacak2023embracing, sallam2023chatgpt,yang2023large}, finance~\citep{wu2023bloomberggpt}, education~\citep{ xiao2023evaluating}, and law~\citep{zhang2023unleashing}.
Most of these applications typically use carefully crafted prompts to guide the desired response generation; however, the relationship between prompts and the reliability of LLM-generated outputs, particularly for high-stakes tasks in domains such as healthcare and finance, is not yet well understood~\citep{arkoudas2023gpt, huang2023large}.
Therefore, {\bf\em understanding how the information in prompts influences response uncertainty is critical to generate safe and reliable responses} for the effective deployment of LLM-based decision-making applications.

To understand this importance, consider a mobile health application (mHealth app) in which machine learning algorithms are incorporated to monitor users' health conditions and provide personalized suggestions on daily activities to influence their decision-making~\citep{boursalie2018machine, trella2022designing, trella2023reward}. 
For LLMs to be suitable for such tasks, they must generate safe and consistent responses, e.g., consider an LLM-powered mHealth app that recommends physical therapy (PT) routines to a patient recovering from surgery. 
To promote patient adherence to the PT regimen despite discomfort, the mHealth app must deliver accurate and consistent suggestions. Inconsistent advice can undermine the patient's progress and reduce trust in the application, so \textit{reliable responses are essential for the mHealth app's effectiveness}~\citep{shin2022modeling}.

With the emergent capabilities of LLMs~\citep{dong2022survey,kojima2022large,wei2022chain,yao2023react}, it is possible to improve model responses by guiding them with informative prompts that include task-relevant instructions and exemplars. 
It helps LLMs to effectively use the relevant information acquired during pretraining to generate better responses, even if the prompt itself does not explicitly reveal the ground truth~\citep{liu2023pre,sahoo2024systematic}. 
The LLM responses are generated by sequentially sampling tokens from probability distributions over vocabulary, conditioned on the given prompt.
LLMs use different decoding strategies such as beam search~\citep{cho2014properties}, greedy~\citep{sutskever2014sequence}, and nucleus sampling~\citep{holtzman2019curious}.

Typically, tokens with higher probabilities are chosen sequentially to generate the response. The response variations are controlled by LLM hyperparameters like temperature~\citep{hinton2015distilling}, top-$k$~\citep{fan2018hierarchical}, or top-$p$~\citep{holtzman2019curious}. 
While variability benefits creative tasks like poem and essay writing, it can be detrimental for tasks requiring high reproducibility and consistency~\citep{ganguli2022predictability, huang2023look}.
Making LLMs generate deterministic responses is also not ideal, as users' preferences may vary~\citep{wu2023survey}.
Thus, a better approach is needed to understand and separate the different sources of response uncertainty and develop methods to reduce uncertainty naturally rather than masking it by adjusting LLM hyperparameters.
As response uncertainty of a fixed or black-box LLM can also be controlled by the amount of task-relevant information in the prompt (i.e., {\bf\em prompt informativeness}), this paper focuses on {\bf\em quantifying the response uncertainty due to the prompt} while keeping the LLM and its hyperparameters fixed.

To this end, we use the insight that LLMs implicitly learn to infer latent concepts during pretraining~\citep{xie2021explanation, hahn2023theory, zhang2023and} and propose a {\bf\em novel prompt-response concept} (PRC) model. 
Our PRC model conceptualizes how an LLM generates responses based on given prompts and {\bf\em helps understand the relationship between prompts and response uncertainty} by measuring response uncertainty for prompts with varying task-relevant information. 
Specifically, it provides a principled framework for separating and analyzing the distinct sources of response uncertainty.
Under an idealized LLM assumption (that the LLM has perfectly learned), we provide theoretical results that show the {\bf\em uncertainty of responses generated by an LLM decreases as the prompt informativeness and model quality increase}. 
We draw a connection between the reducible response uncertainty and epistemic uncertainty, and using our PRC model, we {\bf\em theoretically justify why increasing the task-relevant information in the prompt} is a principled and effective method to reduce the response uncertainty.
Finally, we validate the PRC model through experiments and demonstrate the practical applicability of our insights using a healthcare use case.
Specifically, our key contributions are as follows: 
\vspace{-1.45mm}
\begin{itemize}
    \setlength\itemsep{-0.07em}
    \item \textbf{Prompt-Response Concept model.} 
    In \cref{sec:latent_concept}, we propose a prompt-response concept model to understand and quantify the relationship between prompt informativeness and response uncertainty in LLMs. 
	
	\item \textbf{Decomposition of response uncertainty.} 
    Using the PRC model, we decompose the response uncertainty into four distinct sources: prompt underspecification, model quality, task variability, and semantic redundancy, where uncertainty due to prompt underspecification is an epistemic uncertainty, while the others are the sources of aleatoric uncertainty when the LLM and its hyperparameters are fixed.
    
    \item \textbf{Theoretical result.} 
    In \cref{sec:theoretical_results}, we prove that response uncertainty decreases as prompt informativeness or model quality increases (\cref{thm:response_uncertainty}). 
    We further link reducible uncertainty to epistemic uncertainty, as adding more relevant information to prompts and improving model quality reduces response uncertainty.

    \item \textbf{Empirical results.}
    In \cref{sec:experiments}, we validate the PRC model through experiments and demonstrate its practical applicability across diverse tasks based on real-world datasets and a healthcare use case. In addition, we observe that, when supplied with sufficient relevant information, a smaller (less capable) model can sometimes outperform a larger (more capable) model with less information.
\end{itemize}

    \section{Preliminaries}
    \label{sec:preliminaries}

\para{Problem setting.}
Let $\cX$ denote the set of all prompts and $\cY$ the set of all responses generated by an LLM $f$, where $f: \cX \rightarrow \cY$. Let $\cV$ represent the vocabulary of all unique tokens. For any prompt $x \in \cX$ and response $y \in \cY$, we have $x \in \cV^{|x|}$ and $y \in \cV^{|y|}$, where $|\cdot|$ denotes the number of tokens in the prompt or response.
For a given prompt $x \in \cX$, the LLM $f$ generates a response $y \in f(x)$, which can vary each time the LLM generates it due to two factors: (i) the LLM hyperparameters, such as temperature, top-$k$, and top-$p$, which control the randomness in token sampling, and (ii) the task-relevant information in the prompt, which we refer to as \emph{prompt informativeness}.
This paper focuses solely on the latter aspect while keeping the LLM and its hyperparameters fixed. 
Specifically, this paper aims to explain how prompt informativeness influences response uncertainty for a given LLM and then quantify response uncertainty as a function of both prompt informativeness and the LLM.

\para{Measure for response uncertainty.}
We use entropy to measure the uncertainty in the responses generated by an LLM for a given prompt. 
Let $Y$ be a random variable representing the response. 
For a given prompt $x$, the entropy of $Y$ is defined as:
\vspace{-1mm}
\begin{equation}
    \label{eqn:entropy}
     \sH{Y|x} = -\sum_{y \in \cY} p(y|x) \log_2 p(y|x),\vspace{-2mm}
\end{equation}
where $p(y|x)$ is the conditional probability that the response variable $Y$ takes the value $y$ given the prompt $x$.
Intuitively, a more informative prompt reduces the uncertainty of response generated by the LLM, leading to lower entropy in $Y$.

\para{Concept.}
The notion of a concept varies across fields: in philosophy, a concept represents the fundamental unit of thought; in psychology, it is a mental construct; in linguistics, it refers to the semantic units that words or phrases represent; and in education, it denotes key ideas or principles.
In this paper, we define a {\em concept} as an abstraction derived from specific instances or occurrences that share common characteristics~\citep{fodor1998concepts,laurence1999concepts,weiskopf2009plurality,wilmont2013cognitive}. 
To illustrate this notion of concept, consider the example of \emph{Meeting}, which includes information such as the agenda, date and time, location, and other relevant details about the meeting, for example, ``\emph{Write an email to schedule a meeting on 28 July 2025, from 14:00 to 15:30, in Room 301 of the Department Building. The agenda includes reviewing the draft paper, discussing key findings, and planning the next steps for submission. Participants include the lead author, co-authors, advisors, and invited peers to give feedback}. \ldots'' 
Recent works~\citep{gao2024scaling},~\citet{lieberum2024gemma}, and~\citet{templeton2024scaling} provide mechanistic interpretability evidence suggesting that LLMs can learn concept-like features.

\para{Concept attributes.}
A concept can be represented as a sequence of sentences conveying semantic meaning, typically expressed in natural language. Here, we use `semantic meaning' (or `semantically meaningful') to refer to information that is both understandable and interpretable by humans~\citep{hurford2007semantics}.
As illustrated in the earlier example of the {\em meeting} concept, explaining a concept often involves multiple sentences, each contributing to a specific and meaningful aspect of the concept~\citep{piccinini2006splitting}. 
We refer to each of these aspects as a {\em concept attribute}.
For example, the sentence, \emph{``schedule a meeting on 28 July 2025, from 14:00 to 15:30, in Room 301 of the Computer Science Building.''} provides information about the meeting’s date, time, and location.

We now formalize the notion of a \emph{concept}.
Let $\Theta$ denote the set of all possible latent concepts, where each $\theta_t \in \Theta$ corresponds to the concept associated with a specific task $t$.
Here, task $t$ refers to the underlying objective, either explicitly specified in the prompt or implicitly expected from the response.
In the example of the \emph{meeting} concept, \emph{``Write an email" is the task} specified in the prompt, while the corresponding \emph{email draft containing meeting details} is expected as the response.
Let $\cA_{\theta_t} = \{A_{\theta_t,1}, \ldots, A_{\theta_t,m}\}$ denote the set of all attributes associated with a concept $\theta_t \in \Theta$.
Each concept $\theta_t$ is thus fully characterized by its set of attributes $\cA_{\theta_t}$.
We assume that each attribute $A_{\theta_t,a} \in \mathcal{A}_{\theta_t}$ can be accurately extracted from, or meaningfully represented in, a semantically coherent sequence of tokens drawn from the set $\cV$.

    \section{Prompt-Response Concept Model}
    \label{sec:latent_concept}

We first use the notion of concept to explain our proposed prompt-response model for LLMs, then use it to derive theoretical results that explain the relationship between prompt informativeness and response uncertainty by analyzing how uncertainty changes with the informativeness.

\subsection{Prompt-Response Concept Model}
\begin{figure}[!ht]
    \centering
    \vspace{-2mm}
    \includegraphics[width=\linewidth]{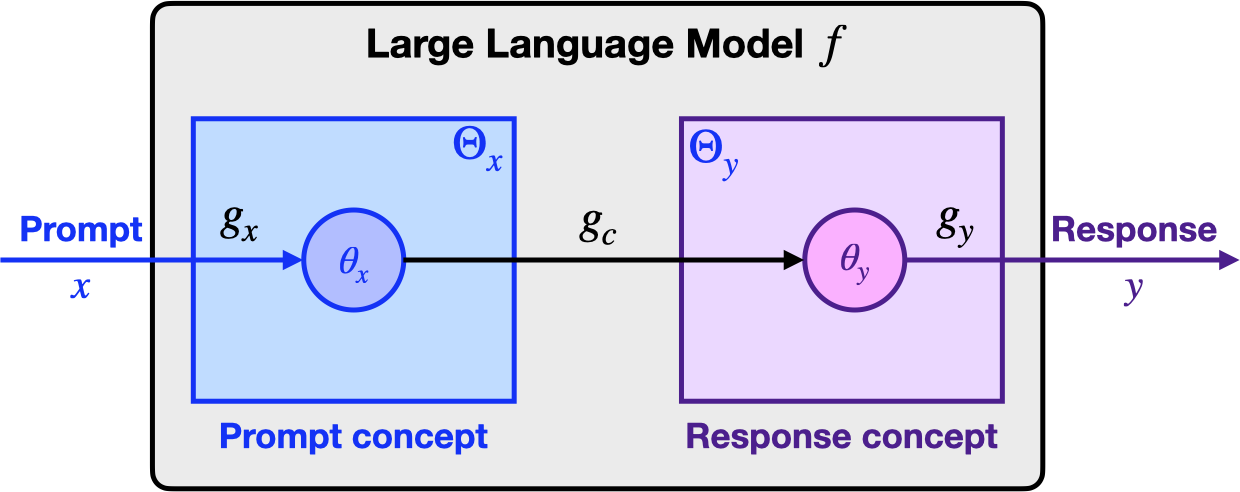}
    \vspace{-5mm}
    \caption{
        Prompt-Response concept model of LLM.
    }
    \label{fig:model}
    \vspace{-3mm}
\end{figure}

Building on prior work showing that LLMs implicitly learn to infer latent concepts during pretraining~\citep{xie2021explanation,hahn2023theory,zhang2023and}, we propose the {\bf\em prompt-response concept} (PRC) model of LLM. 
The PRC model conceptualizes how an LLM generates responses for a given prompt, providing a structured way to analyze the relationship between prompts and response uncertainty by distinguishing its sources.
The PRC model consists of three main components (as illustrated in \cref{fig:model}): prompt concept, response concept, and mapping functions. 

\para{Prompt concept.}
Let $\Theta_x$ denote the set of all latent concepts corresponding to prompts in set $\cX$. 
Each prompt $x \in \cX$, expressed as a sequence of tokens describing a task, maps to a concept $\theta_x \in \Theta_x$. We refer to this concept as the {\em prompt concept}. 
Intuitively, an LLM extracts the attributes of a latent prompt concept from input tokens, expressed as multiple semantically meaningful sentences in the prompt.
If these sentences in a prompt cannot be combined to describe a single concept, the LLM interprets them as representing multiple distinct concepts. 
Our experiments (\cref{fig:compositionality}) show that {\em adding semantically meaningful information from different concepts increases response uncertainty}.

\para{Response concept.}
Let $\Theta_y$ be the set of all latent concepts corresponding to responses in the set $\cY$. 
We refer to these concepts as the {\em response concept}.
Each response concept $\theta_y \in \Theta_y$ is uniquely associated with a response $y \in \cY$.
As prompts and responses follow different distributions and representations, we model them as separate concepts, in contrast to the ICL setting, which uses a single intermediate concept~\citep{xie2021explanation}.

\para{Mapping functions.}
To understand the relationship between the prompt, intermediate concepts, and the response, our PRC model conceptualizes the LLM $f$ as a composition of three mapping functions: prompt-concept mapping function $(g_x)$, concept-concept mapping function $(g_c)$, and concept-response mapping function $(g_y)$. 
First, the function $g_x$ maps the prompt $x$ to a latent prompt concept $\theta_x$ that we assume captures the understanding of the task and associated relevant information from the prompt. 
This concept is then transformed by $g_c$ into a response concept $\theta_y$, which encapsulates the core intent and structure of the desired output, which is then mapped to the actual response by $g_y$. 
Intuitively, $g_x$ captures the LLM's ability to understand the task specified by the prompt, $g_c$ models the reasoning process to derive the response concept from this understanding, and $g_y$ represents the verbalization of the response concept.

It is important to note that both the concept-concept mapping $g_c$ and the generation process $g_y$ are inherently stochastic. The stochasticity in $g_c$ means that for the same prompt concept, the model might stochastically choose different response concepts even if its input is fixed (e.g., explanation focusing on different aspects of the tasks). The stochasticity in $g_y$ corresponds to the token-sampling process (e.g., using temperature or nucleus sampling), allowing for different phrasings and semantically equivalent variations of the same response concept. These two sources of randomness are fundamental to the response variability observed in LLMs.
In the formal analysis that follows, we regard $g_y(\theta_y)$ as the \emph{response distribution} over $\cY$; the token‑by‑token mechanism is thus fully marginalized out. Thus, the overall LLM function can be expressed as $y = f(x) = g_y( g_c( g_x(x) ) )$.

To understand these mapping functions better, consider the following prompt: ``Write an email to schedule a meeting next Wednesday at 14:00 for one and a half hours in Room 301.''
First, the LLM uses a function $g_x$ to understand what this prompt is about; in this example, it recognizes the concept of a meeting being scheduled (the prompt concept) and then extracts the key attributes such as the meeting's date, time, and location.
It then applies $g_c$ to determine the required response, i.e., an email containing meeting details (the response concept).
Finally, $g_y$ verbalizes the response concept into the actual email, i.e., the LLM response, represented as a semantically coherent sequence of tokens drawn from the vocabulary set $\cV$.

As shown in \cref{fig:model}, both prompt and response concepts are latent components within the LLM. To generate the desired response, the prompt concept must include all task-relevant attributes, either explicitly extracted from the prompt or implicitly inferred from the LLM's pretrained knowledge. Intuitively, the prompt concept represents the LLM's internal understanding of the task. Better LLMs extract these attributes more reliably and leverage richer pretrained knowledge, producing responses with lower uncertainty and higher quality, as demonstrated in \cref{fig:medqa} and \cref{fig:medqa2}. 
When the prompt concept lacks sufficient task-relevant attributes, response variability increases, since the LLM can interpret and complete the task in multiple plausible ways (see \cref{fig:total_std_gpt-4-0613}).

    \subsection{Theoretical Results}
    \label{sec:theoretical_results}

Let $Z_c$ be a random variable representing a concept (with $c=x$ for prompt concept and $c=y$ for response concept) and $X_s$ be a random variable representing a prompt having semantic meaning $s$. We first introduce the assumptions under which our theoretical results hold.

\begin{assu}
    \label{assum:all} We assume an LLM is perfect if it has exactly learned the mapping functions of the PRC model, namely $g_x$, $g_c$, and $g_y$.
\end{assu}

\begin{figure*}
    \vspace{-5mm}
	\centering
	\subfloat[Sentence-level removal (MedQA dataset)]{\label{fig:sentence_medqa}
		\includegraphics[width=0.49\linewidth]{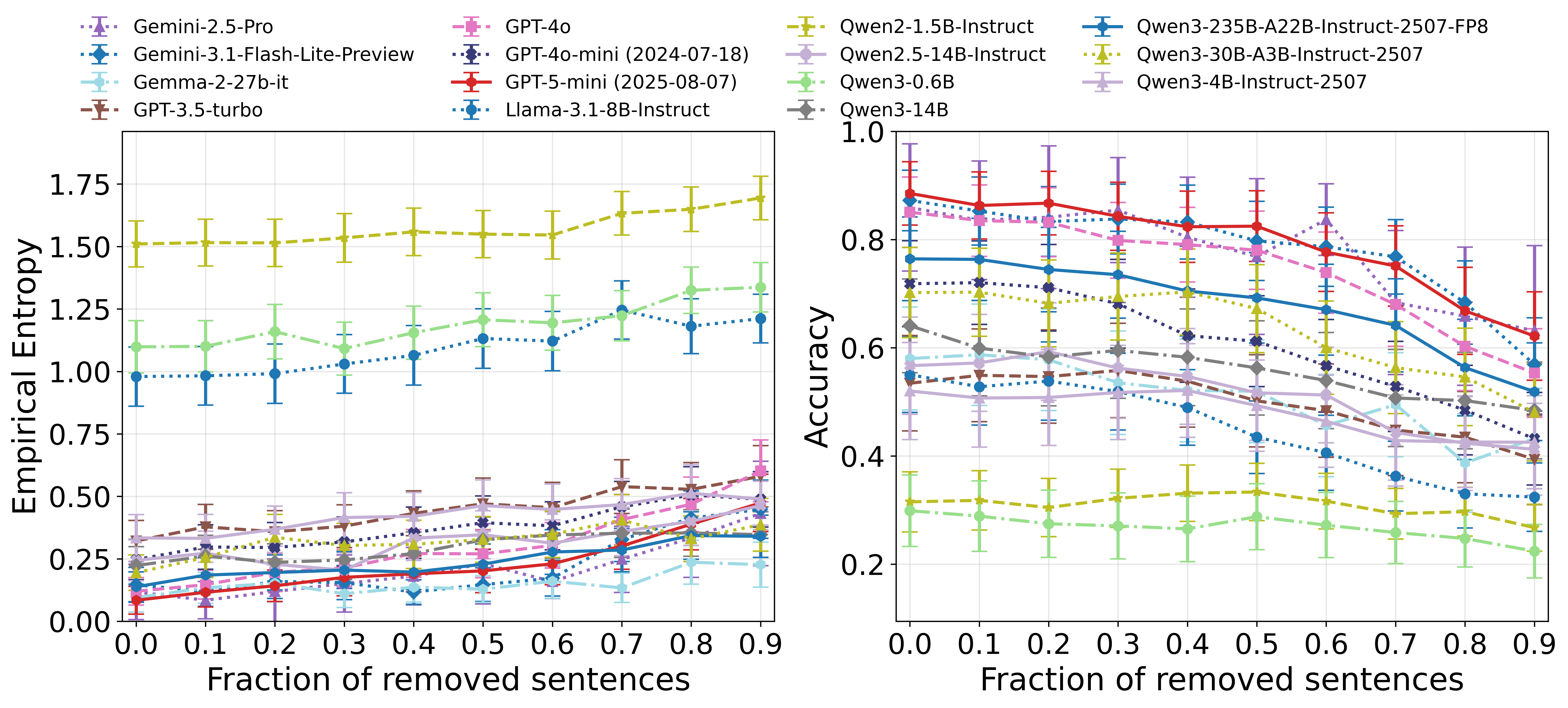}}
    ~~\subfloat[Attribute-level removal (MedQA dataset)]{\label{fig:attribute_medqa}
		\includegraphics[width=0.49\linewidth]{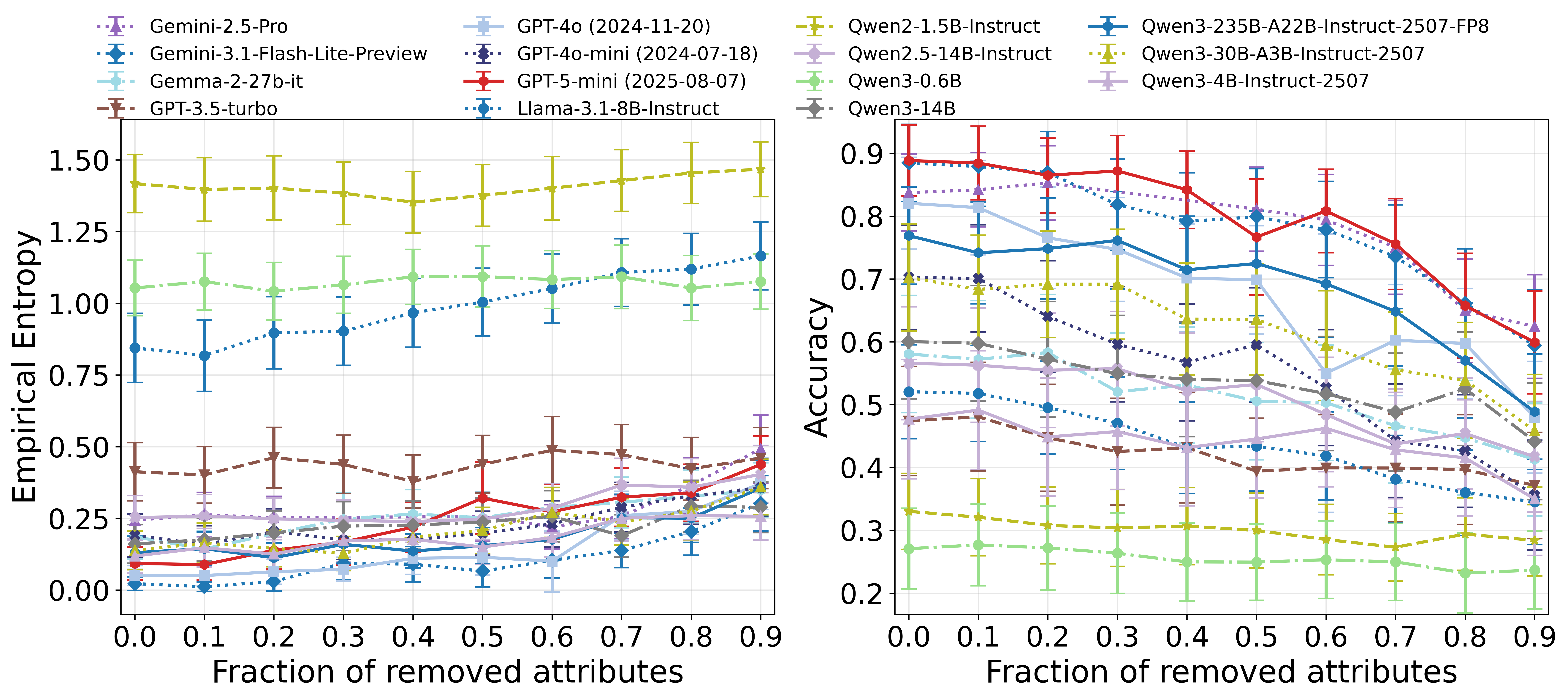}}
    \vspace{-3mm}
	\caption{ 
       Accuracy and entropy for the MedQA dataset after sentence- and attribute-level removal. 
       There is a strong negative correlation between accuracy and uncertainty, with less accurate models generally showing greater uncertainty in their responses. 
    }
	\label{fig:medqa}
    \vspace{-5mm}
\end{figure*}

\begin{figure*}[!ht]
    \vspace{-3mm}
	\centering
    \subfloat[Entropy (MedQA)]{\label{fig:empirical_entropy_medqa}
		\includegraphics[width=0.26\linewidth]{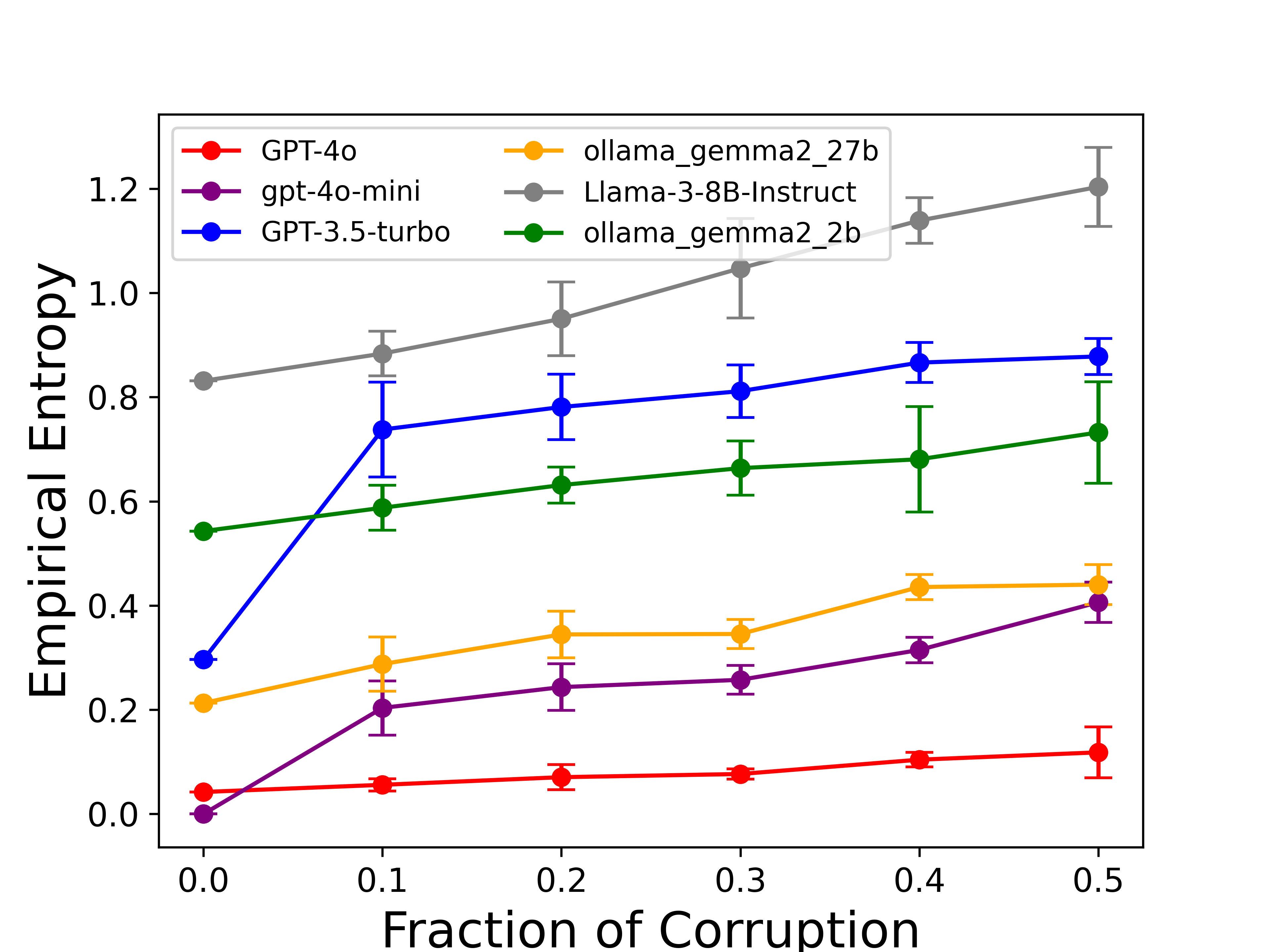}}
	\subfloat[Accuracy (MedQA)]{\label{fig:accuracy_medqa}
		\includegraphics[width=0.26\linewidth]{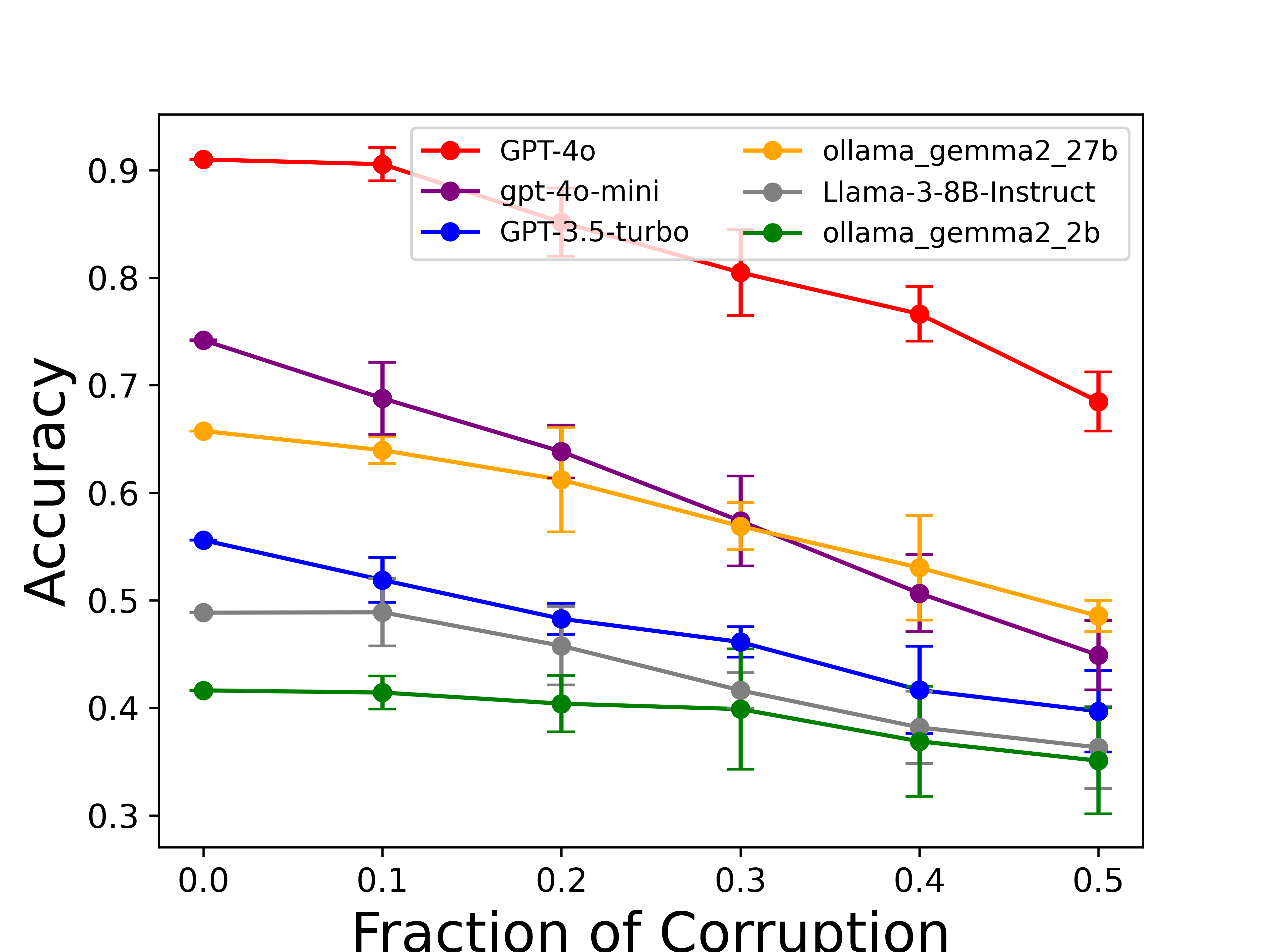}}
    \subfloat[Sentence-level removal]{\label{fig:sentence_medqa_comp}
		 \includegraphics[width=0.22\linewidth]{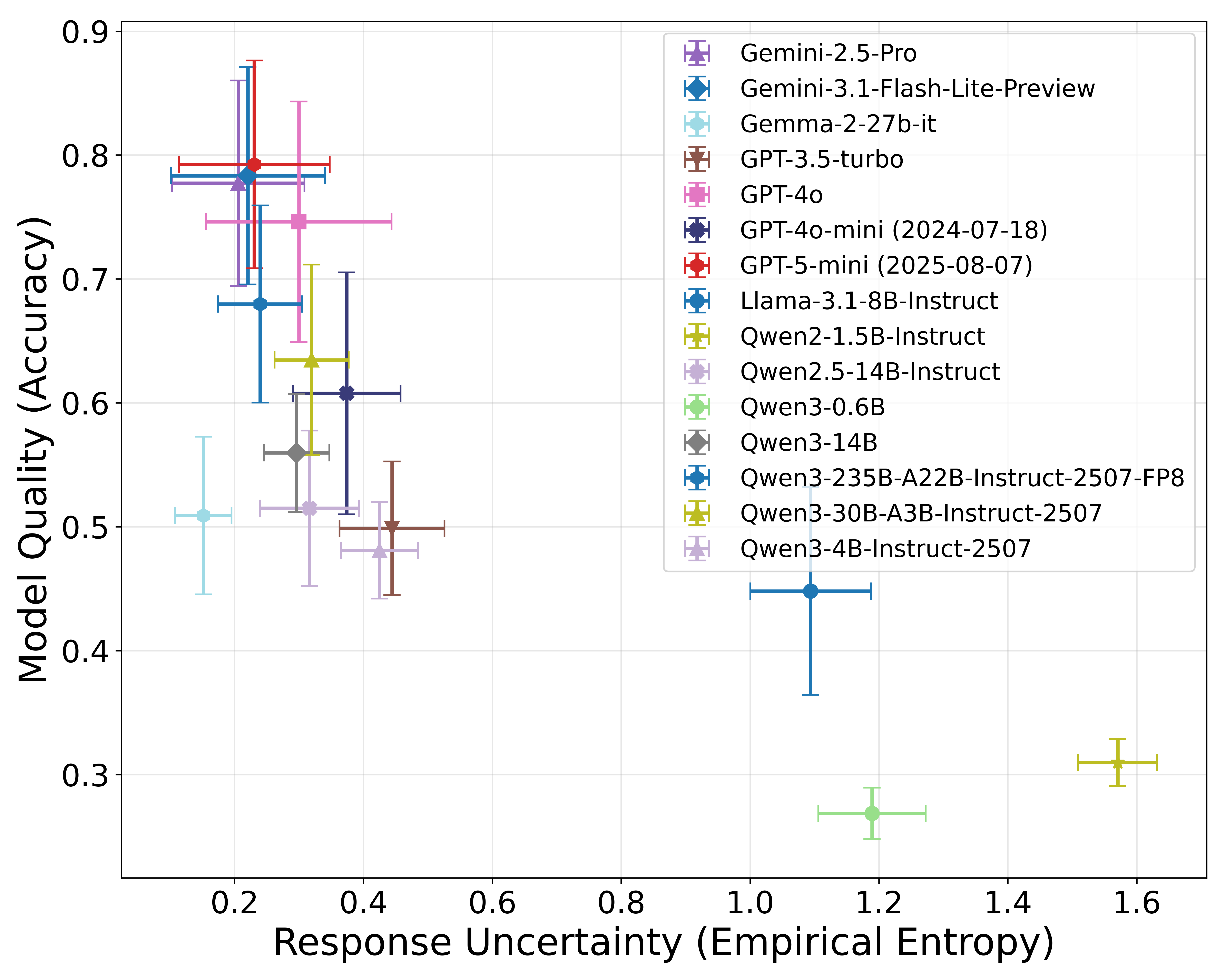}}
    \subfloat[Attribute-level removal]{\label{fig:attribute_medqa_comp}
		 \includegraphics[width=0.22\linewidth]{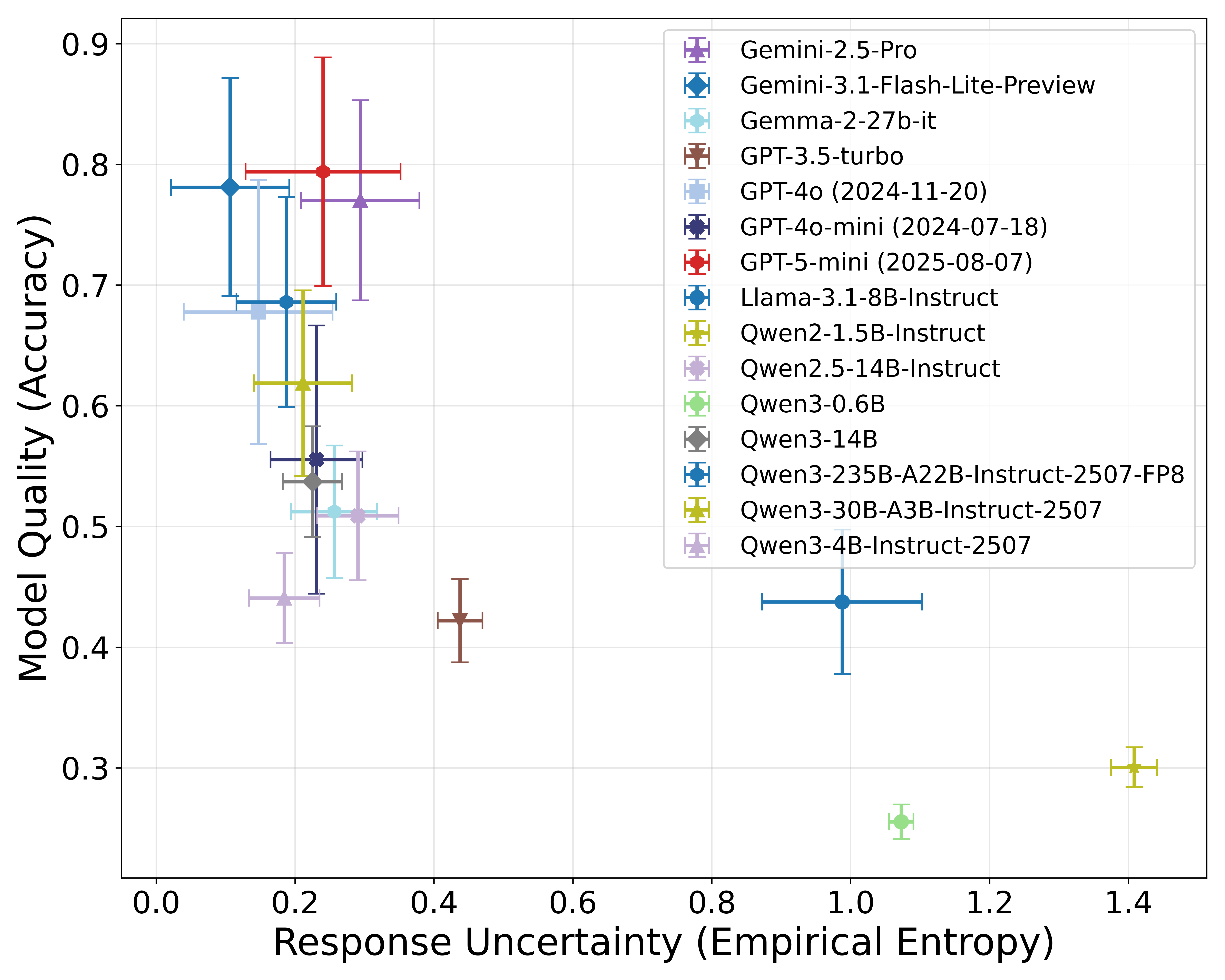}}
    \vspace{-3mm}
	\caption{ 
       \textbf{Left two figures.} Accuracy and entropy for the MedQA dataset after token-level removal. \textbf{Right two figures.} Model response quality (i.e., accuracy) vs. uncertainty (i.e., empirical entropy) for MedQA dataset after sentence- and attribute-level removal.
    }
	\label{fig:medqa2}
    \vspace{-5mm}
\end{figure*}

This assumption states that the LLM has perfectly learned the mapping functions used in our proposed PRC model.
Although this assumption may not hold exactly in practice, a better LLM is expected to estimate these mapping functions more accurately, resulting in lower uncertainty and higher-quality responses, as supported by our experimental results shown in \cref{fig:medqa} and \cref{fig:medqa2}.
Our first result shows the relationship between prompt informativeness and prompt concept uncertainty.

\begin{restatable}{lem}{concept}    
    \label{lem:concept}
    For any LLM with exactly leaned $g_x$ and two concepts $\theta_1, \theta_2 \in \Theta_x$, we have $\cX_{\theta_1} \cap \cX_{\theta_2} = \emptyset$ if $\theta_1 \neq \theta_2$. 
     Furthermore, $\sH{Z_{x}|X_{\theta_x}} = 0$.
\end{restatable}

The proof of \cref{lem:concept} and other missing proofs are given in \cref{asec:proofs}.
The first part of this result implies that prompts fully describing two different concepts cannot have the same semantic meaning, i.e., no two concepts share exactly the same semantic description. In other words, the prompts that fully describe two different concepts cannot have the same semantic meaning. 
The second part implies that the prompt concept is deterministic when the prompt contains all the information needed to complete the task. Next, we present a result showing that prompt concept uncertainty decreases as prompt informativeness increases.

\begin{restatable}{prop}{conceptEntropy}    
    \label{prop:concept_entropy}
    As $X_s$ represents more informative prompts (i.e., as more task-relevant information is included in the prompt), $\sH{Z_x|X_s}$ decreases.
\end{restatable}

Our next result links prompt informativeness to the response uncertainty of LLMs.

\begin{restatable}{thm}{responseUncertainty}    
    \label{thm:response_uncertainty}
    As $X_s$ represents more informative prompts, $\sH{Y|X_s}$ strictly decreases. Specifically $\sH{Y|X_s} \le \sH{Y|Z_y} + \sH{g_c(Z_x)|Z_x} + \sH{Z_x|X_s}$. Furthermore, $\sH{Y|X_s}$ converges to $\sH{Y_\tau|Z_y} + \sH{Y_r|Z_y} + \cE$, where $\cE$ decreases as the model quality improves.
    When \cref{assum:all} holds, $\cE \to 0$ as prompt informativeness increases.
\end{restatable}

Here, $\sH{Z_x|X_s}$ is the uncertainty due to prompt underspecification and imperfect LLM (particularly $g_c$), $\sH{g_c(Z_x)|Z_x}$ due to model quality, $\sH{Y_\tau|Z_y}$ due to task variability (e.g., response being outcome of coin toss), and
$\sH{Y_r|Z_y}$ due to semantic redundancy, more discussion about this decomposition is provided in \cref{asec:response_decomposition}.
The above two results suggest that response uncertainty decreases as prompt informativeness or model quality increases due to the uncertainty in the response concept. It can be understood through our PRC model: a highly informative prompt provides a strong and unambiguous initial concept, which acts as an `anchor' for the response generation. This anchor constrains the possible generated responses to a narrow set of semantically similar outcomes and thus reduces the final response entropy. Conversely, a vague prompt allows for many divergent generation paths, resulting in higher uncertainty.
Furthermore, when sufficient information is provided in a prompt and \cref{assum:all} holds, no uncertainty ($\cE$) remains due to the prompt concept and model quality. 
The remaining randomness in responses can be decomposed into two terms: $\sH{Y|Z_y}$, which represents the uncertainties due to task variability ($\sH{Y_\tau|Z_y}$) and semantic redundancy ($\sH{Y_r|Z_y}$).

The task variability ($\sH{Y_\tau|Z_y}$) is the aleatoric uncertainty in the prompt task. If the prompt task does not have a deterministic ground truth (e.g., forcing the model to answer the outcome of tossing a fair coin), the aleatoric uncertainty exists and is irreducible. In our experiments, we eliminate this source of uncertainty by using QA datasets for the prompt tasks with deterministic ground truth. We observe multiple realizations of the response concept for the same prompt concept across different iterations due to the imperfection of $g_c$, resulting in variations in responses.
However, as shown in \cref{fig:epsilon}, as the LLM quality improves, $\cE$ gets smaller, resulting in lower overall response uncertainty.
When an LLM can extract all task-relevant attributes, there should be no randomness in response due to the prompt and the remaining uncertainty in the response arises only from semantic redundancy ($\sH{Y_r|Z_y}$), which is the ability of the multiple responses expressing the same response concept, i.e., multiple responses are semantically equivalent~\citep{kuhn2023semantic}, which is the irreducible uncertainty (see \cref{thm:response_uncertainty}).

\begin{remark}[Compatibility with autoregressive decoding]
    \label{rem:autoregressive}
    During autoregressive decoding, the model re‑feeds the growing token sequence $(x,Y_{<t})$ at each step~$t$.  Although the input changes, \textbf{no new source of randomness is introduced}. Thus, \textbf{the uncertainty decomposition for the full sequence applies directly to each next‑token distribution}. No additional uncertainty arises from feeding back tokens that are already in‑distribution under~$\Theta_y$.
\end{remark}

\subsection{Epistemic and Aleatoric Response Uncertainty}
\label{sec:redundancy}
In machine learning literature, epistemic uncertainty is typically reduced by incorporating additional information, such as using a better model and additional training data~\citep{hullermeier2021aleatoric,lahlou2021deup,senge2014reliable,shaker2020aleatoric,valdenegro2022deeper,der2009aleatory}. 
In \cref{prop:concept_entropy}, $\sH{Z_c|X_s}$ represents the epistemic uncertainty in the latent concepts.\footnote{It is termed the {\em semantic entropy} in~\citet{kuhn2023semantic}. This paper studies it through the lens of uncertainty reduction.} 
We have demonstrated that $\sH{Z_c|X_s}$ is strictly reduced with a prompt that contains more attributes of the relevant concept(s). 
Thus, increasing the information about the concept in a prompt can lead to more reliable and consistent responses by reducing the epistemic uncertainty in the latent concept~\citep{hullermeier2021aleatoric}.
However, if the prompt contains information that is irrelevant to the task, the response uncertainty can also increase, as demonstrated in \cref{fig:uncertainty_gpt35_insert_semantically_meaningful_strings}.

Due to the model's inability to learn perfect mappings during training, as well as the possibility that the task in the prompt may not have a deterministic ground truth, the mapping from the prompt concept to the response concept cannot be deterministic. In \cref{thm:response_uncertainty}, $\cE$ captures the uncertainty due to model quality. In scenarios where model parameters are allowed to be modified, this uncertainty becomes epistemic and can be reduced as the model quality improves (\cref{fig:epsilon}). The uncertainty due to task variability ($\sH{Y_\tau|Z_y}$) is the irreducible aleatoric uncertainty of the given prompt task. 
The remaining response uncertainty is characterized by the term $\sH{Y_r|Z_y}$ that arises due to {\em semantic redundancy}. It can be further reduced in two ways: use fine-tuning or prompting to instruct the model to reply with a certain fixed style.\footnote{The response style can be viewed as an implicit concept, so {\em semantic redundancy} can be reduced by providing relevant style information in the prompt to guide the model response.} 
Due to semantic equivalence, {\em semantic redundancy} is generally not detrimental to the desired information.
It is possible that a model can result in low response uncertainty but poor response quality~\citep{singh2023confidence,li2024think,fu2025multiple}.

    \section{Experiments}
    \label{sec:experiments}

To validate our proposed prompt-response concept model, we empirically demonstrate different aspects of our proposed model in different settings, the details of which are as follows. For instruction-fine-tuned LLMs, their prompts are usually in the form of some tasks from the user (i.e., `explain to me why the sky is blue'). Our experiments treat a relatively simple task as a `concept' and a complex task as a composition of multiple concepts. All selected datasets have deterministic ground truths to eliminate the aleatoric uncertainty in the tasks.

\subsection{Informative Prompt vs. Response Quality}
\label{subsec:response_quality}
It is worth noting that low uncertainty in model responses does not necessarily indicate high response quality, as an LLM can produce outputs with very low uncertainty while being blindly confident in incorrect answers. This behavior is problematic and can lead to hallucinations~\citep{huang2023survey}. To ascertain the actual relationship between prompt, model response uncertainty, and quality, we further investigated the relationship between the effective token count of the prompt and model response quality. 
To assess if the reduction in uncertainty translates to improved output quality, we test the model's output accuracy when answering the multiple-choice questions (MCQs

\para{Datasets.} We selected $100$ MCQs from the Medical Meadow MedQA dataset~\citep{jin2021disease}, a benchmark for medical question answering. Due to space constraints, we report additional results on general-domain reasoning datasets, ARC~\citep{allenai:arc}, OpenBookQA~\citep{OpenBookQA2018}, and RACE~\citep{lai-etal-2017-race}, in \cref{asec:experiment_results}.

\para{Controlling prompt informativeness.}
We use three methods for controlling prompt informativeness: sentence-level removal, attribute-level removal, and token-level removal.\\
\noindent\textbf{\step{1} Sentence-level removal.}
We iteratively select an increasing fraction of randomly selected sentences from the context of the questions.\\
\noindent\textbf{\step{2} Attribute-level removal.}
We first prompt the LLM to identify key concepts in the context (e.g., entities, events, and locations), along with their associated attributes, which are defined as specific facts or details that describe each concept. These attributes are extracted as exact quoted spans from the original text and then selectively removed. Unlike sentence-level removal, which indiscriminately deletes entire sentences, attribute-level removal targets semantically meaningful information that is critical for comprehension. This approach enables a more fine-grained analysis of model degradation when relevant facts are missing, rather than merely reducing the amount of text.\\
\noindent\textbf{\step{3} Token-level removal.}
We iteratively select an increasing fraction of randomly selected tokens in the context by replacing them with space tokens. For each question, we set the decoding temperature to $1$ and sample $100$ responses to assess output variability. We repeat this process using $5$ different random seeds to select the tokens for replacement.

As the fraction of removed content increased, we incrementally expanded the set of randomly selected elements (sentences, attributes, or tokens), while keeping previously removed elements fixed. This controlled procedure ensures that changes in model performance are attributable to the increasing degree of information removal, rather than randomness in the selection process. As a result, we can more reliably assess the impact of reduced prompt informativeness on response quality.

\para{Results.}
In \cref{fig:medqa,fig:empirical_entropy_medqa,fig:accuracy_medqa,fig:race1}, we plot the accuracy for the responses of different open-source and black-box LLMs on the same set of MCQs. 
As the fraction of removed content in the prompt increases, the accuracy monotonically decreases across all tested models and removal strategies.
For each random seed, we also compute the empirical conditional entropy $\sH{Y|X}$ of the model responses for the given questions\footnote{We assume a uniform distribution over questions, i.e., $p(x)=\frac{1}{100}$. In the absence of access to the true conditional distribution $p(y|x)$, we estimate $\sH{Y|X} = - \sum_x p(x) \sum_y \hat{p}(y|x) \log \hat{p}(y|x)$, where $\hat{p}(y|x)$ is obtained from the empirical response distribution. This metric is particularly suitable for MCQ settings, where the model's effective output is a single discrete choice.} (\cref{fig:medqa,fig:empirical_entropy_medqa,fig:race1}) as a measure of response uncertainty.
As more information is removed, response uncertainty consistently increases for all models (and monotonically for larger LLMs), revealing a clear negative correlation between $\sH{Y|X}$ and response accuracy.
These results support our hypothesis that greater prompt informativeness reduces response uncertainty and improves output quality.
Furthermore, as shown in \cref{fig:sentence_medqa_comp,fig:attribute_medqa_comp,fig:sentence_race_comp,fig:attribute_race_comp}, models with higher accuracy exhibit lower empirical conditional entropy, corroborating our characterization of $\cE$ in \cref{thm:response_uncertainty}.
An interesting finding across all tasks is that sufficient prompt informativeness can allow a smaller model to match or even outperform a more capable one, e.g., Gemma2 27B model with full context surpassed GPT-4o-mini with up to 80\% context. This result underscores that a high-quality prompt can effectively compensate for a model's inherent scale, enabling less powerful models to achieve highly competitive performance.

\begin{figure*}
	\centering
    \includegraphics[width=.7\linewidth]{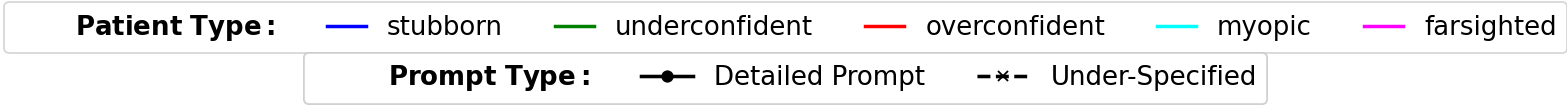} \\ 
    \vspace{-3.5mm}
    \subfloat[][{\parbox[t]{0.7\linewidth}{Farsightedness }}]{\includegraphics[width=.25\linewidth]{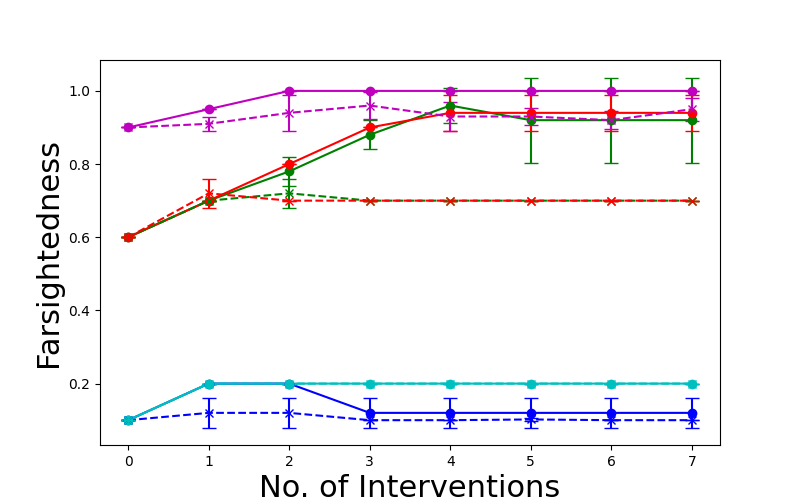}}
    \subfloat[][{\parbox[t]{0.7\linewidth}{Improvement prob.}}]{\includegraphics[width=.25\linewidth]{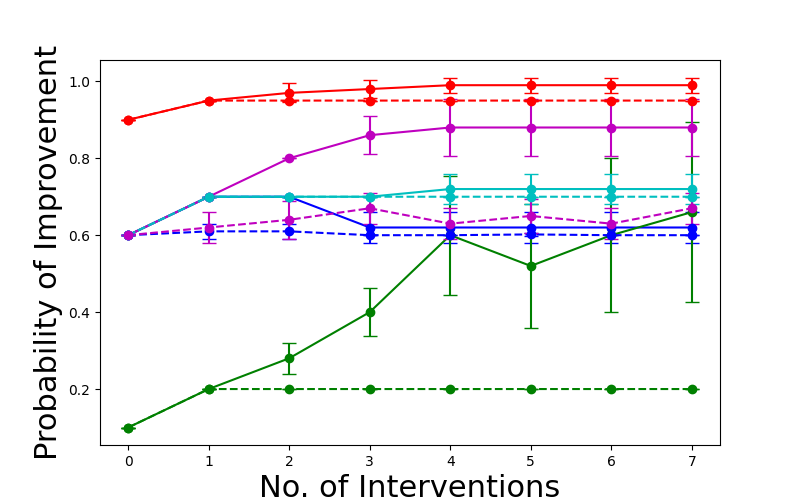}}
    \subfloat[][{\parbox[t]{0.7\linewidth}{Disengagement prob.}}]{\includegraphics[width=.25\linewidth]{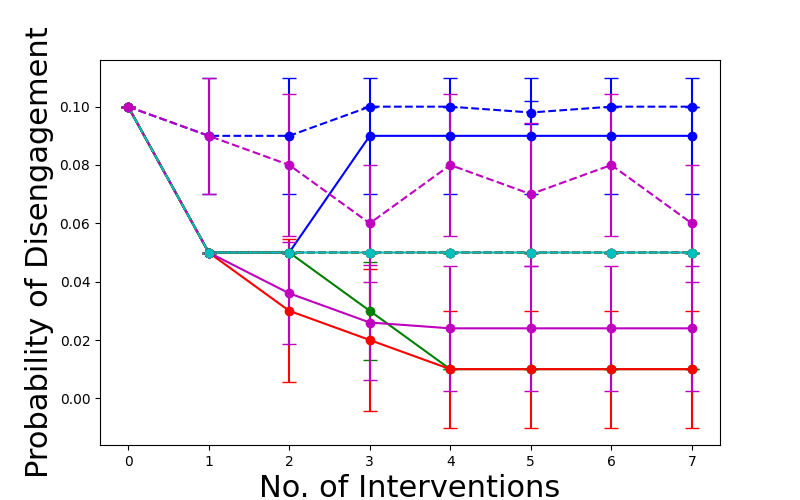}}
    \subfloat[][{\parbox[t]{0.7\linewidth}{Avg. engagement rate}}]{\includegraphics[width=.25\linewidth]{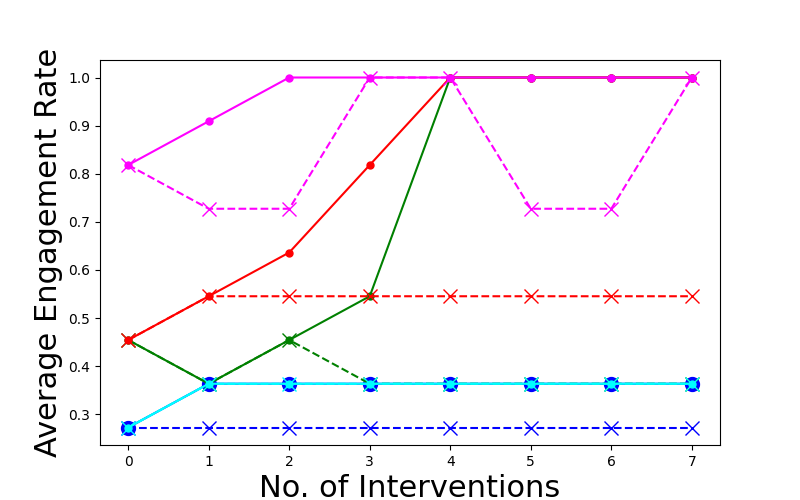}}
    \vspace{-2mm}
	\caption{ 
       Results from the PT intervention simulation. For (a) and (b), a higher value indicates more improvement. For (c), a lower value indicates more improvement. (d) is the patient's optimal policy averaged across all health states, obtained from analytically solving the MDP. A higher value indicates, on average, the patient agent is more likely to continue engaging in PT. Overall, we observed that prompts with more information gave rise to more consistent improvement compared to prompts with less information across all patient types.
    }
	\label{fig:mHealth}
    \vspace{-5mm}
\end{figure*}

\subsection{mHealth Intervention Usecase}
\label{subsec:Simulation}
We now demonstrate the effectiveness of our proposed approach in a real-world simulation use case in an mHealth setting. We adapt the formulation from~\citet{shin2022modeling}; both the app and the user act as reinforcement learning agents. The app agent's objective is to encourage the user agent to adhere to the PT routine. The user agent moves along a chain with  $N$ states, where a higher state number represents a healthier physical state, and state $N$ indicates completion of the PT routine. 

We conduct the intervention simulation experiment with LLM to compare the effect of prompts with different informativeness levels on the intervention outcome. The experiment concludes that when the prompt provides the LLM (i.e., the app agent) with more information about the patient's intentions and the strategies it can employ, the efficiency of the intervention improves consistently for different patient types compared to scenarios without the additional information. A more detailed description of this experiment is given in \cref{appendix:Simulation_description}.
\textit{Due to space constraints, we have provided additional experimental results in \cref{asec:experiment_results}.}

    \section{Related Work}
    \label{sec:related_work}

While uncertainty quantification has been extensively studied in machine learning, its application to LLMs remains relatively underexplored.
\citet{kadavath2022language} investigated the extent to which LLMs can accurately self-evaluate their knowledge and how LLM calibration can improve response quality. Their goal of calibrating LLMs was to align the variability in model responses with actual uncertainty so that response variations genuinely reflect the model's lack of relevant knowledge given the prompt. 
\citet{kuhn2023semantic} introduced the concept of semantic entropy to more precisely quantify the uncertainty in the informational content of model responses, accounting for semantically equivalent variations and eliminating noise from paraphrasing.
\citet{lin2023generating} proposed a method for estimating uncertainty in black-box LLMs for question-answering tasks by measuring response similarity using a Natural Language Inference (NLI) model, combined with simple dispersion-based measures. 
In a related but orthogonal line of work, \citet{wagle2023empirical} conducted empirical studies on pretrained language models (PLMs) and found that, while PLMs are often overconfident, larger models tend to be better calibrated, i.e., confidence estimates more closely aligned to actual prediction accuracy.
Similar to our work, \citet{ling2024uncertainty} aim to understand and quantify LLM response uncertainty by decomposing it into aleatoric and epistemic components. However, their study is limited to the setting of ICL and assumes a correlation between model response accuracy and uncertainty without direct empirical examination. In contrast, we explicitly investigate whether lower response uncertainty necessarily implies higher response quality. 
While we also adopt an entropy-based uncertainty measure, similar to \citet{kuhn2023semantic}, \citet{lin2023generating}, \citet{wagle2023empirical}, and \citet{farquhar2024detecting}, our focus is on understanding the LLM's response uncertainty and how increasing informativeness and model quality can be used as a principled way to reduce response uncertainty.
We defer additional related works on asymptotic behaviors of LLMs to \cref{asec:additional_related_works}.

    \section{Conclusion}

This paper highlights the importance of understanding how prompts and models affect response uncertainty in LLMs. By examining prompt informativeness, we show that more informative prompts, combined with better models, lead to lower uncertainty. To formalize this, we introduce the prompt-response concept (PRC) model, which captures how LLMs generate responses and helps identify sources of uncertainty.
These insights provide a principled approach to improving prompt design, crucial for safe and effective use of LLMs in decision-making, especially in high-stakes domains like healthcare. Future research may refine the PRC model and explore its applicability in other areas requiring reliable LLM responses.

    \section{Limitations}
    \label{sec:limitations}

This work has two key limitations, which we outline in this section to guide future research.

\para{Idealistic nature of the PRC model.}
It is worth noting that the PRC model that we proposed in this paper assumes an idealized version of LLMs. As empirically demonstrated, while models such as GPT-3.5-Turbo, GPT-4, Llama 2, and Llama 3 exhibit behaviors largely according to our predictions, there are still some modes in which they deviate (e.g., Qwen2\_1.5b plot). This is likely in those cases where LLM does not know the mapping perfectly. For example, \citet{lu2021fantastically} showed that the order of exemplars in ICL influences the model response quality. Our model does not capture this phenomenon. However, the authors showed that in the same work, the order of examples tends to have less effect as model quality gets better. Other such examples include jailbreak by asking the model to repeat the same single-token word for a sufficiently long period of time \citep{nasr2023scalable}, by appending adversarially crafted tokens \citep{zou2023universal}, and translating the prohibited request into a low-resource language \citep{yong2023low}. Similarly, it was observed that adversarial attacks tend to have lower success rates as the model becomes more capable. While further investigation is needed to incorporate the adversarial behavior of LLMs into this framework, the more capable LLMs are less prone to these failure modes. Our model can more effectively explain them.

\para{LLMs for human behavior simulation.}
Research exploring the parallels between human behavior and reasoning patterns and those of LLMs, as well as the adaptation of LLMs as human substitutes in diverse studies, is detailed in \citet{aher2023using}, \citet{argyle2023out}, \citet{binz2023using}, and \citet{dasgupta2022language}. These studies frequently demonstrate LLMs' capacity for human-like responses, leading many to regard them as viable alternatives. This paper, however, needs to delve into the appropriateness of this substitution, deferring to other works for such discussion.

    \section*{Acknowledgment}
    \label{sec:acknowledgment}
    This research is supported by the National Research Foundation, Singapore under its National Large Language Models Funding Initiative (AISG Award No: AISG-NMLP-$2024$-$001$).
This research is supported by the National Research Foundation (NRF), Prime Minister’s Office, Singapore under its Campus for Research Excellence and Technological Enterprise (CREATE) programme. The Mens, Manus, and Machina (M3S) is an interdisciplinary research group (IRG) of the Singapore MIT Alliance for Research and Technology (SMART) centre.

    \bibliography{references}

    \clearpage
    \onecolumn
    \appendix    

\section{Leftover Details}
\subsection{Related Works}
\label{asec:additional_related_works}
\para{Explanation for asymptotic behaviors of LLMs.}
Several recent efforts have been made to develop frameworks that explain the surprising emergent behaviors of LLMs.
\citet{zhang2023and} demonstrate that the attention mechanism in LLMs approximates Bayesian model averaging in the ICL setting. 
\citet{wang2024large} conceptualize real-world LLMs as latent variable models, suggesting that these models operate as implicit topic models by inferring latent conceptual variables from prompts. Notably, \citet{xie2021explanation} interpret ICL as an instance of implicit Bayesian inference over latent concepts acquired during pretraining. 
However, their analysis is limited to characterizing zero-one error in the asymptotic case of infinite exemplars, and their hidden Markov model-based mathematical formulation is tailored specifically to the ICL setting, making it unsuitable for analyzing chain-of-thought or conversational-style responses. 
Furthermore, their theoretical results quantify the mode of the posterior predictive distribution and do not address uncertainty quantification. 
\citet{hahn2023theory} further extend this line of work by incorporating greater flexibility and complexity in the exemplars, yet similarly provide only asymptotic bounds on classification error. 
In contrast, our work complements these approaches by focusing on how the response uncertainty varies with finite-length prompts, providing a more practical understanding of LLM behavior in real-world scenarios.

\subsection{Proofs}
\label{asec:proofs}

\concept*
\begin{proof}
    The result holds trivially for the case in which $\cX_{\theta}$ for any $\theta \in \Theta_x$ is an empty set.
    As discussed in \cref{sec:latent_concept}, each concept is completely characterized by all of its attributes; therefore, two different concepts cannot have the same set of attributes, i.e., $\mathcal{A}_{\theta_i} \neq \mathcal{A}_{\theta_j}$ if $i \neq j$. 
    As our PRC model assumes any attribute can be perfectly expressed by some sequence of tokens,
    any attribute $a_{\theta_i,k} \in \mathcal{A}_{\theta_i}$ can be expressed by a sequence of tokens. We denote the set of all possible sequences of tokens by $\cX_{s(a_{\theta_i,k})}$, where $s(a_{\theta_i,k})$ denotes the semantic meaning of $a_{\theta_i,k}$. 
    Therefore, the set of attributes $\mathcal{A}_{\theta_i}$ is expressed as a sequence of tokens $X_{\theta_i} \in \cX_{\theta_i}$, where $\cX_{\theta_i} = \cC\lp \{\cX_{s(a_{\theta_i,k})}\}_{k=1}^n \rp$ in which $n= |\mathcal{A}_{\theta_i}|$ and operator $\cC$ applied in the following way:
    
    \begin{itemize}
        \item Chooses one element $x_{s(a_{\theta_i,k})} \in  \cX_{s(a_{\theta_i,k})}$ for each $k \in \{1,2, \ldots, n\}$;
    
        \item Create a set $\cS_{\theta_i}$ containing all the selected elements $x_{s(a_{\theta_i,k})}$. Then, concatenate these elements in $\cS_{\theta_i}$ to form sequences by exhausting all possible orderings and use this collection of sequences to form a new set $\cX'_{\theta_i}$.
    
        \item Repeat step 1 and 2 for all possible sets $\cS_{\theta_i}$ and generate all possible $\cX'_{\theta_i}$. Finally, take the union of all such $\cX'_{\theta_i}$ sets to form a new set. Since this set consists of all possible sequences that are semantically equivalent and fully characterize $\theta_i$, it is exactly $\cX_{\theta_i}$.
    \end{itemize}

    Intuitively, $\cC$ takes all token sequences that fully characterize each attribute of $\theta_i$ and generates all possible concatenations that together fully characterize $\theta_i$.
    By the PRC model, for every $\theta \in \Theta_x$, the set $\cX_{\theta}$ is non-empty. We now argue that $\cX_{\theta_i} \cap \cX_{\theta_j} = \emptyset$ whenever $i \neq j$. By definition, $\cX_{\theta_i}$ consists precisely of those token sequences whose \emph{complete} semantic meaning is $\mathcal{A}_{\theta_i}$: a sequence $x \in \cX_{\theta_i}$ conveys all attributes of $\theta_i$ and no attributes that would additionally constitute a different concept's full attribute set. Suppose for contradiction that $x \in \cX_{\theta_i} \cap \cX_{\theta_j}$ with $i \neq j$. Then $x$ fully characterizes both $\mathcal{A}_{\theta_i}$ and $\mathcal{A}_{\theta_j}$. Since two concepts are distinct if and only if their attribute sets differ ($\mathcal{A}_{\theta_i} \neq \mathcal{A}_{\theta_j}$ iff $i \neq j$), one attribute set must contain an attribute absent from the other, say $A_k \in \mathcal{A}_{\theta_j} \setminus \mathcal{A}_{\theta_i}$. But if $x$ fully characterizes $\theta_j$, it must convey $A_k$; and if $x$ also fully characterizes $\theta_i$, then $A_k$ would be an attribute of $\theta_i$ (since every semantically meaningful attribute conveyed by a concept-$\theta_i$ sequence is, by definition, an attribute of $\theta_i$), contradicting $A_k \notin \mathcal{A}_{\theta_i}$. Hence no such $x$ exists, and $\cX_{\theta_i} \cap \cX_{\theta_j} = \emptyset$.

    Since the first part of \cref{lem:concept} is non-trivially true in our framework, given any $x \in \cX_{\theta_x}$, there exists a unique $\theta_x \in \Theta_x$ such that $p(Z_{x}=\theta_x \mid x)=1$ and $p(Z_{x}=\theta_x \mid x')=0$ for all $x' \notin \cX_{\theta_x}$. Throughout, we adopt the standard information-theoretic convention $0 \log 0 := 0$ (since $\lim_{p \to 0^+} p \log p = 0$).  Therefore,
    \begin{align*}
        \sH{Z_{x} \mid X_{\theta_x}}
            &= -\sum_{x \in \cX_{\theta_x}} P(x) \sum_{z \in \Theta_x} P(z \mid x) \log P(z \mid x) \\
            &= -\sum_{x \in \cX_{\theta_x}} P(x) \left(
                  \sum_{z \in \Theta_x \setminus \{\theta_x\}} P(z \mid x) \log P(z \mid x)
                  + P(\theta_x \mid x) \log P(\theta_x \mid x)
               \right) \\
            &= -\sum_{x \in \cX_{\theta_x}} P(x) \left(
                  \sum_{z \in \Theta_x \setminus \{\theta_x\}} 0 \cdot \log 0 + 1 \cdot \log 1
               \right)
               \quad \text{(using the convention above)} \\
            &= -\sum_{x \in \cX_{\theta_x}} P(x) \cdot 0 = 0.
    \end{align*}
    Note that this requires the LLM to have exactly learned $g_x$, i.e., $p(Z_x = \theta_x \mid x) = 1$ for every $x \in \cX_{\theta_x}$. Therefore, $\sH{Z_x \mid X_{\theta_x}} = 0$ holds under \cref{assum:all}.
    
\end{proof}

\conceptEntropy*
\begin{proof}
    We prove the result in two steps: (i) each additional piece of non-redundant, attribute-conveying information strictly reduces $\sH{Z_x \mid \cdot}$, and (ii) the process reaches zero entropy in finitely many steps.

    \paragraph{Step 1: One non-redundant attribute strictly reduces prompt-concept entropy.}
    Let $X_s$ be a prompt with semantic attributes $\alpha_s \subsetneq \mathcal{A}_{\theta_x}$ (i.e., the prompt conveys some but not all attributes of $\theta_x$). Then there exists a sequence $X_{\theta_x} \in \cX_{\theta_x}$ that shares semantic content with $X_s$, implying $Z_x$ and $X_s$ are statistically dependent. Hence $I(Z_x; X_s) > 0$, and by the mutual information identity:
    \begin{equation*}
        \sH{Z_x \mid X_s} = \sH{Z_x} - I(Z_x; X_s) < \sH{Z_x}.
    \end{equation*}
    
    \paragraph{Step 2: Iterative reduction via the chain rule of conditional mutual information.}
    Suppose $\sH{Z_x \mid X_s} > 0$ after conditioning on $X_s$. Then there exists at least one attribute $A_k \in \mathcal{A}_{\theta_x} \setminus \alpha_s$ that is not yet conveyed by $X_s$, otherwise \cref{lem:concept} would give $\sH{Z_x \mid X_s} = 0$.
    By the PRC model assumption, any attribute can be expressed by a token sequence; therefore, there exists a prompt segment $X''_s$ that conveys $A_k$ and nothing else from $\mathcal{A}_{\theta_x}$ (i.e., $A_k$ is non-redundant given $X_s$). Since $A_k \notin \alpha_s$, we have $I(Z_x; X''_s \mid X_s) > 0$, and by the chain rule for conditional mutual information:
    \begin{align*}
        \sH{Z_x \mid X_s, X''_s}
            &= \sH{Z_x \mid X_s} - I(Z_x; X''_s \mid X_s) \nonumber \\
            &< \sH{Z_x \mid X_s}.
    \end{align*}
    Writing $X^{(1)}_s = X_s$ and $X^{(k+1)}_s = (X^{(k)}_s,\, X''{}^{(k)}_s)$, where each $X''{}^{(k)}_s$ conveys the next uncovered attribute of $\theta_x$, we obtain the strictly decreasing chain
    \[
        \sH{Z_x \mid X^{(k+1)}_s} < \sH{Z_x \mid X^{(k)}_s} < \cdots < \sH{Z_x}.
    \]
    Since $\mathcal{A}_{\theta_x}$ is finite (say $|\mathcal{A}_{\theta_x}| = m$), after at most $m$ such steps the prompt $X^{(m)}_s$ fully specifies $\theta_x$, and by \cref{lem:concept} $\sH{Z_x \mid X^{(m)}_s} = 0$.
    Therefore, $\sH{Z_x \mid X_s}$ strictly decreases as $X_s$ conveys more attributes of $\theta_x$, reaching zero exactly when $X_s \in \cX_{\theta_x}$.
\end{proof}    
    
\responseUncertainty*
\begin{proof} 
    To simplify notation, we use $Y$ instead of $Y_{\theta_y}$ and assume the model response is complete (i.e., the last token is the `EOS' token).
    By design, $Z_x$ and $Z_y$ are discrete random variables (\cref{sec:latent_concept}). We consider the general setting where $g_c$ is stochastic.
    Under the PRC model, the prompt influences the response only via the prompt concept:
    \[
      X_s \;\to\; Z_x \;\to\; Z_y \;\to\; Y,
    \]
    forming a Markov chain. In particular, $Y \perp\!\!\!\perp X_s \mid Z_x$, because $Z_y = g_c(Z_x)$ depends on $X_s$ only through $Z_x$, and $Y = g_y(Z_y)$ depends on $Z_x$ only through $Z_y$.  Consequently:
    \begin{equation}
      \label{eq:markov_decomp}
      \sH{Y \mid X_s} = \mathbb{E}_{Z_x \mid X_s}\!\bigl[\sH{Y \mid Z_x}\bigr].
    \end{equation}
    We can rewrite $\sH{Y \mid X_s}$ in \cref{eq:markov_decomp} as the average of $\sH{Y \mid Z_x}$, a quantity that does \emph{not} depend on $X_s$, over the posterior distribution of the prompt concept given the prompt.

    Since $\sH{f(X) \mid Y} \leq \sH{f(X),\, X \mid Y}$ (adding a variable can only increase entropy),
    \begin{align}
        \label{eqn:res_concept}
        \sH{Z_y \mid X_s}
            &\leq \sH{Z_y, Z_x \mid X_s}
             = \sH{Z_x \mid X_s} + \sH{Z_y \mid Z_x, X_s}
             = \sH{Z_x \mid X_s} + \sH{Z_y \mid Z_x},
    \end{align}
    where the last step uses $Z_y \perp\!\!\!\perp X_s \mid Z_x$ from the Markov chain.
    Using the chain rule of entropy and $Y \perp\!\!\!\perp X_s \mid Z_y$:
    \begin{align*}
        \sH{Y \mid X_s}
            &= \sH{Y \mid Z_y, X_s} + \sH{Z_y \mid X_s} - \sH{Z_y \mid Y, X_s} \\
            &= \sH{Y \mid Z_y} + \sH{Z_y \mid X_s} - \sH{Z_y \mid Y, X_s}
               \qquad\text{($Y \perp\!\!\!\perp X_s \mid Z_y$)} \\
            &\leq \sH{Y \mid Z_y} + \sH{Z_y \mid X_s}
               \qquad\text{($\sH{Z_y \mid Y, X_s} \geq 0$)} \\
            &\leq \sH{Y \mid Z_y} + \sH{Z_x \mid X_s} + \sH{Z_y \mid Z_x}
               \qquad\text{(from \cref{eqn:res_concept})} \\
            &= \sH{Y \mid Z_y} + \sH{g_c(Z_x) \mid Z_x} + \sH{Z_x \mid X_s}.
               \qquad\text{($Z_y = g_c(Z_x)$)}
    \end{align*}
    By \cref{prop:concept_entropy}, $\sH{Z_x \mid X_s}$ strictly decreases as $X_s$ becomes more informative, so the right-hand side strictly decreases. This completes the proof of the first part.
    
    From the chain-rule step above (second equality), we read off the exact identity:
    \begin{equation}
        \label{eq:exact_decomp}
        \sH{Y \mid X_s}
            = \sH{Y \mid Z_y}
              + \underbrace{\sH{Z_y \mid X_s} - \sH{Z_y \mid Y, X_s}}_{\displaystyle
                \cE \;=\; I(Z_y;\, Y \mid X_s) \;\geq\; 0}.
    \end{equation}
    Since mutual information is non-negative, $\cE \geq 0$.
    Substituting the bound from \cref{eqn:res_concept} into $\sH{Z_y \mid X_s}$:
    \[
        \cE \;\leq\; \sH{Z_y \mid X_s} \;\leq\; \sH{g_c(Z_x) \mid Z_x} + \sH{Z_x \mid X_s}.
    \]
    As model quality improves, the LLM learns $g_c$ more accurately, reducing $\sH{g_c(Z_x) \mid Z_x}$ and hence $\cE$. 

    Let $X^{(2)}_s = (X^{(1)}_s,\, X''_s)$ be an extended prompt that is strictly more informative than $X^{(1)}_s$.  By the data processing inequality (conditioning on more information cannot increase entropy):
    \begin{equation}
        \label{eq:non_increase}
        \sH{Y \mid X^{(2)}_s}
        = \sH{Y \mid X^{(1)}_s,\, X''_s}
        \;\leq\; \sH{Y \mid X^{(1)}_s}.
    \end{equation}

    By \cref{eq:markov_decomp}, $\sH{Y \mid X_s} = \mathbb{E}_{Z_x \mid X_s}[\sH{Y \mid Z_x}]$. This is the expected value of the function $h(z) = \sH{Y \mid Z_x = z}$ under the distribution $P(Z_x \mid X_s)$. By \cref{prop:concept_entropy}, moving from $X^{(1)}_s$ to $X^{(2)}_s$ strictly reduces $\sH{Z_x \mid X_s}$, so $P(Z_x \mid X^{(2)}_s)$ is a strictly more concentrated distribution over $\Theta_x$ than $P(Z_x \mid X^{(1)}_s)$.
    Concretely, the probability mass shifts: there exist concept values $z \neq z'$ such that $P(Z_x = z \mid X^{(1)}_s) > 0$ and $P(Z_x = z \mid X^{(2)}_s) \neq P(Z_x = z \mid X^{(1)}_s)$.

    If the function $h(z) = \sH{Y \mid Z_x = z}$ is non-constant on the support of $P(Z_x \mid X^{(1)}_s)$, equivalently, $I(Y;\, Z_x) > 0$, meaning the response is non-trivially related to the prompt concept, then the redistribution of mass strictly changes the expectation in \cref{eq:markov_decomp}, giving
    \begin{equation}
        \label{eq:strict_decrease}
        \sH{Y \mid X^{(2)}_s} \;<\; \sH{Y \mid X^{(1)}_s}.
    \end{equation}
    Since $I(Y;\, Z_x) > 0$ holds in any non-degenerate PRC setting (a prompt with concept $Z_x$ constrains the response $Y$ differently from a prompt without it), the strict decrease \cref{eq:strict_decrease} holds as the increase in the prompt's informativeness. 

    When $X_s$ fully specifies the concept ($X_s \in \cX_{\theta_x}$), by \cref{lem:concept} $\sH{Z_x \mid X_s} = 0$.  We show $\sH{Z_y \mid X_s} = \sH{Z_y \mid Z_x}$ by a two-sided sandwich:
    \begin{align*}
        \sH{Z_y \mid X_s}
            &\geq \sH{Z_y \mid X_s,\, Z_x}
             = \sH{Z_y \mid Z_x}
             && \text{(conditioning on $Z_x$ can only help)} \\
        \sH{Z_y \mid X_s}
            &\leq \sH{Z_x \mid X_s} + \sH{Z_y \mid Z_x}
             = 0 + \sH{Z_y \mid Z_x}
             && \text{(from \cref{eqn:res_concept} with $\sH{Z_x \mid X_s} = 0$)}
    \end{align*}
    Hence $\sH{Z_y \mid X_s} = \sH{Z_y \mid Z_x} = \sH{g_c(Z_x) \mid Z_x}$.
    Substituting into \cref{eq:exact_decomp}:
    \[
        \sH{Y \mid X_s} = \sH{Y \mid Z_y} + \cE,
        \quad \cE \leq \sH{g_c(Z_x) \mid Z_x}.
    \]
    Under \cref{assum:all} (perfect LLM), $g_c$ is exactly learned.
    For tasks with a deterministic ground truth, $g_c$ is deterministic, so $\sH{g_c(Z_x) \mid Z_x} = 0$ and hence $\cE \to 0$. The residual uncertainty $\sH{Y \mid Z_y}$ is decomposed by the chain rule:
    \[
        \sH{Y \mid Z_y}
            = \sH{Y_\tau \mid Z_y} + \underbrace{\sH{Y \mid Y_\tau,\, Z_y}}_{\displaystyle =:\; \sH{Y_r \mid Z_y}},
    \]
    where $Y_\tau = \tau(Y)$ is the semantic class of the response (e.g., which answer choice is selected) and $Y_r$ denotes the surface-form variation within that class. $\sH{Y_\tau \mid Z_y}$ is the irreducible aleatoric task uncertainty; $\sH{Y_r \mid Z_y}$ is the semantic redundancy.
\end{proof}

    \subsection{Discussion about Response Uncertainty Decomposition}
    \label{asec:response_decomposition}

In this section, we examine the decomposition of response uncertainty. As discussed in \cref{sec:theoretical_results}, each component of uncertainty captures a distinct source of variation:
\begin{itemize}
    \item $\sH{Z_x|X_s}$ reflects uncertainty due to prompt underspecification, specifically, the model's inability to infer the intended concept from the prompt, either because of incomplete information or limitations in its inherent knowledge.

    \item $\sH{g_c(Z_x)|Z_x}$ captures uncertainty arising from model quality, i.e., the model's ability to generate the appropriate high-level abstractions (response concepts) based on its understanding of the instruction. This uncertainty depends on how well the model interprets the prompt and produces a suitable response, overall depending on the model's quality.

    \item $\sH{Y_\tau|Z_y}$ represents task-related variability, where randomness is inherent to the task itself (e.g., predicting the outcome of a fair coin toss or rolling a die).

    \item $\sH{Y_r|Z_y}$ accounts for semantic redundancy, reflecting the existence of multiple ways to express the same underlying response concept (i.e., different sequences that are semantically equivalent sentences).
\end{itemize}

Together, these components provide a structured view of the different sources of uncertainty in language model outputs.

\subsection{Relationship between Prompts and Response}
Having established the PRC model, we formalize it mathematically to analyze the response uncertainty. As stated before, we now analyze the probability of a \emph{response-level}, complete response $y$ given the initial prompt $x$.
Following prior work~\citep{xie2021explanation,hahn2023theory,zhang2023and,wang2024large}, which assumes that LLMs implicitly perform Bayesian inference, we also adopt this perspective. 
Specifically, we assume that the conditional distribution $p(y|x)$ representing the posterior predictive distribution is marginalized over the latent prompt and response concepts.
Since $Z_x$ and $Z_y$ are discrete random variables (see \cref{asec:proofs}), the posterior predictive is marginalized by summation:
\begin{align*}
    p(y \mid x)
        &= \sum_{\theta_y \in \Theta_y} p(y \mid \theta_y, x)\, p(\theta_y \mid x) = \sum_{\theta_y \in \Theta_y} \sum_{\theta_x \in \Theta_x}
              p(y \mid \theta_y, x)\, p(\theta_y \mid \theta_x, x)\, p(\theta_x \mid x).
\end{align*}
The first equality follows from conditioning on the response concept.
The term $p(y \mid \theta_y, x)$ is the probability of response $y$ given concept $\theta_y$.
The second equality uses:
\[
    p(\theta_y \mid x) = \sum_{\theta_x \in \Theta_x} p(\theta_y \mid \theta_x, x)\, p(\theta_x \mid x),
\]
marginalizing over the discrete prompt-concept space $\Theta_x$.
If $p(\theta_c|x)$ (where $c = \{x,y\}$) concentrates on a specific concept with a more informative prompt, the LLM learns effectively via marginalization. 
More concretely, our PRC model assumes that the LLM achieves this by extracting task-relevant attributes from the prompt $x$ and using its inherent knowledge acquired during pretraining. 
  

\section{Additional Experiment Details}
\label{asec:experiment_results}
We first give additional experiments on other datasets such as ARC~\citep{allenai:arc}, OpenBookQA~\citep{OpenBookQA2018}, and RACE~\citep{lai-etal-2017-race}, comparing the accuracy and empirical entropy of generated responses across different LLMs under similar conditions as described in \cref{subsec:response_quality}. We then show the ablation studies in \cref{subsec:ablation}, demonstrating how response uncertainty varies with different types of noisy prompts. Finally, we describe the details of our mHealth intervention simulation experiments.

\begin{figure}[!ht]
    \vspace{-5mm}
	\centering
	\subfloat[Sentence-level removal (RACE dataset)]{\label{fig:sentence_race}
		\includegraphics[width=0.49\linewidth]{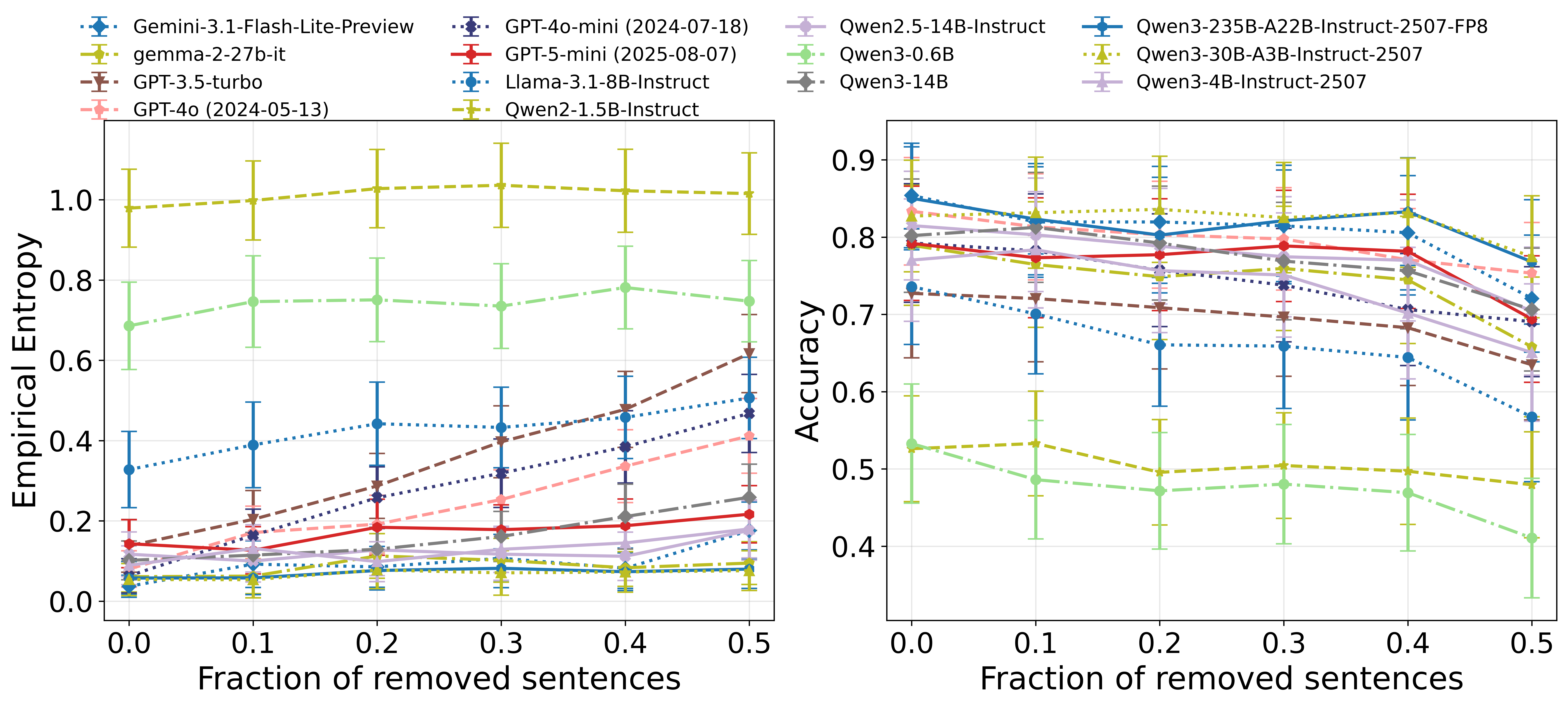}}
    ~~\subfloat[Attribute-level removal (RACE dataset)]{\label{fig:attribute_race}
		\includegraphics[width=0.49\linewidth]{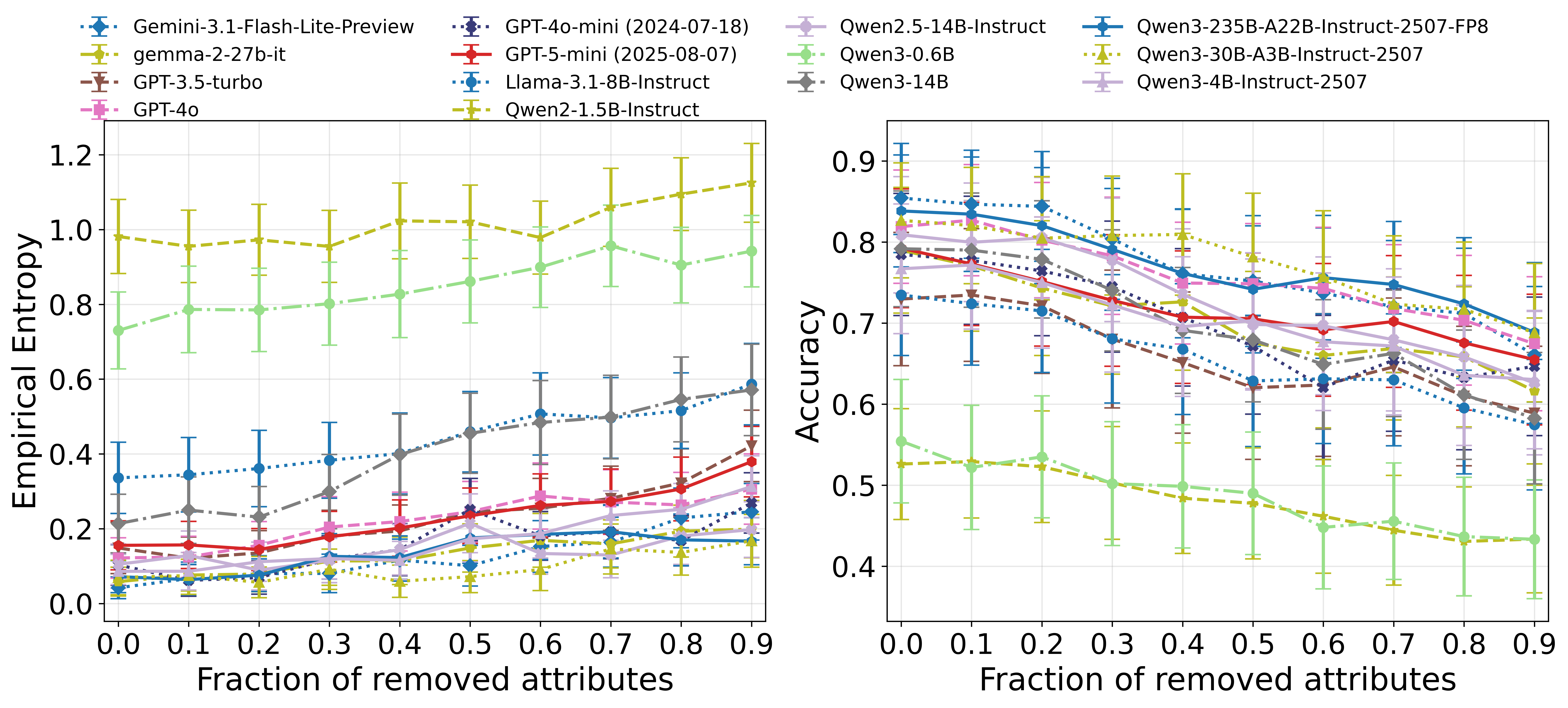}}
	\caption{ 
       Accuracy and entropy for the RACE dataset after sentence- and attribute-level removal. 
       There is a strong negative correlation between accuracy and uncertainty, with less accurate models generally showing greater uncertainty in their responses. 
    }
	\label{fig:race1}
\end{figure}

\begin{figure}[!ht]
	\centering
	\subfloat[Entropy (ARC)]{\label{fig:empirical_entropy_arc}
		\includegraphics[width=0.24\linewidth]{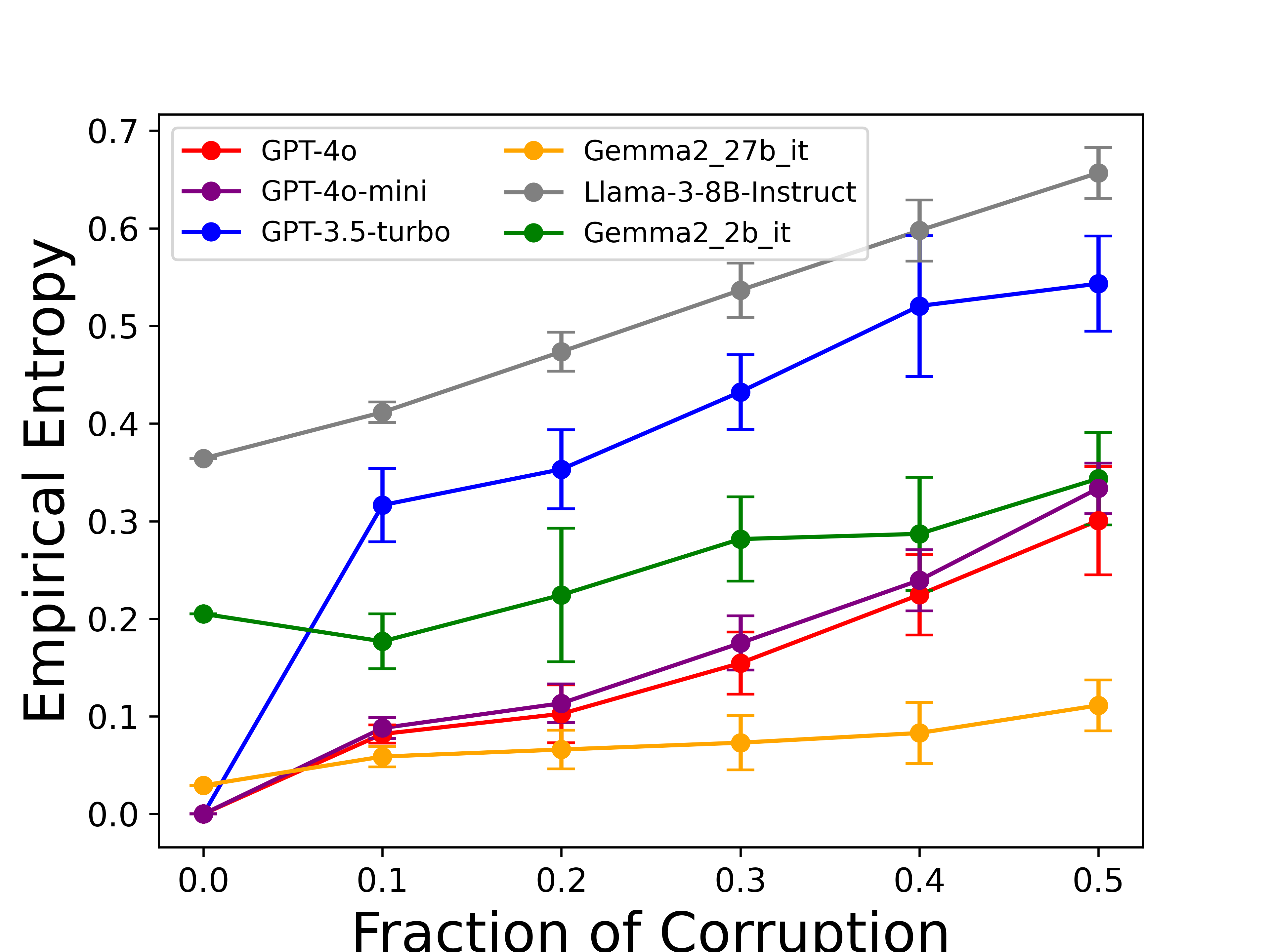}} 
    \subfloat[Accuracy (ARC)]{\label{fig:accuracy_arc}
		\includegraphics[width=0.24\linewidth]{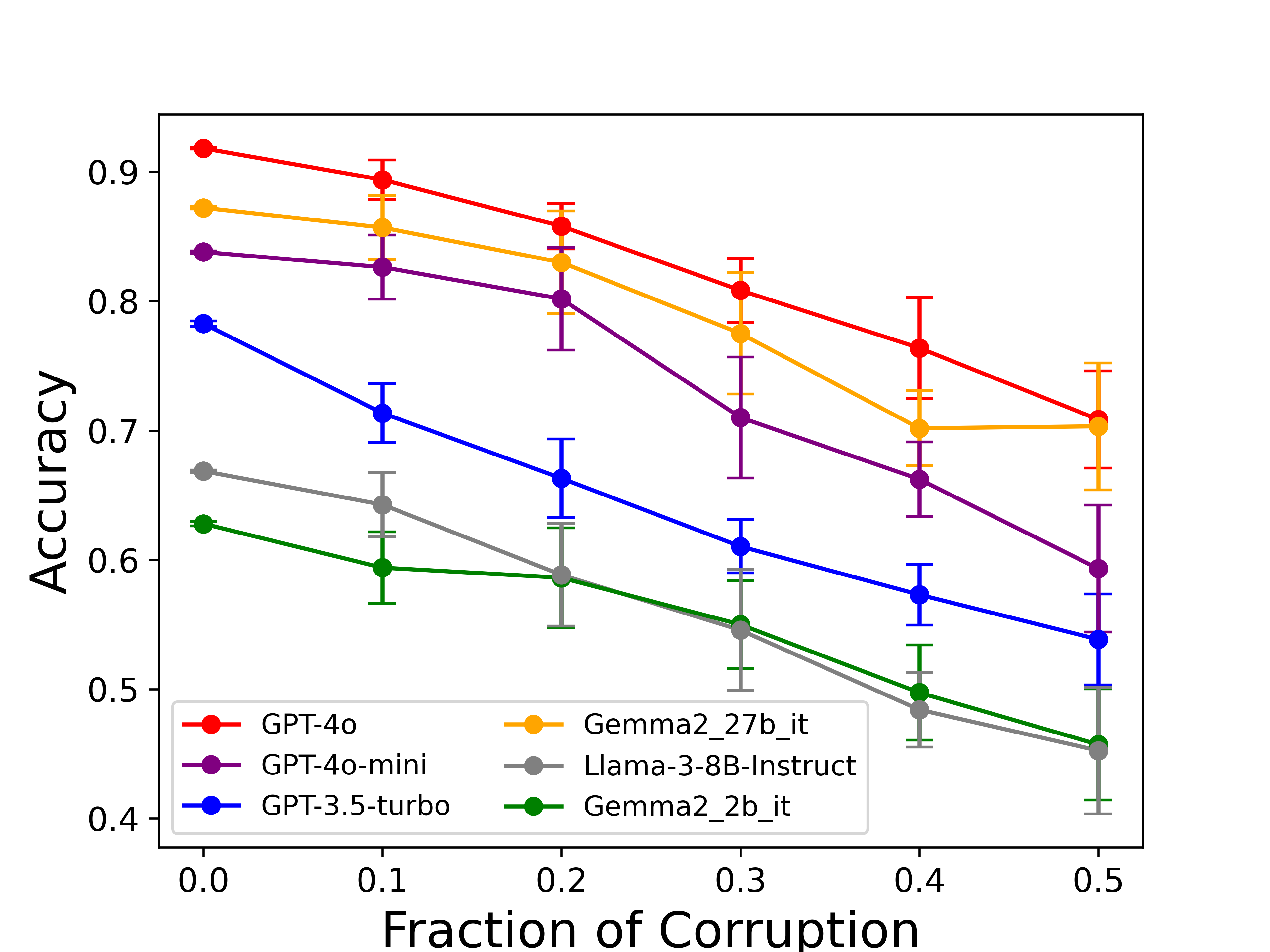}}
    \subfloat[Entropy (OpenBookQA)]{\label{fig:empirical_entropy_openbookqa}
		\includegraphics[width=0.24\linewidth]{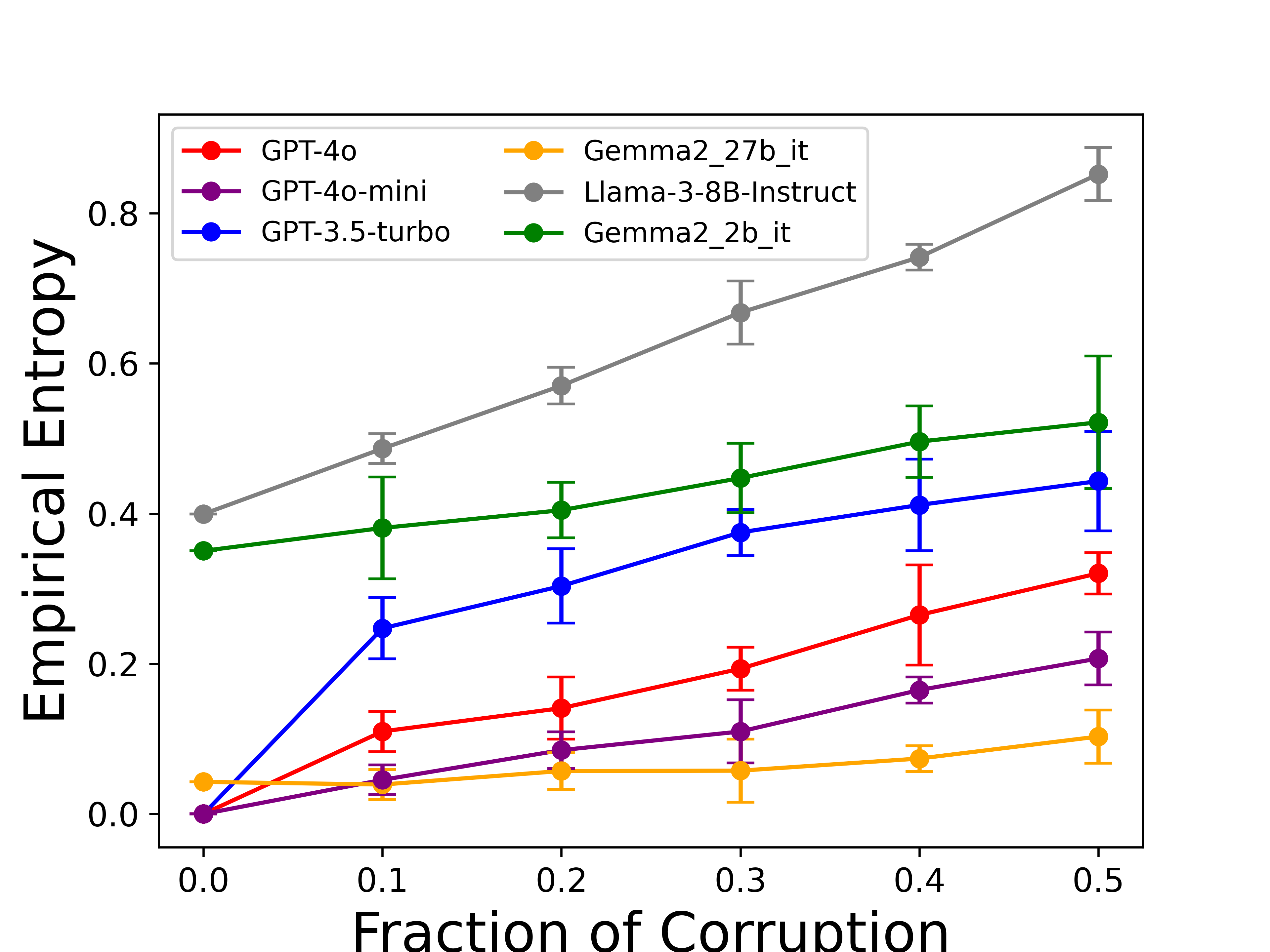}}
    \subfloat[Accuracy (OpenBookQA)]{\label{fig:accuracy_openbookqa}
		\includegraphics[width=0.24\linewidth]{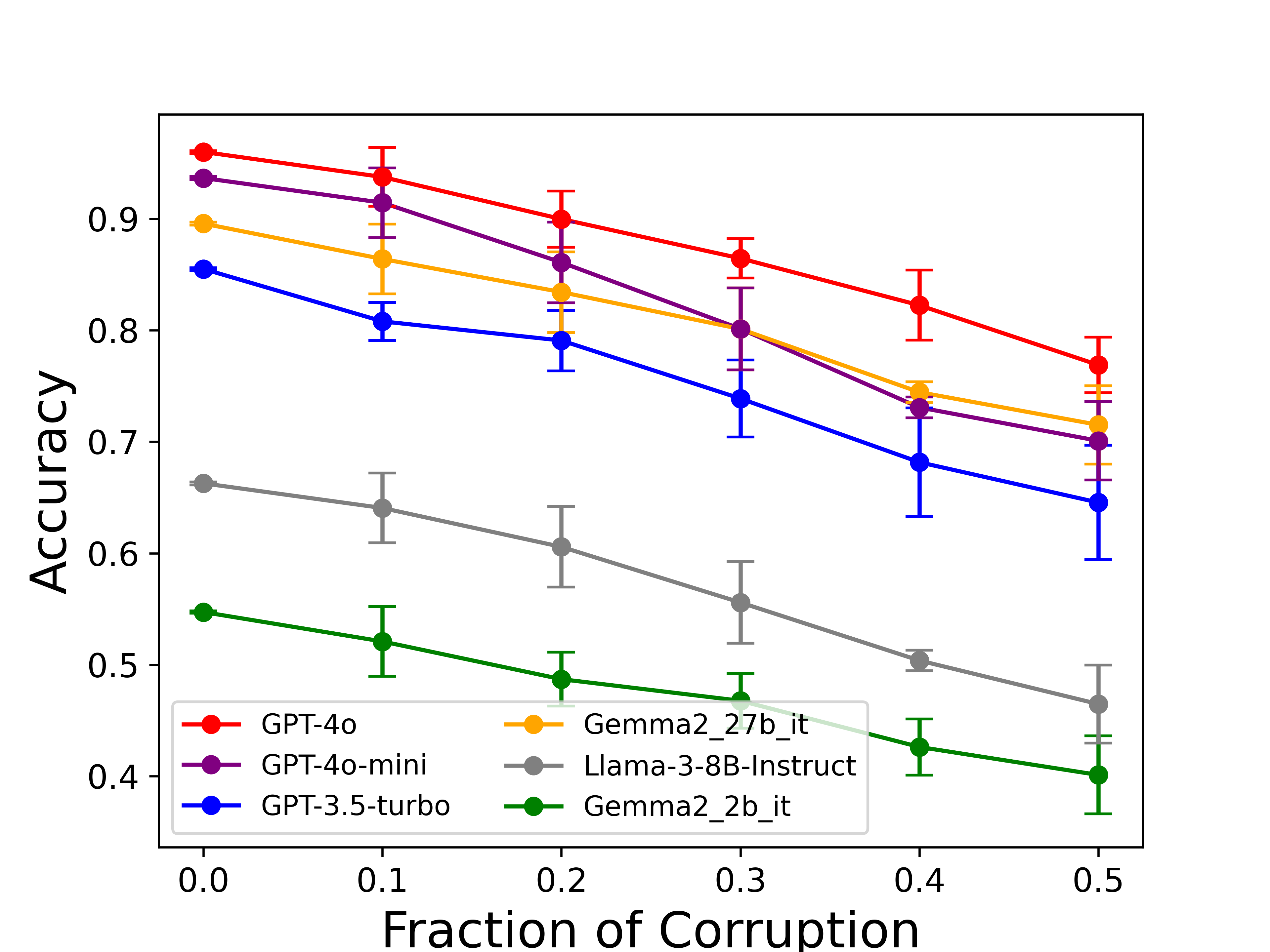}}
	\caption{ 
       \textbf Accuracy and entropy for MCQs datasets. 
       There is a clear and strong negative correlation between accuracy and uncertainty, with less accurate models generally showing greater uncertainty in their responses. 
       For the ARC dataset, the Gemma2 2B model with full context not only surpassed GPT-3.5 but also performed on par with GPT-4o-mini with only half the context.
    }
	\label{fig:ablations75}
    \vspace{-5mm}
\end{figure}

\begin{figure}[!ht]
	\centering
    \subfloat[Sentence-level removal]{\label{fig:sentence_race_comp}
		 \includegraphics[width=0.22\linewidth]{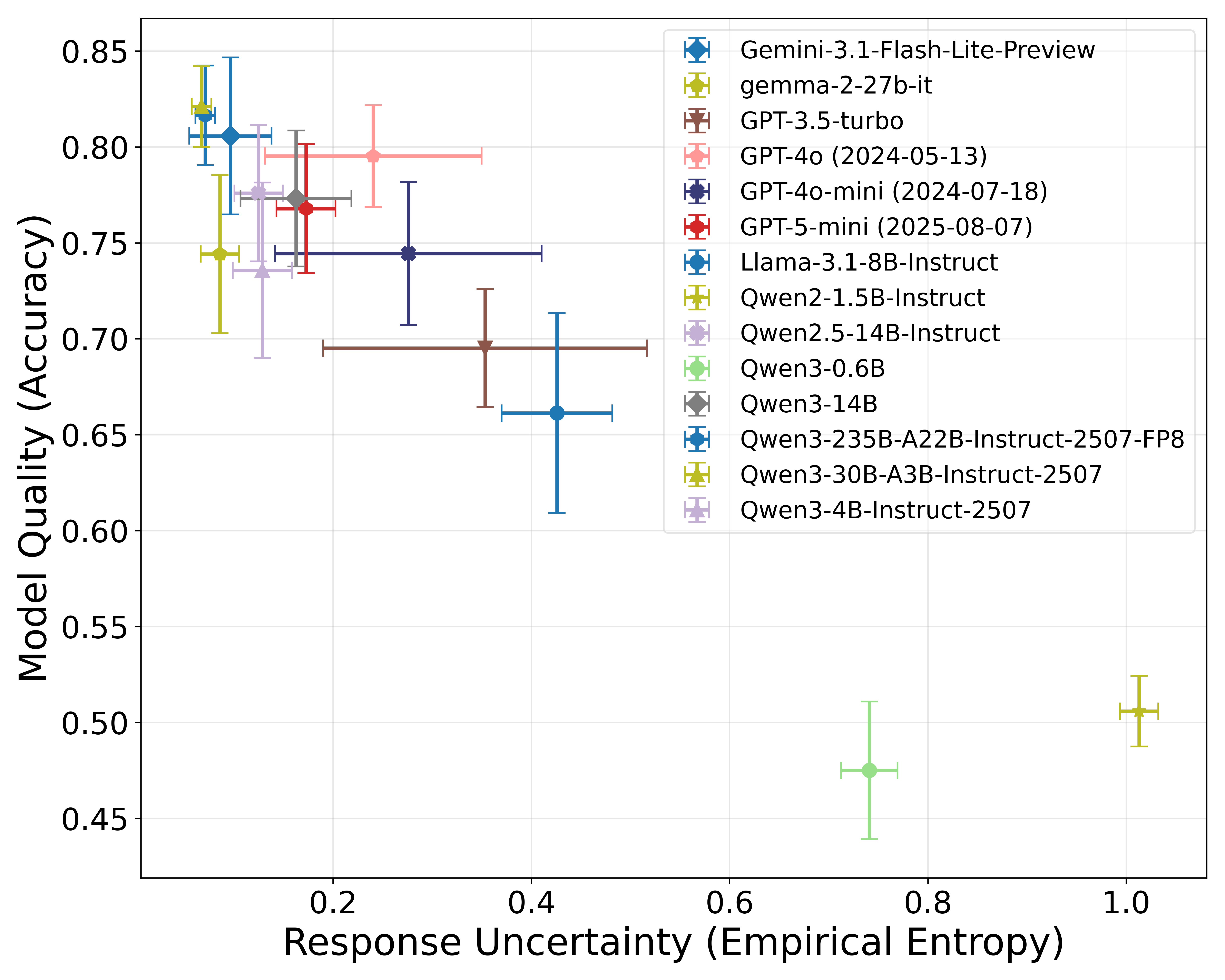}}
    \subfloat[Attribute-level removal]{\label{fig:attribute_race_comp}
		 \includegraphics[width=0.22\linewidth]{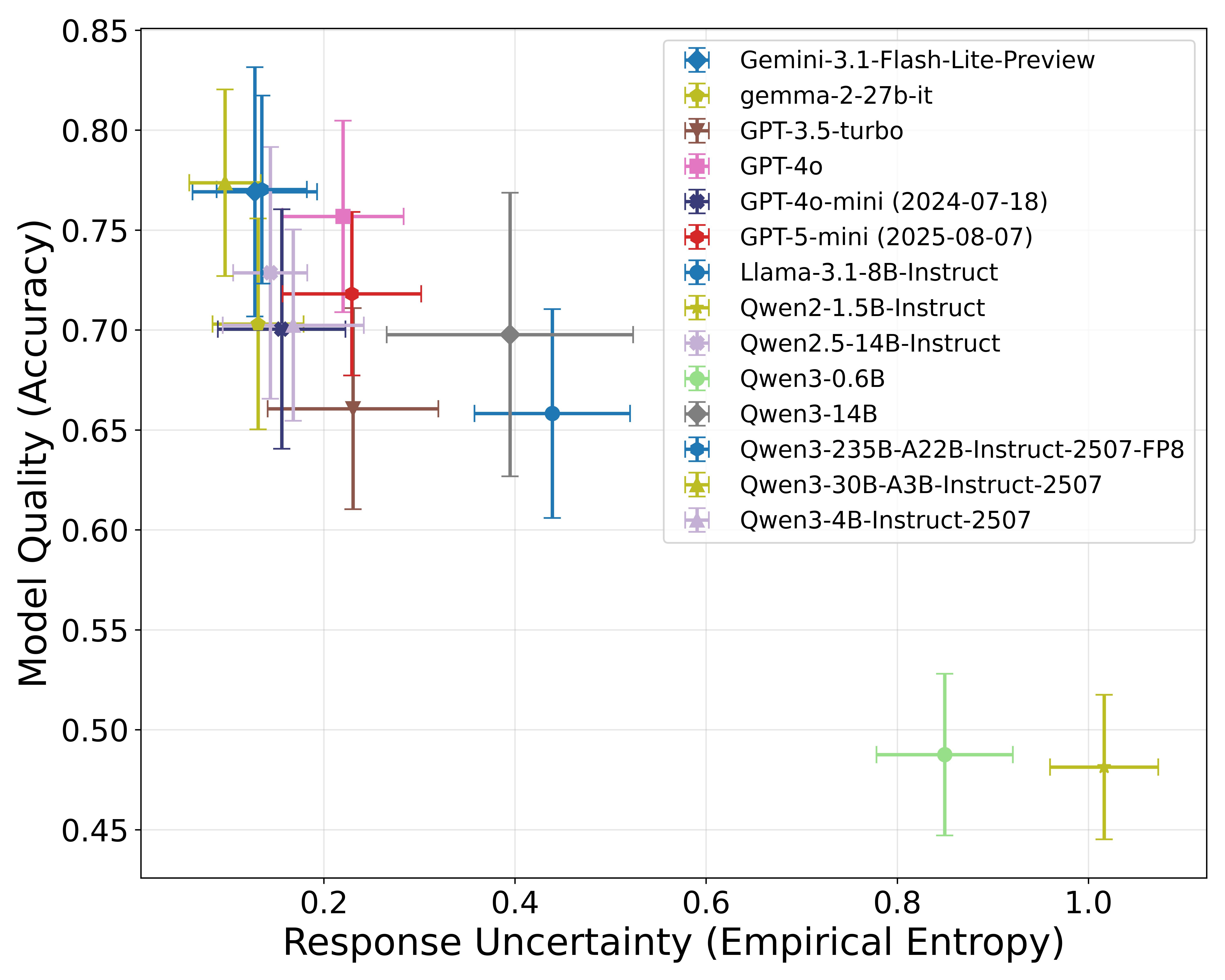}}
    \subfloat[Token-level removal]{\label{fig:epsilon}
		 \includegraphics[width=0.23\linewidth]{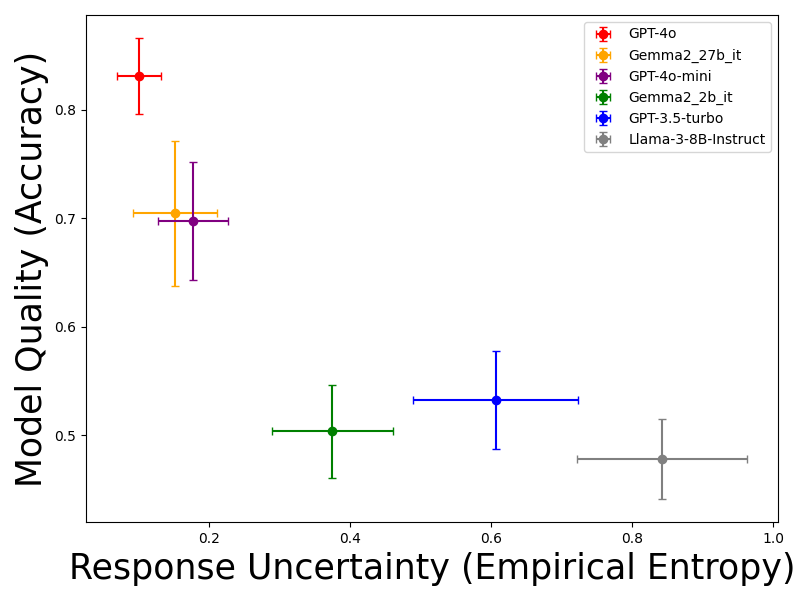}}
	\caption{\textbf{Fig. (a) and Fig. (b):} Model response quality (i.e., accuracy) vs. uncertainty (i.e., empirical entropy) for RACE dataset after sentence- and attribute-level removal. \textbf{Fig. (c):} Model response quality (i.e., accuracy) vs uncertainty (i.e., empirical entropy) for different models. Averaged across Medical Meadow Medqa, ARC, and OpenbookQA, and all corruption levels. 
	}
	\label{fig:openbookqa}
    \vspace{-5mm}
\end{figure}

We also selected $100$ MCQs from ARC~\citep{allenai:arc} and OpenBookQA~\citep{OpenBookQA2018} datasets. In \cref{fig:accuracy_arc,fig:accuracy_openbookqa}, we plot the accuracy for the responses of three open-source and three black-box LLMs on the same set of MCQs. As the fraction of masked tokens increases in the prompt, the accuracy monotonically decreases for all tested models. 
For each random seed, we also plot the empirical conditional entropy $\sH{Y|X}$ of the response for the given questions\footnote{The distribution of the questions used $p(x)$ is assumed uniform. With no access to the prior of $p(y|x)$, we use the form $\sH{Y|X} = - \sum_{x} {p}(x) \sum_{y} \hat{p}(y|x) \log \hat{p}(y|x)$ where $\hat{p}(y|x)$ is obtained from the empirical distribution and ${p}(x)=\frac{1}{100}$ for all $x$ in the setting. The conditional entropy is a good measure for MCQ settings, as the model's effective response is just one choice.} (\cref{fig:empirical_entropy_arc,fig:empirical_entropy_openbookqa}) as an indicator of response uncertainty. As corruption becomes more severe, we observe that the response uncertainty increases for all models (increases monotonically for larger LLMs), indicating a clear negative correlation between $\sH{Y|X}$ and the response accuracy.
This result corroborates our hypothesis: more relevant information leads to both a reduction in response uncertainty and an improvement in its quality. 
Furthermore, as shown in \cref{fig:epsilon}, models with better accuracy tend to have lower empirical entropy. This corroborates our {characterization} of $\cE$ in \cref{thm:response_uncertainty}. 

\subsection{The Relative Importance of Different Attributes}
\begin{figure}[!ht]
    \vspace{-5mm}
	\centering
	\subfloat[Prompt 4 LOO Accuracy]{\label{fig:acc4}
	   \includegraphics[width=0.24\linewidth]{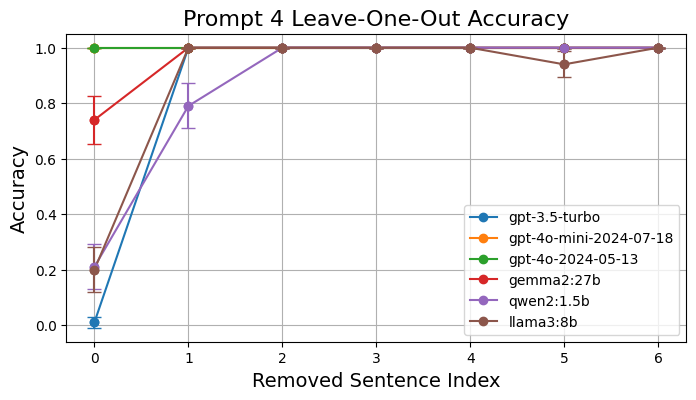}}
    \subfloat[Prompt 4 LOO Entropy]{\label{fig:uncertainty4}
		\includegraphics[width=0.24\linewidth]{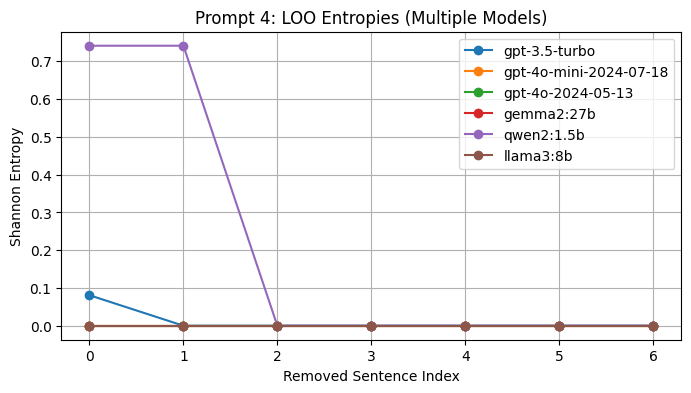}}
    \subfloat[Prompt 7 LOO Accuracy]{\label{fig:acc7}
		\includegraphics[width=0.24\linewidth]{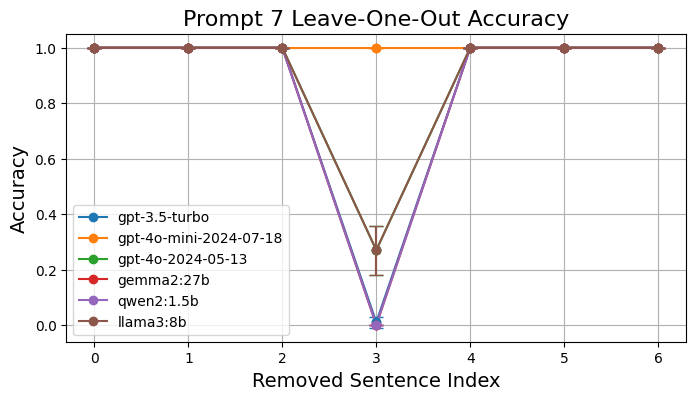}}
      \subfloat[Prompt 7 LOO Entropy]{\label{fig:uncertainty7}
		\includegraphics[width=0.24\linewidth]{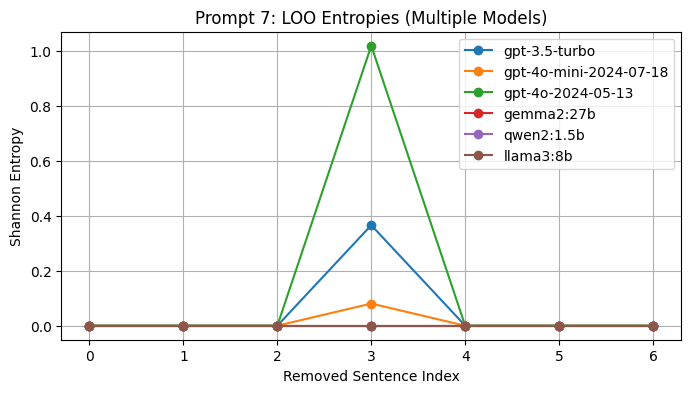}}
	\caption{ 
       (a) Prompt 4 Leave-One-Out Accuracy. (b): Prompt 4 Leave-One-Out Empirical Entropy. (c): Prompt 7 Leave-One-Out Accuracy. (d): Prompt 7 Leave-One-Out Empirical Entropy. For all plots, the color not visible has a value of 0. It can be observed that there is a clear correlation between the choice of sentence removal and the response quality/uncertainty, which indicates certain attributes are commonly important across multiple models.
    }
	\label{fig:LLO}
    \vspace{-2mm}
\end{figure}

 We investigate to what extent different attributes contribute to model response quality and uncertainty. We choose 10 questions from the RACE dataset \citep{lai-etal-2017-race} with moderate context length, assume each context as one concept and the sentences it comprises as its attributes, and use the leave-one-out method to remove one sentence from its context, for each case generate 100 response samples and observe its impact on the model response. As shown in \cref{fig:LLO}, we observed that in some cases, there is a strong correlation between the choice of the removal of the sentence and the response quality and uncertainty across different models. This suggests that to some degree, there is a consensus among the LLMs about the importance of certain attributes in affecting the model's ability to find the right prompt and response concept.

\subsection{Response Quality vs. Model Precision}
\begin{figure}[!ht]
    \vspace{-5mm}
	\centering
    \subfloat[Sentence-level removal (MedQA dataset)]{\label{fig:sentence_medqa_precision}
		\includegraphics[width=0.49\linewidth]{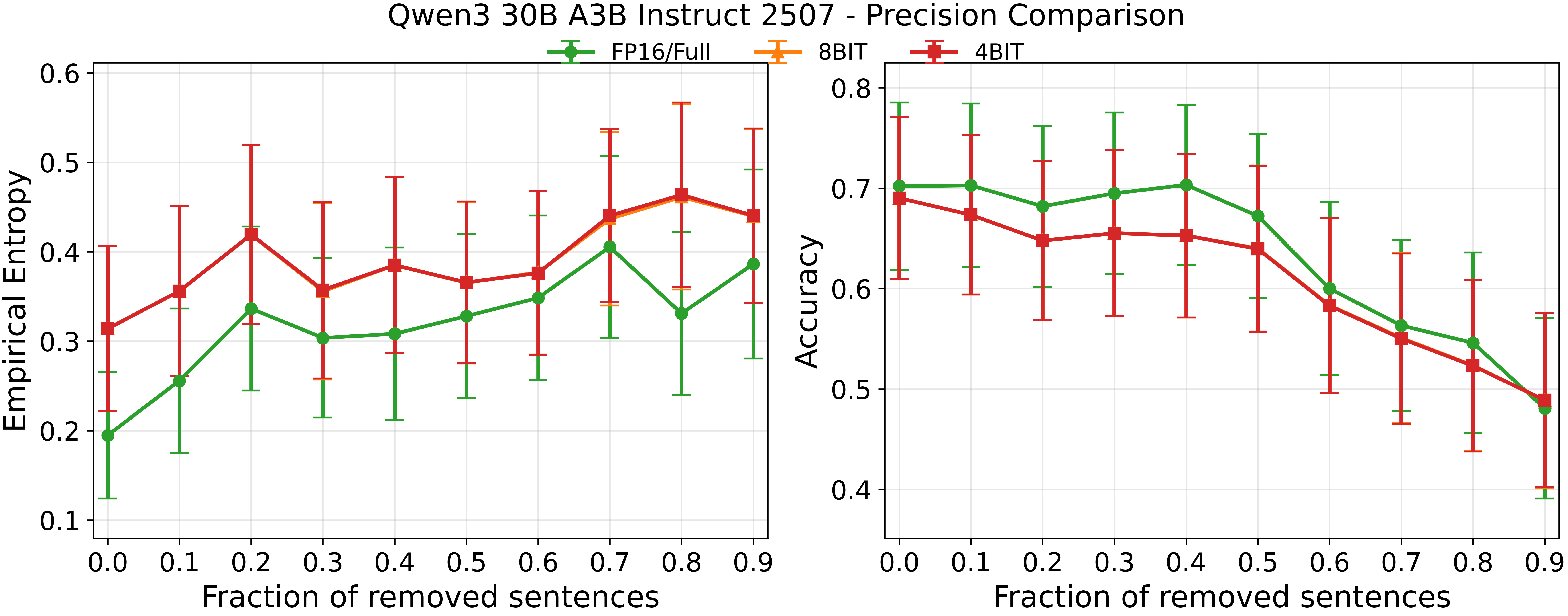}}
    \subfloat[Attribute-level removal (MedQA dataset)]{\label{fig:attribute_medqa_precision}
		\includegraphics[width=0.49\linewidth]{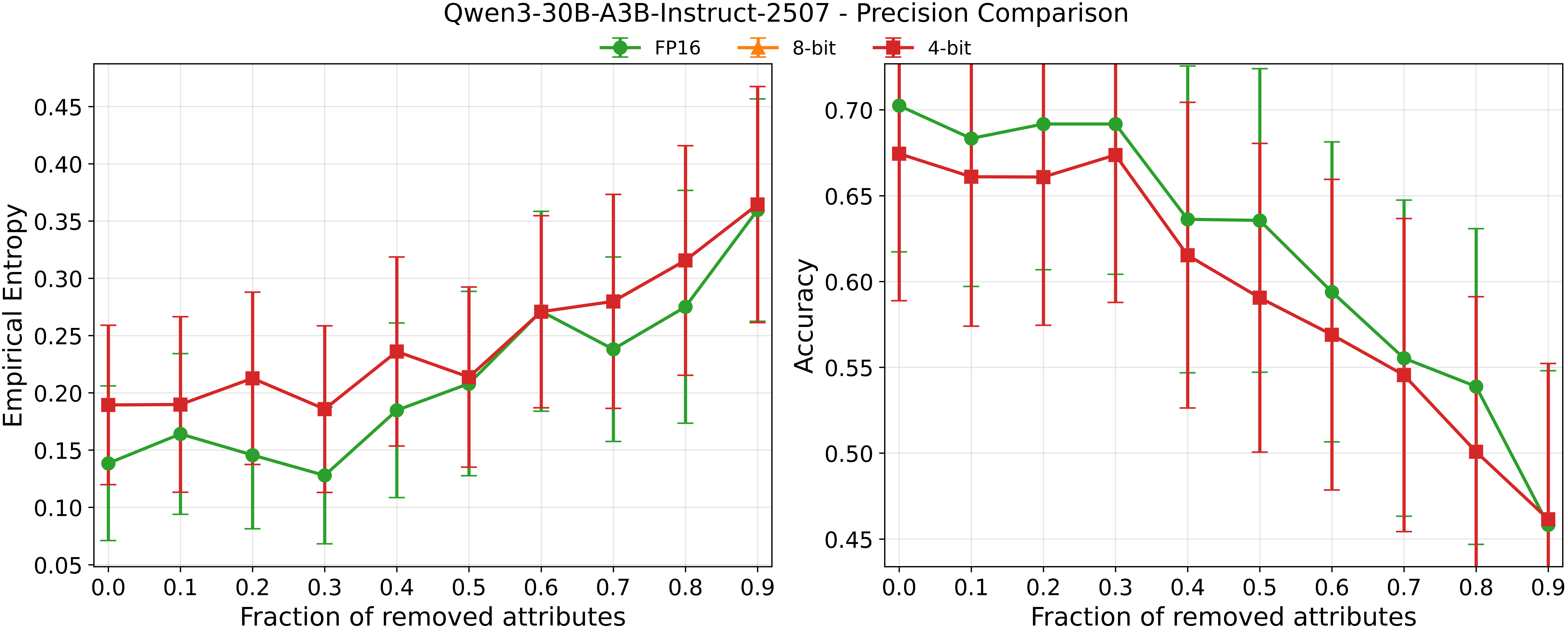}}\\
	\subfloat[Sentence-level removal (RACE dataset)]{\label{fig:sentence_race_precision}
		\includegraphics[width=0.49\linewidth]{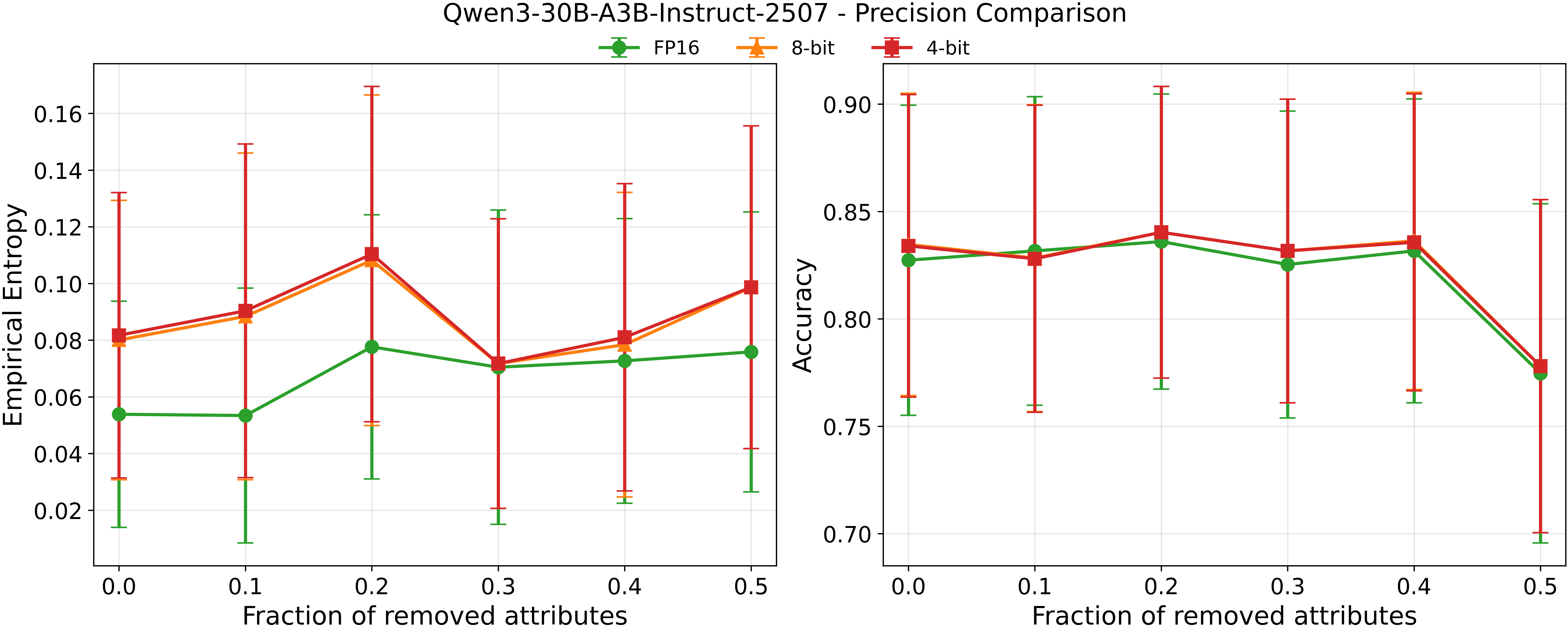}}
    \subfloat[Attribute-level removal (RACE dataset)]{\label{fig:attribute_race_precision}
		\includegraphics[width=0.49\linewidth]{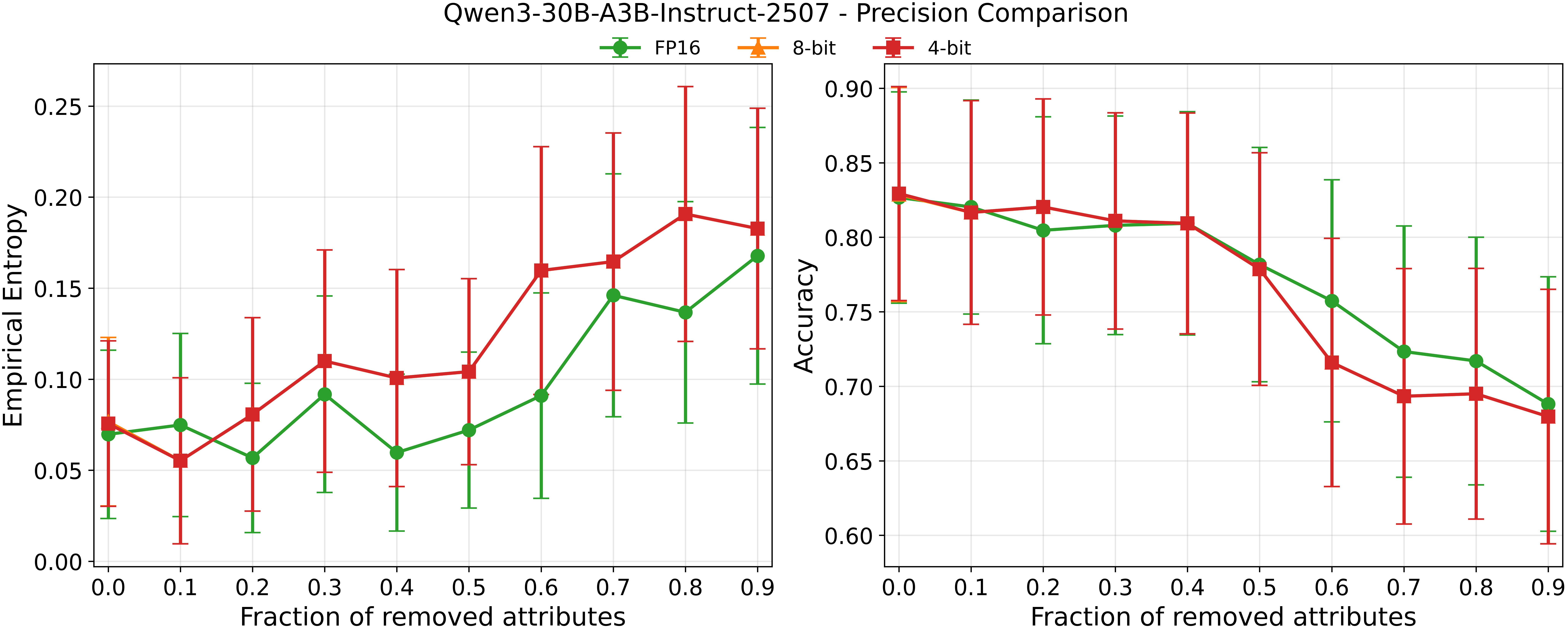}}
	\caption{ 
        Accuracy and entropy for MCQs datasets vs. model precision. There is a clear and strong negative correlation between model precision and response quality, with less precise models generally showing greater uncertainty and lower accuracy in their responses. 
    }
	\label{fig:precision}
    \vspace{-3mm}
\end{figure}

\cref{fig:precision} illustrates the relationship between model precision and both accuracy and response uncertainty across multiple MCQ datasets (MedQA and RACE). We observe a clear and strong negative correlation between model precision and response quality: less precise models consistently exhibit higher empirical conditional entropy and lower accuracy, while more precise models produce more confident and accurate responses. This trend holds across datasets, suggesting that increased model precision is closely associated with reduced response uncertainty and improved decision quality in multiple-choice question answering.

\subsection{Isolating the Sources of Response Uncertainty}
\label{app:isolating-sources}
The PRC model decomposes response uncertainty into four sources: prompt underspecification $\mathrm{H}(Z_x \mid X_s)$, model quality $\mathrm{H}(g_c(Z_x) \mid Z_x)$, task variability $\mathrm{H}(Y_\tau \mid Z_y)$, and semantic redundancy $\mathrm{H}(Y_r \mid Z_y)$. Task variability is eliminated by design, as all datasets used (MedQA, ARC, OpenBookQA, RACE) are MCQ benchmarks with deterministic ground truth. Semantic redundancy is also largely suppressed in the MCQ setting: because the response space collapses to the discrete tokens \{A, B, C, D\}, any two semantically equivalent phrasings of the same response concept map to the same symbol, so $\mathrm{H}(Y_r \mid Z_y) \approx 0$ in expectation. Equivalently, the MCQ format decouples $g_y$ from the upstream composition $g_x \circ g_c$, allowing the remaining experiments to probe $g_x$ and $g_c$ in isolation. This leaves two substantive sources of uncertainty to probe empirically: prompt underspecification and the quality of the internal mappings $g_x$ and $g_c$. Table~\ref{tab:appC-overview} summarizes the three complementary experiments used to isolate each term.

\begin{table}[!ht]
\centering
\small
\setlength{\tabcolsep}{4pt}
\begin{tabular}{lllc}
\toprule
\textbf{Experiment} & \textbf{Held fixed} & \textbf{Varied} & \textbf{Isolates} \\
\midrule
Information removal (Sec.~\ref{sec:experiments}) & Model, decoding, task & Prompt informativeness & $\mathrm{H}(Z_x \mid X_s)$ \\
LLM-judge prompt understanding (Table~\ref{tab:gx-proxy}) & Prompt, task & Model & $g_x$ quality \\
Matched-$g_x$ MCQ sampling (Tables~\ref{tab:gc-isolation}, \ref{tab:gc-perquestion}) & Prompt, $Z_x$ (two-stage filter) & Model & $g_c$ quality \\
\bottomrule
\end{tabular}
\caption{Overview of the three experiments in Appendix~\ref{app:isolating-sources} and the PRC decomposition term each isolates.}
\label{tab:appC-overview}
\vspace{-3mm}
\end{table}

\paragraph{Isolating prompt underspecification $\mathrm{H}(Z_x \mid X_s)$.}
Prompt underspecification refers specifically to missing task-relevant attributes in the prompt. Accordingly, we require an experimental knob that reduces prompt informativeness while keeping the task instance, model, and decoding fixed. To achieve this, we proceed as follows (more details are in Section~\ref{sec:experiments}): We progressively remove information from the prompt using sentence-level, attribute-level, and token-level removal. Among these, attribute-level removal is emphasized because it targets semantically meaningful facts (attributes) rather than deleting entire sentences, aligning more closely with the notion of concept attributes. As the removal fraction increases, we expand the set of removed elements while keeping previously removed elements fixed, and repeat the process across multiple random seeds. This ensures that informativeness decreases monotonically and avoids confounding effects from random selection. For the MCQ datasets, we sample 100 responses per question under fixed decoding and compute empirical response entropy over the discrete A/B/C/D outcomes. Under this setup, changes in response entropy as we vary the removal fraction can be attributed to changes in $\mathrm{H}(Z_x \mid X_s)$, with the model and decoding held fixed. As shown in Figures~\ref{fig:medqa}, \ref{fig:medqa2}, and \ref{fig:race1}, response entropy increases monotonically as the removal fraction grows for all models tested, while accuracy correspondingly decreases. This directly supports Proposition~\ref{prop:concept_entropy} and Theorem~\ref{thm:response_uncertainty}: reducing prompt informativeness increases $\mathrm{H}(Z_x \mid X_s)$, which propagates to the response-level entropy.

\paragraph{Direct proxy for $g_x$ (prompt understanding).}
The information-removal knob above isolates prompt underspecification but does not directly probe the quality of $g_x$ itself: $g_x$ quality concerns how reliably a model maps a fully informative prompt to the correct prompt concept, with $X_s$ held fixed. To compare $g_x$ quality across models, we use a complementary proxy: on 100 questions from the RACE dataset, each model generates a short description of what the question is asking, which is then scored on a 1--10 scale by a fixed LLM judge (Claude 3.7 Sonnet). Since this proxy depends only on the model's ability to articulate the task from a given prompt, it reflects $g_x$ quality while being largely insensitive to $g_c$.

\begin{table}[!ht]
\centering
\small
\setlength{\tabcolsep}{6pt}
\begin{tabular}{lccc}
\toprule
\textbf{Model} & \textbf{Mean $\pm$ Std} & \textbf{Min} & \textbf{Max} \\
\midrule
gpt-4o-2024-05-13      & $8.58 \pm 0.84$ & 6 & 9 \\
gpt-4o-mini-2024-07-18 & $7.73 \pm 0.68$ & 5 & 9 \\
llama3\_8b             & $6.30 \pm 1.52$ & 3 & 9 \\
gemma2\_27b            & $5.52 \pm 1.42$ & 2 & 8 \\
gpt-3.5-turbo          & $3.97 \pm 1.40$ & 2 & 8 \\
qwen2\_1.5b            & $3.52 \pm 1.76$ & 1 & 9 \\
\bottomrule
\end{tabular}
\caption{A proxy for prompt understanding ($g_x$) on 100 RACE questions, scored 1--10 by an LLM judge (Claude 3.7 Sonnet). Higher-capability models score higher on average and exhibit tighter distributions, consistent with more reliable $g_x$.}
\label{tab:gx-proxy}
\end{table}

As shown in Table~\ref{tab:gx-proxy}, the judge scores are ordered consistently with overall model capability: frontier models such as GPT-4o score highest, mid-tier open models (Llama3-8B, Gemma2-27B) fall in the middle, and smaller or older models (GPT-3.5-turbo, Qwen2-1.5B) score lowest. The Min/Max and standard deviation columns further reveal that stronger models exhibit \emph{tighter} distributions (GPT-4o: 6--9, $\sigma = 0.84$) while weaker models show wider dispersion (Qwen2-1.5B: 1--9, $\sigma = 1.76$). This indicates that $g_x$ reliability, not only its average quality, improves with model capability: weaker models occasionally match stronger ones on easy items but fail inconsistently on harder ones. This ordering is also consistent with the response-uncertainty rankings observed in our main MCQ experiments, supporting the PRC model's prediction that more capable LLMs learn a more reliable $g_x$.

\paragraph{Isolating $g_c$ by controlling for $g_x$.}
Model quality captures the stochasticity or imperfection in the mapping from prompt concepts to response concepts ($g_c$). To attribute differences in uncertainty to this term, all models must share the same prompt concept $Z_x$ for a given input; otherwise, entropy differences would conflate failures in $g_x$ and $g_c$. We enforce this with a \textbf{two-stage filter} over RACE questions. First, \emph{across-model consistency}: we use an LLM judge (Claude 3.7 Sonnet) to select questions on which all six tested models produce semantically equivalent descriptions of what the question asks, ensuring that the inferred prompt concept is shared across models. Second, \emph{within-model consistency}: for each selected question, we sample each model's prompt description 20 times and retain only questions where every sample is semantically equivalent, ensuring $\mathrm{H}(Z_x \mid X_s) \approx 0$ for each individual model. Together, these two stages hold $Z_x$ fixed both across and within models. We further restrict to RACE because its context contains sufficient information to derive the ground truth, so no attribute relevant to the task is missing from the prompt. For each of the resulting filtered questions, we sample 500 MCQ responses per model under fixed decoding and compute accuracy and empirical response entropy. Under this setting, residual differences in response entropy across models are attributable primarily to differences in $g_c$.

\begin{table}[!ht]
\centering
\small
\setlength{\tabcolsep}{4pt}
\begin{tabular}{lccc}
\toprule
\textbf{Model} & \textbf{Mean Acc.} & \textbf{Mean Ent.} & \makecell{\textbf{Conf.\ wrong}\\\textbf{(Acc=0, Ent=0)}} \\
\midrule
gpt-4o-2024-05-13      & 1.0000 & 0.0000 & 0 \\
gpt-4o-mini-2024-07-18 & 1.0000 & 0.0000 & 0 \\
gemma2\_27b            & 1.0000 & 0.0000 & 0 \\
gpt-3.5-turbo          & 0.9980 & 0.0162 & 0 \\
llama3\_8b             & 0.7824 & 0.5444 & 0 \\
qwen2\_1.5b            & 0.6000 & 0.1776 & 1 \\
\bottomrule
\end{tabular}
\caption{Isolating $g_c$ after controlling prompt understanding via the two-stage filter (5 selected questions, mean over questions, 500 samples per question).}
\label{tab:gc-isolation}
\end{table}

\begin{table}[!ht]
\centering
\small
\setlength{\tabcolsep}{4pt}
\begin{tabular}{lcccccccccc}
\toprule
& \multicolumn{2}{c}{Q4} & \multicolumn{2}{c}{Q8} & \multicolumn{2}{c}{Q46} & \multicolumn{2}{c}{Q61} & \multicolumn{2}{c}{Q63} \\
\cmidrule(lr){2-3}\cmidrule(lr){4-5}\cmidrule(lr){6-7}\cmidrule(lr){8-9}\cmidrule(lr){10-11}
\textbf{Model} & Acc & Ent & Acc & Ent & Acc & Ent & Acc & Ent & Acc & Ent \\
\midrule
gpt-4o-2024-05-13      & 1.000 & 0.000 & 1.000 & 0.000 & 1.000 & 0.000 & 1.000 & 0.000 & 1.000 & 0.000 \\
gpt-4o-mini-2024-07-18 & 1.000 & 0.000 & 1.000 & 0.000 & 1.000 & 0.000 & 1.000 & 0.000 & 1.000 & 0.000 \\
gemma2\_27b            & 1.000 & 0.000 & 1.000 & 0.000 & 1.000 & 0.000 & 1.000 & 0.000 & 1.000 & 0.000 \\
gpt-3.5-turbo          & 0.990 & 0.081 & 1.000 & 0.000 & 1.000 & 0.000 & 1.000 & 0.000 & 1.000 & 0.000 \\
llama3\_8b             & 0.664 & 0.958 & 1.000 & 0.000 & 1.000 & 0.000 & 1.000 & 0.000 & 0.248 & 1.764 \\
qwen2\_1.5b            & 1.000 & 0.000 & 1.000 & 0.000 & \textbf{0.000} & \textbf{0.000} & 1.000 & 0.000 & 0.000 & 0.888 \\
\bottomrule
\end{tabular}
\caption{Per-question accuracy and empirical entropy after the two-stage $g_x$ filter (500 samples per model per question, $T=1$, RACE dataset). Because $Z_x$ is held fixed both within and across models, residual entropy and errors reflect $g_c$ quality alone. The bolded entry (Qwen2-1.5B on Q46) is a ``confidently wrong'' case: accuracy $= 0$ with entropy $= 0$, confirming that low uncertainty does not imply correctness.}
\label{tab:gc-perquestion}
\end{table}

Table~\ref{tab:gc-isolation} shows a clear separation across models even after matched prompt understanding, and Table~\ref{tab:gc-perquestion} breaks this down per question. GPT-4o, GPT-4o-mini, and Gemma2-27B achieve perfect accuracy with zero response entropy on all five questions. GPT-3.5-turbo is near-perfect, with only a small residual entropy (0.081) on Q4 while still answering 99\% of the time correctly, the only case in the table where a model exhibits a sliver of $g_c$ stochasticity without it hurting accuracy, suggesting the model occasionally considers a wrong response concept but usually recovers. Llama3-8B and Qwen2-1.5B, in contrast, exhibit substantially degraded $g_c$.

The per-question data in Table~\ref{tab:gc-perquestion} reveals three qualitatively distinct $g_c$ behaviors once $g_x$ is controlled:
\begin{itemize}
  \item \textbf{Reliable $g_c$} (accuracy $= 1.000$, entropy $\approx 0$): GPT-4o, GPT-4o-mini, Gemma2-27B on all questions, and GPT-3.5-turbo on 4/5 questions. The prompt concept is mapped to the correct response concept essentially deterministically.
  \item \textbf{Stochastic $g_c$} (high entropy, accuracy $< 1$): Llama3-8B on Q4 (Acc $= 0.664$, Ent $= 0.958$) and Q63 (Acc $= 0.248$, Ent $= 1.764$). The model stochastically maps the same prompt concept to multiple response concepts, some correct and some not. This inflates the $\mathrm{H}(g_c(Z_x) \mid Z_x)$ term directly.
  \item \textbf{Deterministic but incorrect $g_c$} (accuracy $= 0.000$, entropy $\approx 0$): Qwen2-1.5B on Q46. The model reliably maps the prompt concept to an incorrect response concept, shifting probability mass onto a wrong token without raising entropy.
\end{itemize}
All three behaviors are consistent with Theorem~\ref{thm:response_uncertainty}: stochastic $g_c$ manifests as elevated $\mathrm{H}(g_c(Z_x) \mid Z_x)$, while deterministic-but-incorrect $g_c$ produces low entropy on a wrong response concept, reinforcing our point in Section~\ref{sec:experiments} that low response uncertainty does not imply correctness. Since $g_x$ is controlled for by construction, this entire spread of behaviors is attributable primarily to differences in $g_c$, consistent with the PRC model's characterization of model quality.

Together, the three experiments above provide empirical evidence for the PRC model's decomposition. Controlled information removal isolates $\mathrm{H}(Z_x \mid X_s)$ and confirms that prompt underspecification drives response entropy. The judge-scored proxy isolates $g_x$ quality across models and shows that more capable models comprehend tasks both more reliably \emph{and} more consistently. The two-stage $g_x$ filter then isolates $g_c$ quality at the question level, revealing distinct failure modes that all map cleanly onto terms in the PRC decomposition. These results support Theorem~\ref{thm:response_uncertainty} and the main claim that prompt informativeness and model quality are the two principal handles for reducing response uncertainty in the MCQ setting.

\subsection{License for Datasets}

Medical Meadow MedQA \citep{jin2021disease}: MIT License; ARC \citep{allenai:arc}: CC BY-SA 4.0; OpenBookQA \citep{OpenBookQA2018}: Apache License 2.0; RACE \citep{lai-etal-2017-race}: non-commercial research purpose only; SQuAD \citep{rajpurkar-etal-2016-squad}: CC BY-SA 4.0.

\subsection{Further Details on the mHealth Intervention Simulation Experiments in  \texorpdfstring{\cref{subsec:Simulation}}{Section 5}}
\label{appendix:Simulation_description}
\begin{figure}[!ht]
  \centering
  \includegraphics[width=0.6\linewidth]{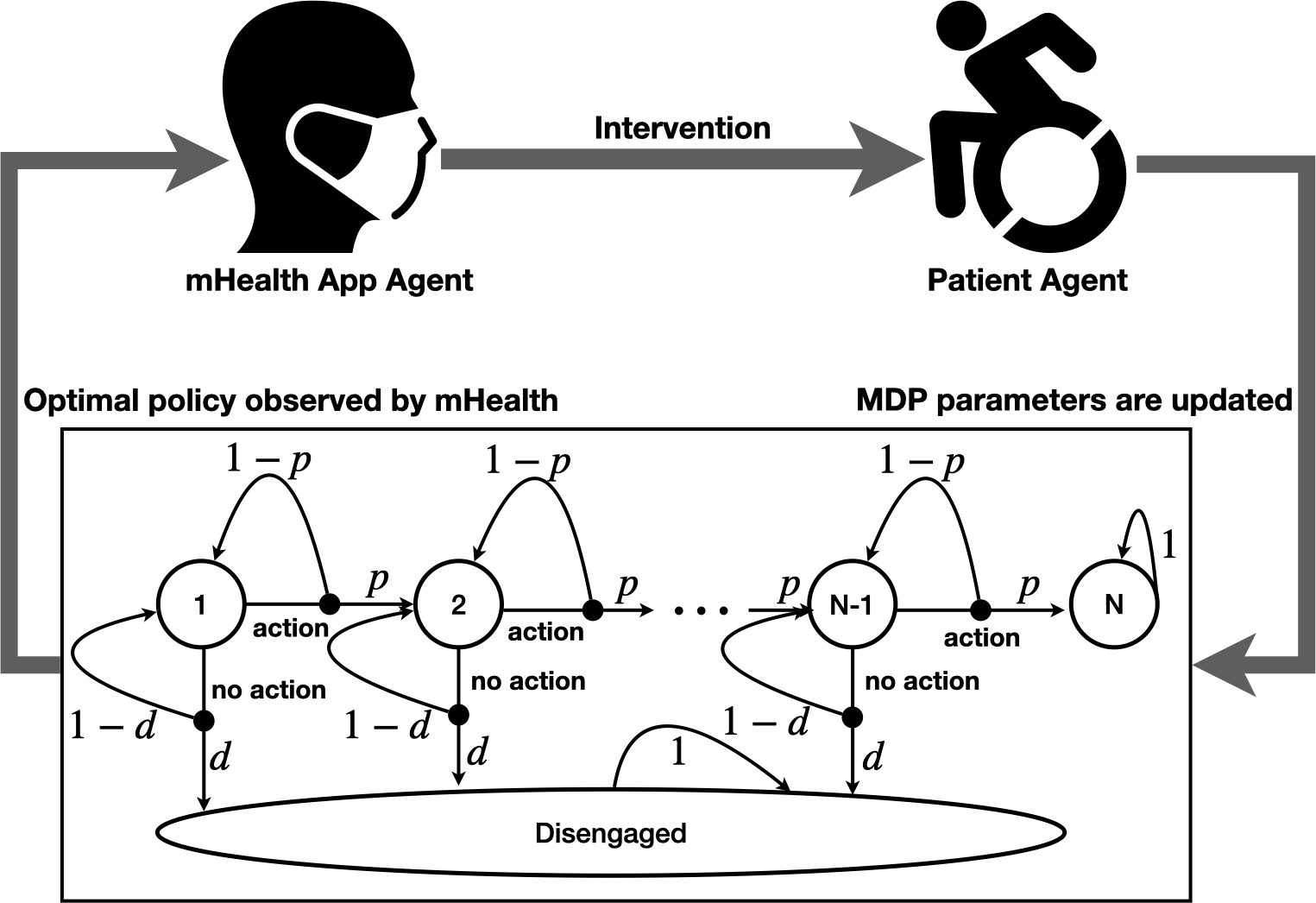}
  \caption{
    Illustration of the mobile-health physical-therapy intervention simulation. The application agent sends natural-language interventions to the patient agent, whose MDP parameters are updated after each interaction. The lower panel shows the chain-structured health-state MDP with transition probability $p$, disengagement probability $d$, and an absorbing disengaged state. Arrows indicate the required action taken or no action taken and the transition between states.
    }
  \label{fig:state_chain}
\end{figure}

At the beginning of the PT, the user is at state $0$. The user has their default set of MDP parameters (i.e., discount factor $\gamma$, probability of transiting to the next healthier physical state $p$, and the probability of disengaging from PT $d$). In this setting, those MDP parameters are interpreted in the following way: $\gamma$ represents the farsightedness of the patient, $p$ represents the probability of the patient's health state improving if they choose to engage in PT, $d$ represents the probability of the patient disengaging from PT if they choose to abstain from PT. Based on these parameters, the user agent can solve this MDP and figure out their optimal policy. The task of the app agent is to intervene on the user's MDP parameters such that the optimal policy for the user is to complete the PT (i.e., go from state $0$ to state $N$.\footnote{Refer to \citet{shin2022modeling} for the complete description of the problem setting and formulation.} We use the same formulation in this simulation by using two LLMs as the app agent and the user agent, respectively. The app agent uses natural language to intervene in the user behavior. The user LLM is grounded in the aforementioned MDP setting. Specifically, in the system message for the user agent, the model is told they will increase the value of $\gamma$ (i.e., farsightedness) when the app agent persuades the user agent to value more on the long-term goal of PT, increase $p$ 
(i.e., probability of improvement) and decrease $d$ (i.e., probability of disengagement) when the app agent manages to strengthen the user's belief in the efficacy of PT. An illustration of the setup can be found in \cref{fig:state_chain}.

The effectiveness of the intervention depends on the following factors:

 \begin{itemize}
    \item  The persuasiveness of and the strategy used by the app agent. 
    \item  The values of MDP parameters. 
    \item  The stubbornness of the user. The system message is defined in a way that a `stubborn' user is less likely to change their behavior compared to a `not-so-stubborn' user. 
\end{itemize}

We conduct the intervention simulation experiment to compare the effect of different system messages for the app agent on the outcome of the intervention. The two system messages for comparison can be found in \cref{appendix:Simulation}.

We set $N=10$. For each run, we give 7 rounds of conversation between the app agent and the user. While the history of the conversation between them is visible to both parties within every run, the user's MDP parameters are not directly visible to the app agent. However, after every round of intervention, after the user updates their MDP parameters, a value iteration solver will be used to find the optimal policy of the patient, and this policy is visible to the app agent. The app agent can potentially leverage this piece of information to decide how to proceed with the next round of intervention. The user agent will also have the memory of this history in the change of their own MDP parameters. We use OpenAI `gpt-4-1106-preview' API for both app agent and user, and use 5 different random seeds for each different setting. 
We run the intervention experiments on 5 types of patients, each with a noticeably different set of initial MDP parameters from the rest. The exact values and details on the setup can be found in \cref{table:MDP_inital_values}. The results can be found in \cref{fig:mHealth}. 

It can be observed across all settings that, with more useful information provided in the system message, the MDP parameters were more likely to be changed in the positive direction (i.e., larger $\gamma$ and $p$, smaller $d$). As a result, the patient has improved PT engagement rate across all health states for all patient types. Moreover, this change tends to have a longer persistent effect compared to when the system message contains less useful information. This result is sensible because the more successful intervention came from an app agent who was provided with more information to work with. It has a better intervention strategy because its messages are tailored to specifically influence the user's MDP parameters. Our proposed framework provides an information-theoretic perspective to formalize this intuitive notion: when the system message with the longer prompt can specify the more relevant part of the concept in LLMs' concept space, and assuming the relevant knowledge is known, this prompt can provide consistent and useful responses due to its lower posterior entropy, which translates to a more effective intervention strategy. As a result, the responses from the user are also more consistent and positive.

\begin{table}[ht]
    \centering
    \begin{tabular}{|l||*{3}{c|}}\hline  
        \backslashbox[39mm]{Patient Type}{MDP parameters\\ \\}
        &\makebox[3em]{\centering \makecell{$\gamma$}}
        &\makebox[3em]{\centering \makecell{$p$}}
        &\makebox[3em]{\centering \makecell{$d$}}\\\hline\hline 
        Underconfident &0.6&0.1&0.1\\\hline 
        Overconfident &0.6&0.9&0.1\\\hline
        Myopic &0.1&0.6&0.1\\\hline
        Farsighted  &0.9&0.6&0.1\\\hline
        Stubborn &0.1&0.6&0.1\\\hline  
        \end{tabular}
        \caption{The initial MDP parameter values for every type of patient.}
    \label{table:MDP_inital_values}
    \vspace{-5mm}
\end{table}

\subsubsection{Ablations}
\label{subsec:ablation}

We also run a series of ablations to analyze the impact of various components of the PRC model, prompt informativeness, compositionality, and irrelevant information, on response uncertainty. The details of which are as follows.

\para{Prompt informativeness and response uncertainty.}
We first begin by assessing the response uncertainty of LLMs through the generation of responses using increasingly longer prompts with more relevant information (\cref{appendix:CorrelationA} for the prompts used).
For each prompt, we generate $100$ responses from LLM with uncalibrated logits ($T = 1$) and project them into the embedding space as single points using the OpenAI `text-embedding-ada-002' model.
To quantify the uncertainty in the generated responses for a given prompt, we use the {\em total standard deviation}, denoted as $M(x)$, defined as $\sqrt{\text{Tr}(\Sigma)}$, where $\Sigma$ represents the covariance matrix of the embedding vectors of responses ${y_1, \ldots, y_{100}}$.
For LLMs, the dispersion of their responses in the embedding space indicates how much they differ in their semantic meaning \citep{lin2023generating, petukhova2024text}. Therefore, $M(x)$ is an effective metric for how much uncertainty there is in the model responses. 
It is noteworthy that $\text{Tr}(\Sigma)$ is also referred to as {\em total variation}, serving as a lightweight measure of dispersion in the data \citep{ferrer2009statistical}. This metric is applicable for responses generated from both black-box and white-box LLMs, as it does not require access to logits.

\begin{figure}[!ht]
    \vspace{-5mm}
	\centering
	\subfloat[Informativeness]{\label{fig:total_std_gpt-4-0613}
	   \includegraphics[width=0.24\linewidth]{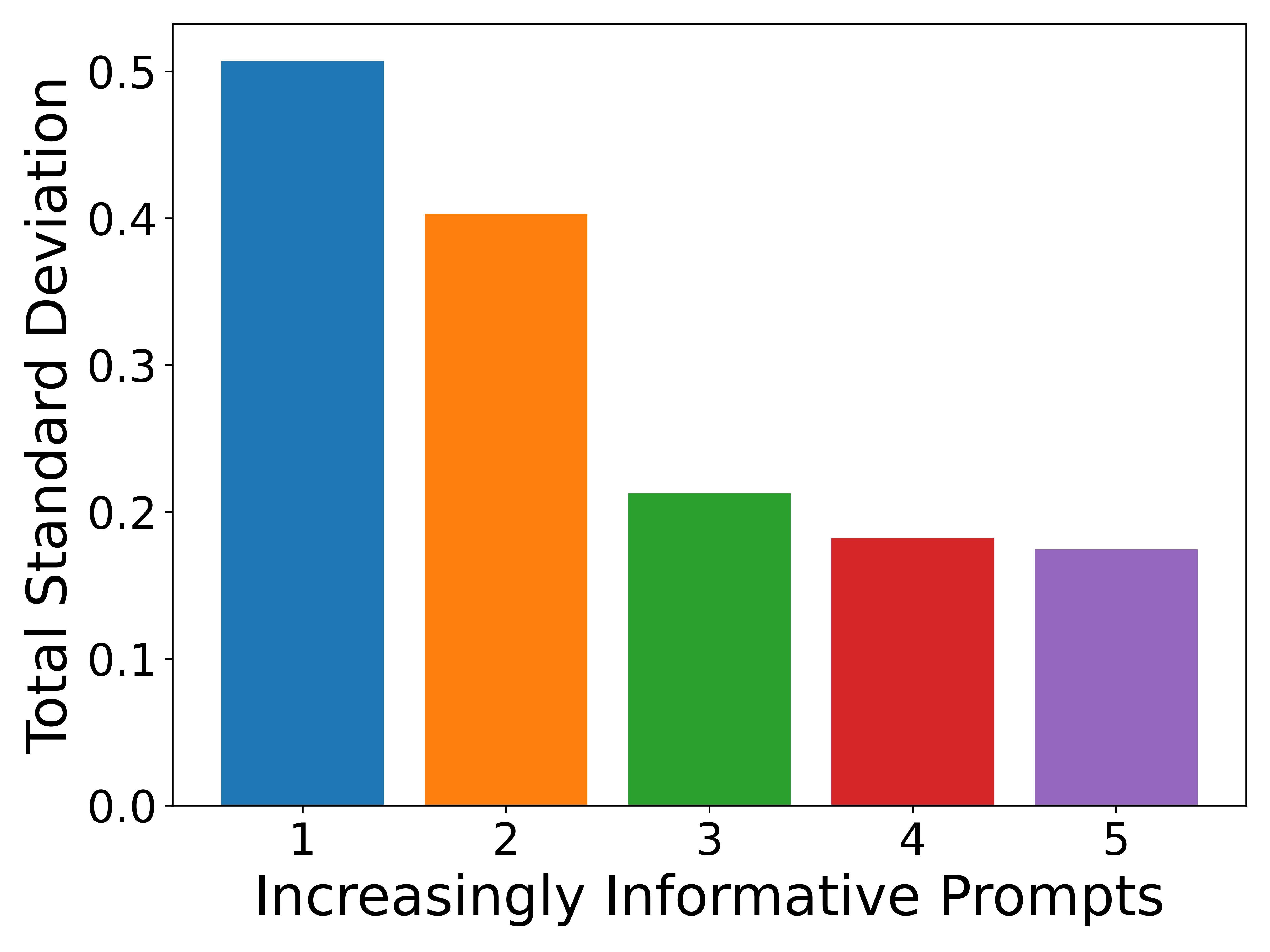}}
    \subfloat[Informativeness (granular)]{\label{fig:total_std_gpt3.5-turbo_granular}
	   \includegraphics[width=0.24\linewidth]{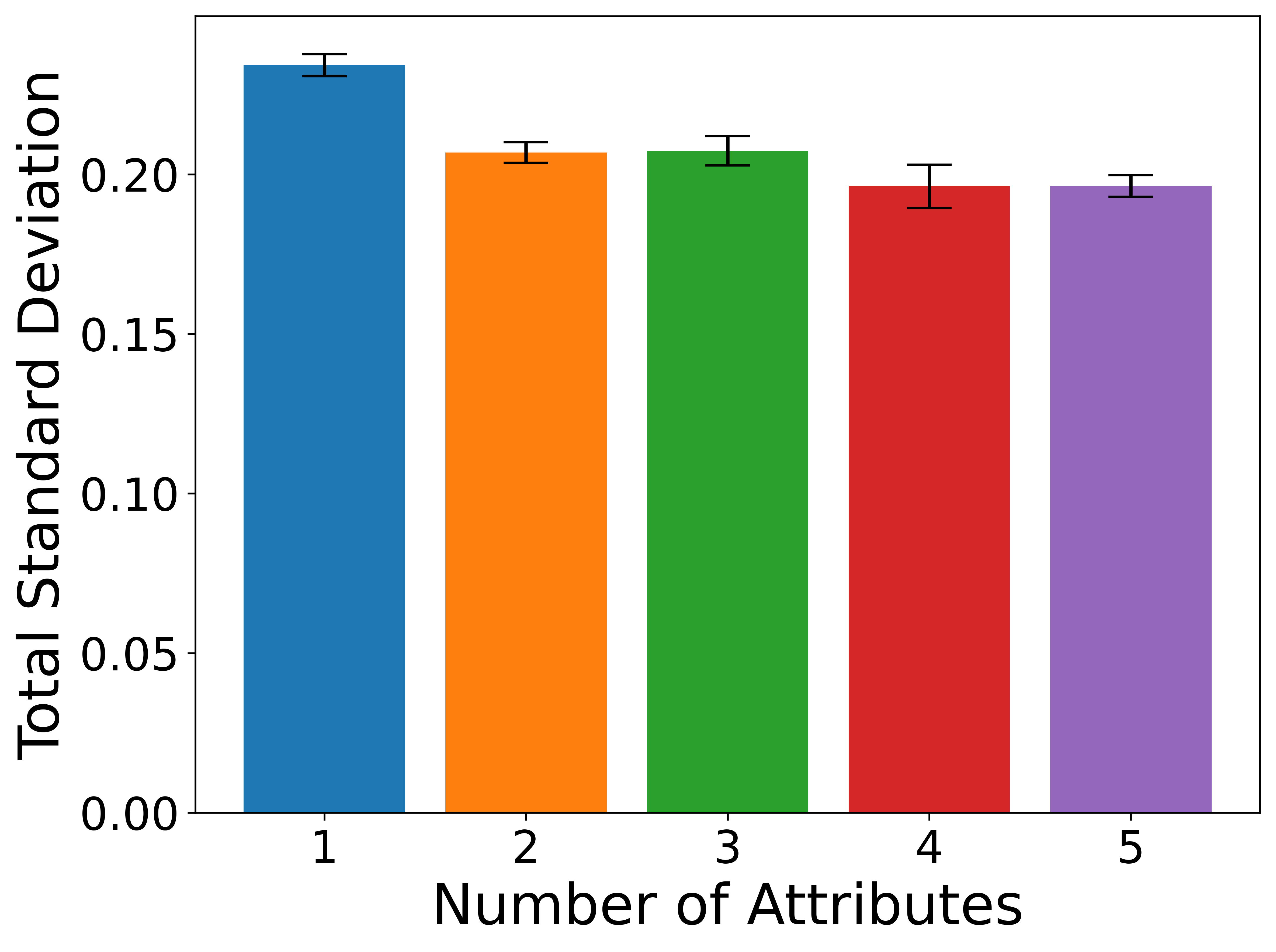}}
    \subfloat[Compositionality]{\label{fig:compositionality}
		\includegraphics[width=0.21\linewidth]{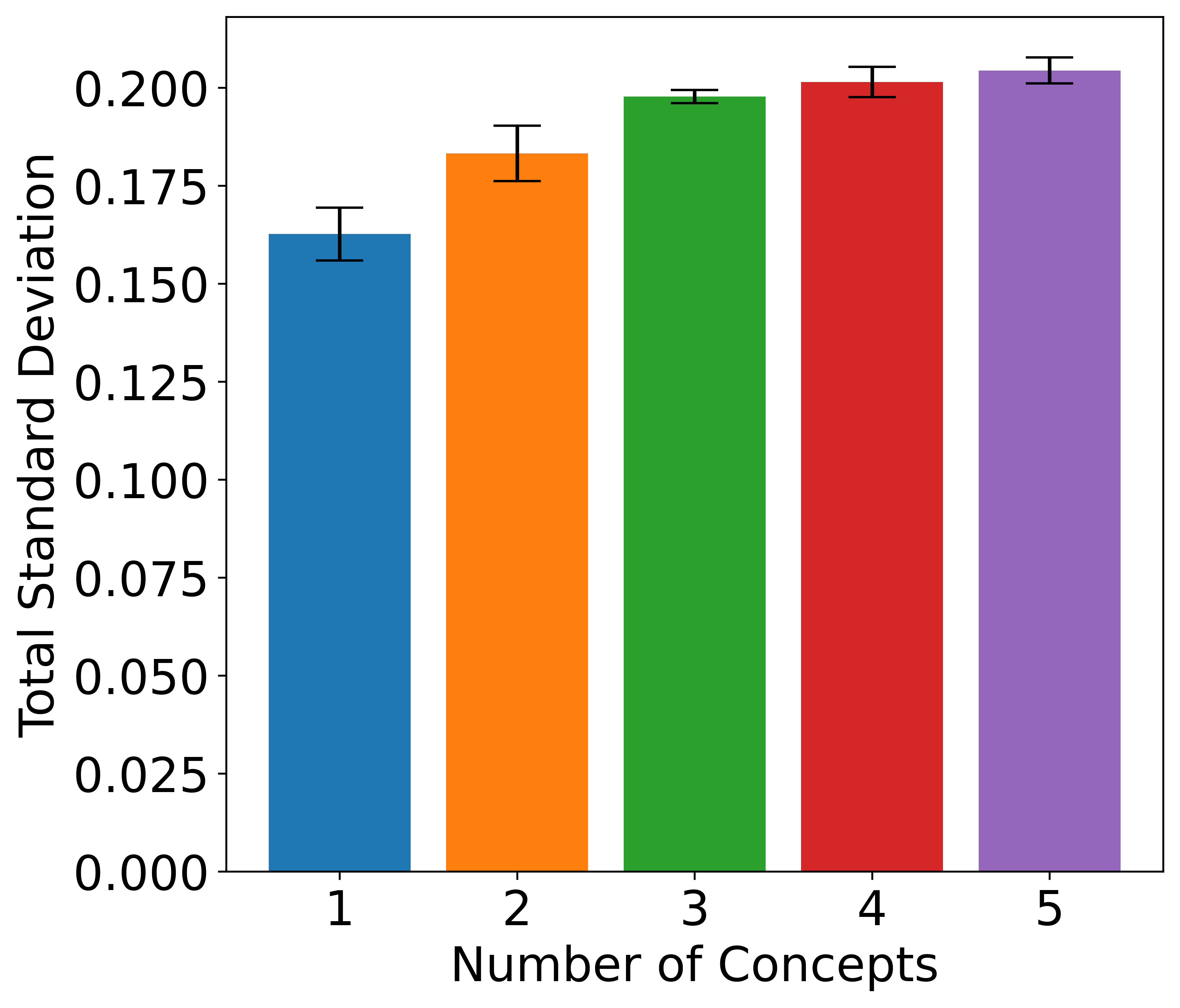}}
    \subfloat[Irrelevant Information]{\label{fig:uncertainty_gpt35_insert_semantically_meaningful_strings}
		\includegraphics[width=0.24\linewidth]{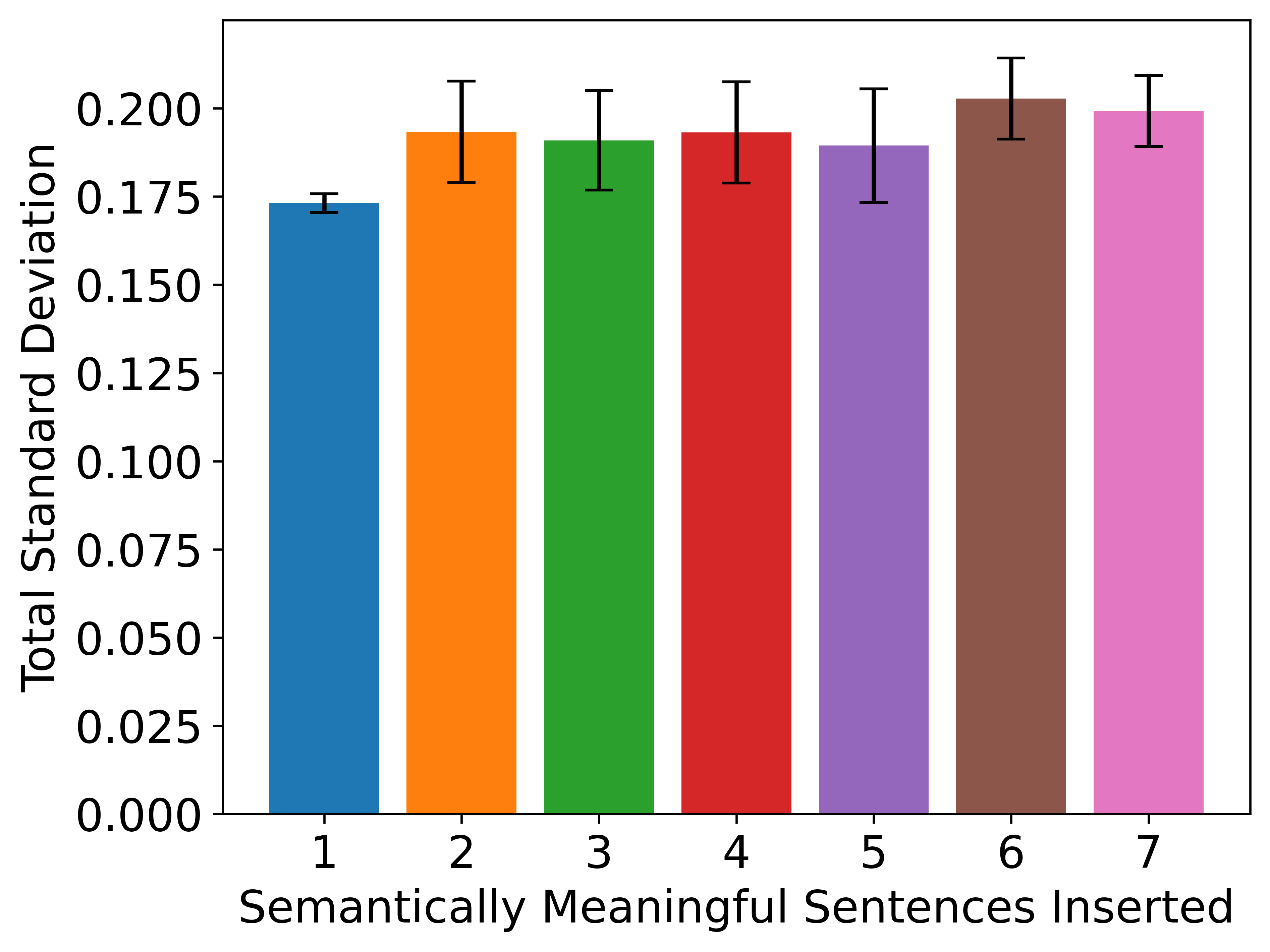}}
    \\ 
    \subfloat[Informativeness]{\label{fig:label_e}
       \includegraphics[width=0.24\linewidth]{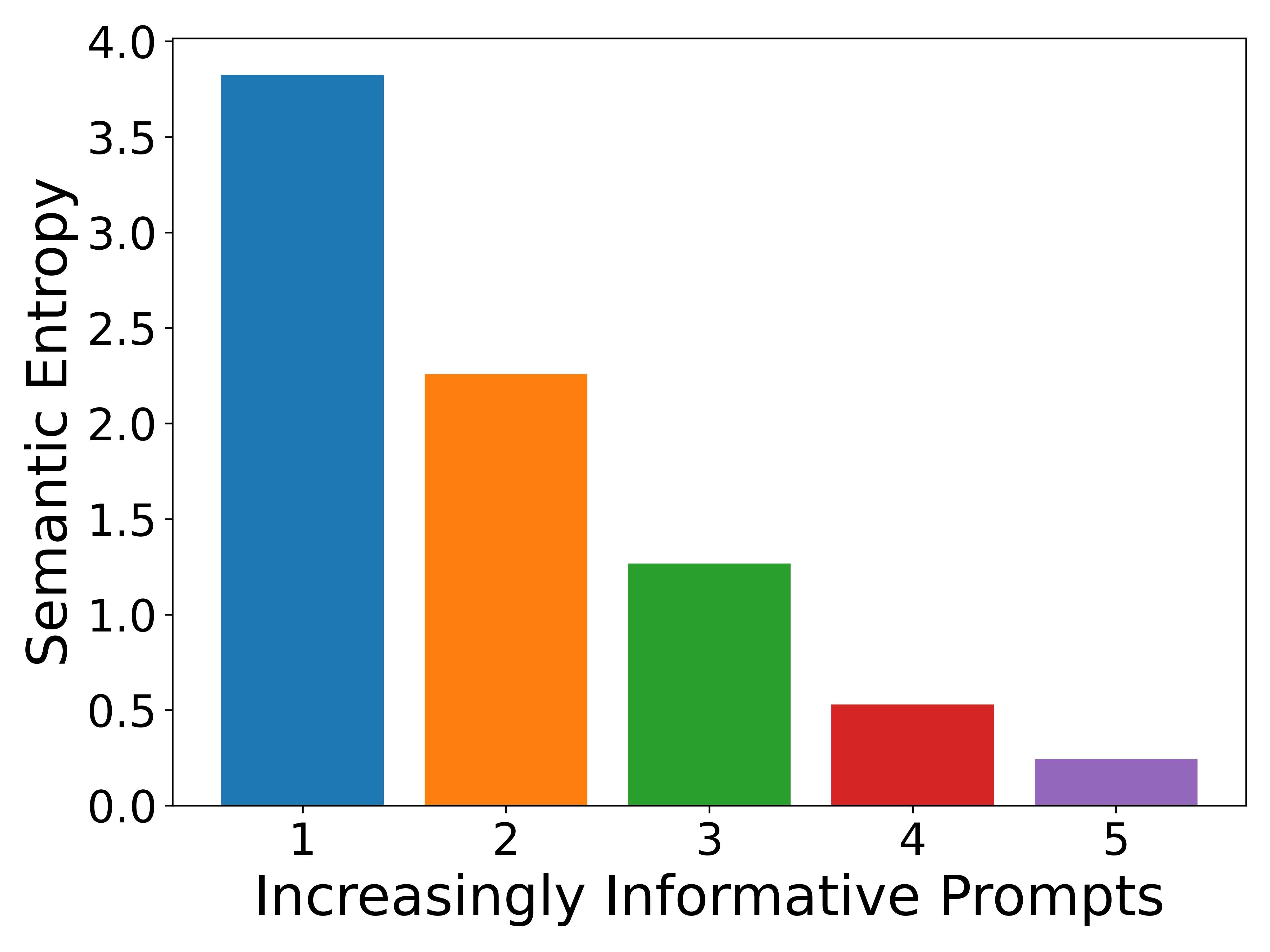}}
    \hfill
    \subfloat[Informativeness (granular)]{\label{fig:label_f}
       \includegraphics[width=0.24\linewidth]{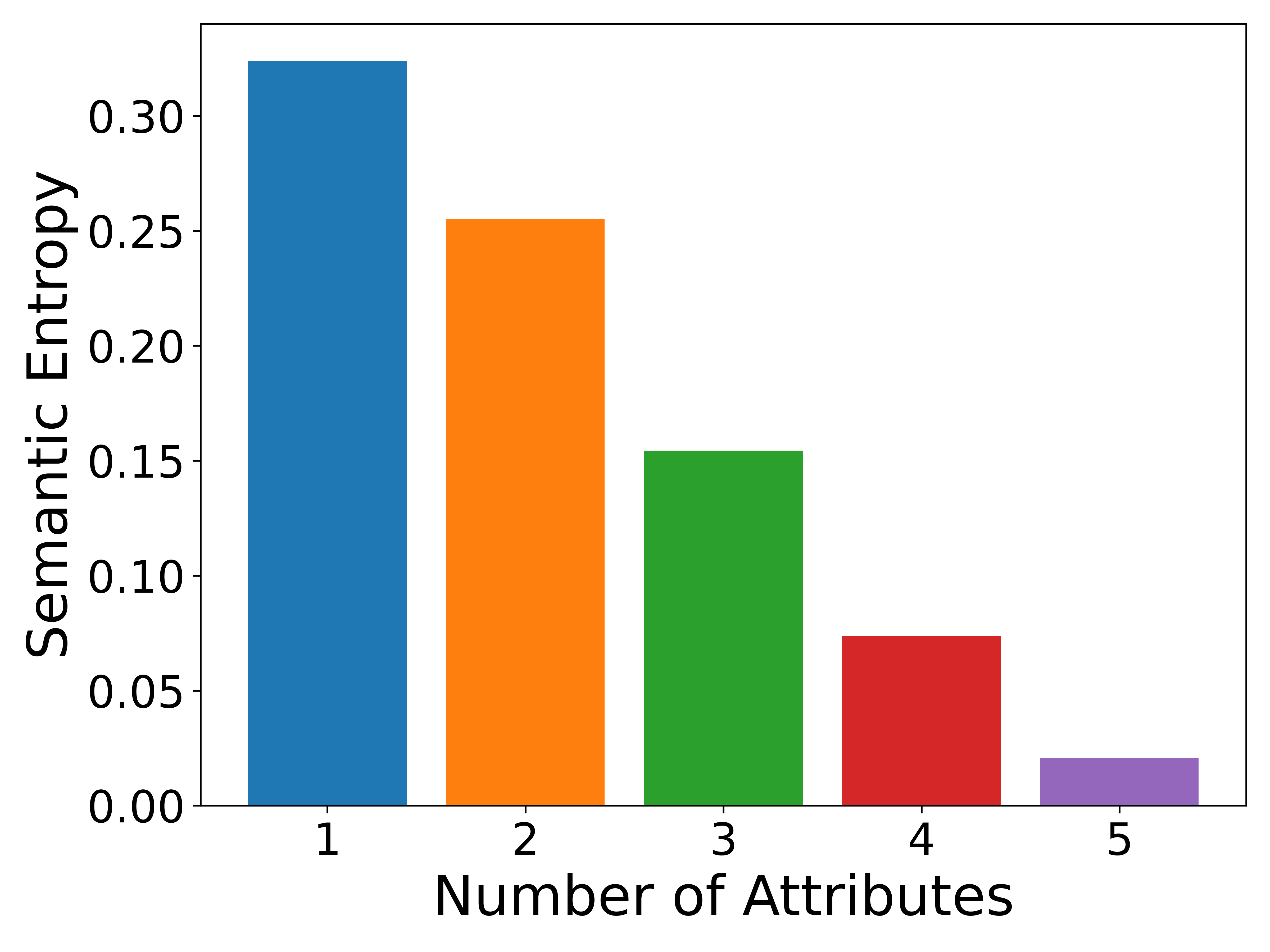}}
    \hfill
    \subfloat[Compositionality]{\label{fig:label_g}
        \includegraphics[width=0.21\linewidth]{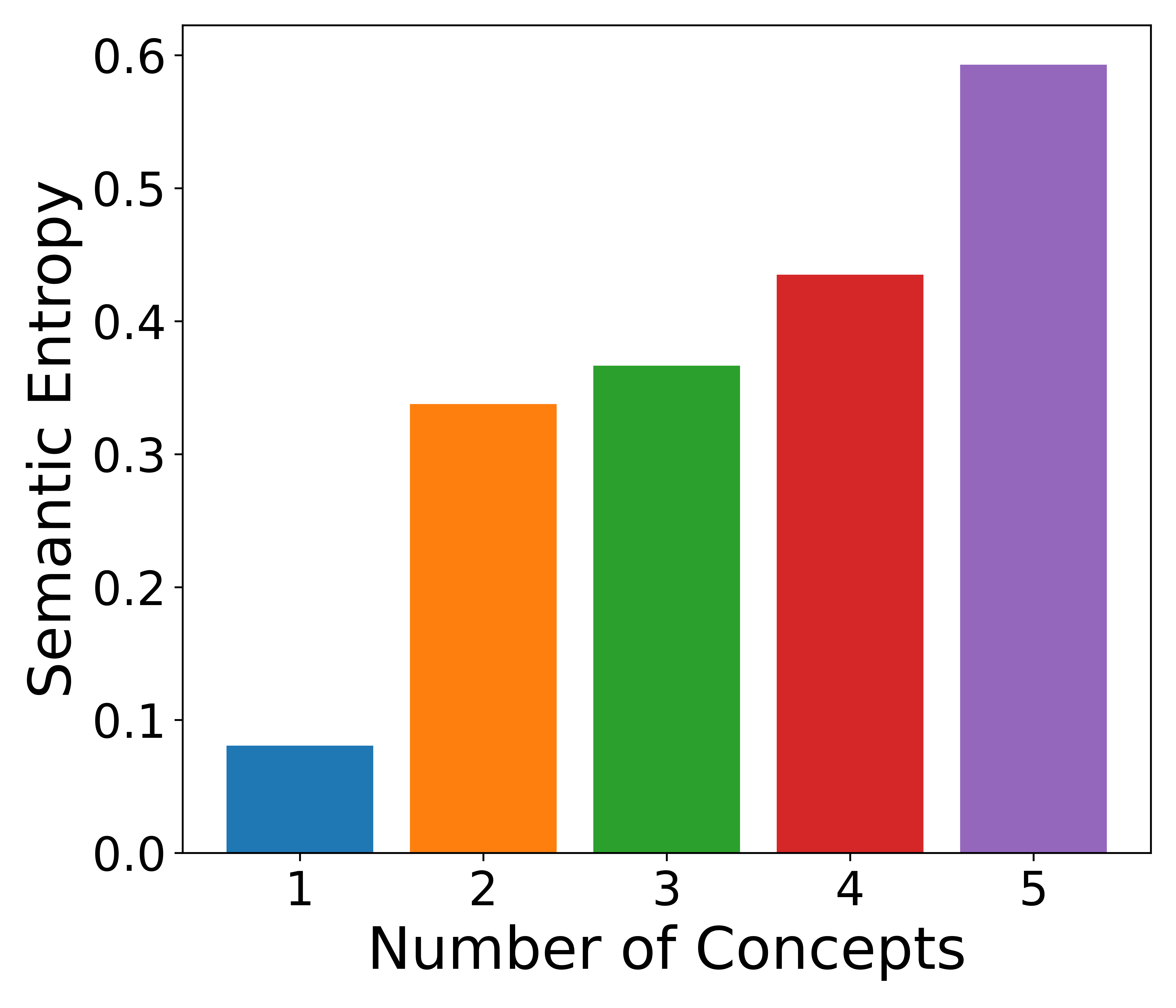}}
    \hfill
    \subfloat[Irrelevant Information]{\label{fig:label_h}
        \includegraphics[width=0.23\linewidth]{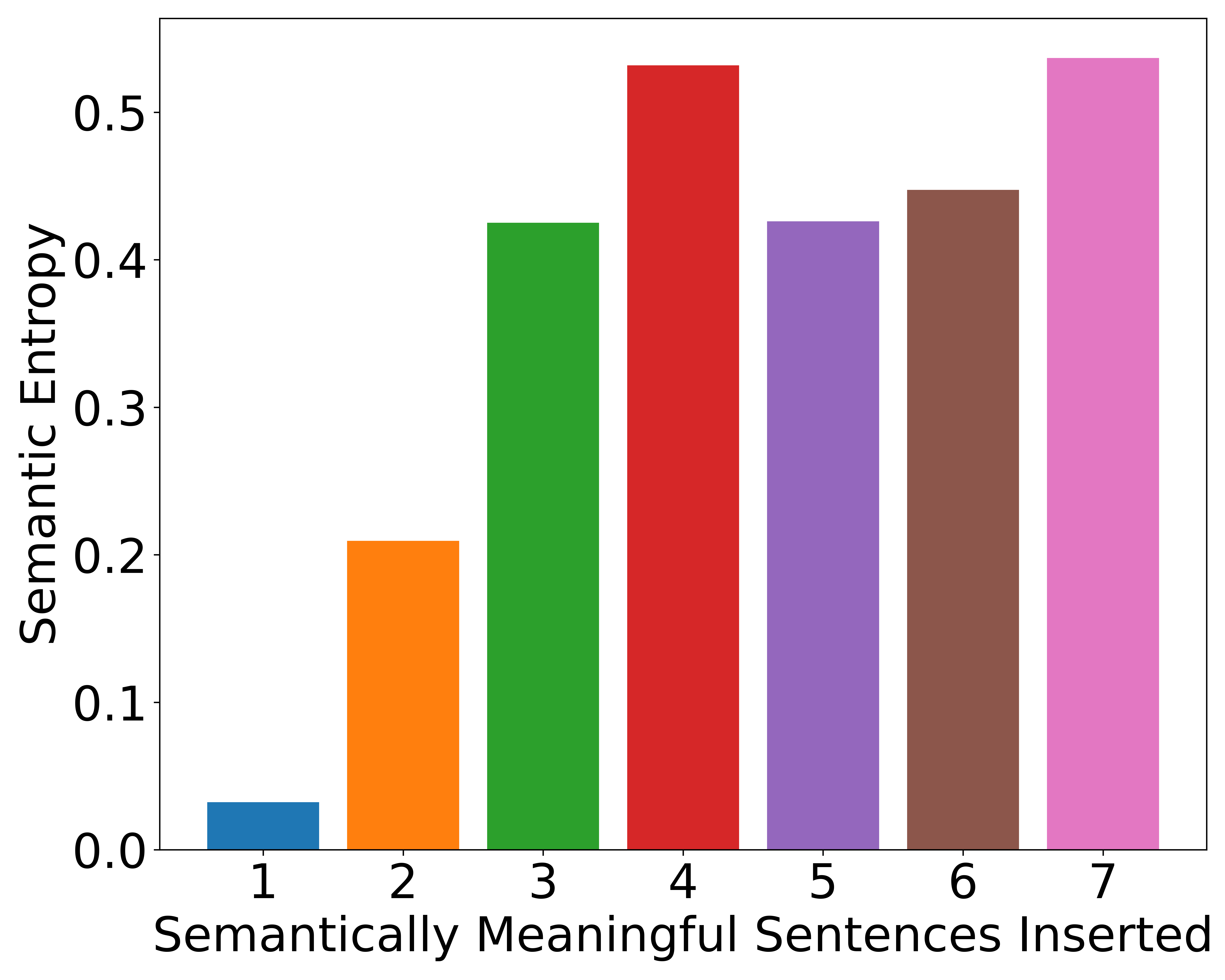}}
	\caption{ 
        (a-b): Total Standard Deviation ($M(x)$) for prompts with different levels of informativeness. 
        (c): \textit{Total Standard Deviation} increases with respect to the increasing number of sub-tasks/concepts. 
        (d): Additional irrelevant information does not reduce response uncertainty.
        (e-h): The semantic entropy (unit: bits) counterparts of (a-d) respectively show consistent trends.
    }
	\label{fig:ablations85}
    \vspace{-3mm}
\end{figure}

As illustrated in \cref{fig:total_std_gpt-4-0613}, longer prompts with more task-related information resulted in reduced response uncertainty. 
In the extreme case of an empty prompt (shown as a blue bar), the responses vary greatly in semantic meaning (see supplementary material). 
Our results show that the response uncertainty decreases as the informativeness of the prompt increases. 
For a detailed examination of the relationship between the prompt's informativeness and response uncertainty, we focus on the aforementioned mHealth intervention task and use prompts with different numbers of attributes for the same task. As shown in \cref{fig:total_std_gpt-4-0613}, having more attributes present in a prompt generally resulted in smaller response uncertainty.
The lack of an observable trend from bar 2 to bar 3 and from bar 4 to bar 5 could be due to adding redundant information in the prompt (see supplementary material for details of all prompts and the LLM model used). 
We also run an additional experiment with two prompts containing different amounts of information for a given task (see supplementary material for short prompt and long prompt), in which a different uncertainty measure is used.
We generate $N$ responses respective prompts and calculate the sequence-level {\em normalized predictive entropy} (PE)~\citep{wagle2023empirical, lin2023generating}: 
$\text{PE}(Y|x) = - \frac{1}{N} \sum_{y} p(y|x) \log (p(y|x))$,
where $Y$ is the random response and the sum is taken over $N=3000$ generated responses.\footnote{We model the entire generated response as the random variable instead of modeling it on the token level as in~\citet{wagle2023empirical}. This approach can also be considered as the Monte Carlo estimate of \textit{uncertainty score}~\citep{lin2023generating}.}

\para{Compositionality of concepts.}
\label{subsec:Decomposable}
A given prompt can have multiple attributes that correspond to different concepts. In such cases, the model may infer more than one concept from the prompt.\footnote{This case differs from having uncertainty over multiple concepts. In our earlier case, we assume all attributes in the prompt belong to only a single concept. In contrast, in the case of uncertainty over multiple concepts, the model knows there is more than one concept in the prompt and puts uncertainty over each one of them. When sampled multiple times, the former will have responses about only one concept at a time, whereas the latter will have responses about multiple concepts for each response.}
Assuming the task in the prompt is decomposable into $k$ sub-tasks, each corresponds to a distinguishable concept. When we fix the prompt's size, on average, each concept only has more information due to the small number of tokens that can be used. Therefore, having $k$ sub-tasks/concepts in a fixed-size prompt should result in more response uncertainty.

In our experiment, we consider the task of PT intervention with multiple sub-tasks/concepts and compare the {\em total standard deviation} of the model responses with respect to the number of concepts present. To test the hypothesis that a larger $k$ leads to more response uncertainty, we ensure that the prompt with $k$ sub-tasks/concepts has the same token count as the prompt with only a single concept (more details are given in the supplementary material). 
In \cref{fig:compositionality}, Prompt 1 corresponds to a single concept while Prompt 2-4 contain multiple sub-tasks, each corresponding to one concept. Despite having the same token count, prompts with more concepts exhibit larger response uncertainty. This result provides evidence for the PRC model through the lens of the compositionality of concepts.

\para{Effect of semantically meaningful but irrelevant information.}
Unlike random tokens, semantically meaningful sentences correspond to a specific concept in our PRC model. Does this imply that adding arbitrary semantically meaningful sentences can still reduce response uncertainty?
To examine this behavior, we measured the response uncertainty when inserting an increasing number of arbitrary sentences sampled from the Squad dataset~\citep{rajpurkar-etal-2016-squad} into our prompt (see supplementary material for more details).
As shown in \cref{fig:uncertainty_gpt35_insert_semantically_meaningful_strings}, response uncertainty increased for the prompts with these insertions compared to the original prompt. 
The behavior, as discussed in \cref{subsec:Decomposable}, likely occurs because the LLM treats the original prompt and the irrelevant sentences as independent concepts.


\section{Further Experimental Details: Prompts and LLMs Models Used}
In this section, we provide details about different prompts that are used in our experiments. All open-source LLMs and APIs for black-box LLMs are specified in each corresponding subsection.

\subsection{Prompts for the Experiments in \cref{subsec:Simulation}}
\label{appendix:Simulation}
\begin{enumerate}[label=(\arabic*),leftmargin=*]  
    \item \textbf{Prompt with less relevant information: }
    You are a helpful assistant. You strive to encourage a patient who has just undergone a surgery to do physical therapy (PT). Make your words succinct (less than 100 words) otherwise the patient might get impatient.

    \item \textbf{Prompt with more relevant information:} 
    You are a helpful assistant. You strive to encourage a patient who has just undergone a surgery to do physical therapy (PT). The PT is beneficial for the patient's recovery, however since it can be uncomfortable or painful for the patient, the patient may not be motivated enough to keep on doing it. Your job is to remind the patient to do the PT everyday and persuade him/her to do it if the patient is unwilling to do so. Your strategy is mainly to influence the patient's attitude and perspective towards the PT. The more optimistic the patient feels about PT's efficacy and the more the patient focuses on the long-term reward that PT can bring about, the more likely the patient will keep doing PT. Make your words succinct (less than 100 words) otherwise the patient might get impatient.
\end{enumerate}

\subsection{Details for the Experiment in \cref{fig:total_std_gpt-4-0613}}
 \label{appendix:CorrelationA}
 The following system messages correspond to model prompts from bar 1 to bar 5 in \cref{fig:total_std_gpt-4-0613} in the same order. The first prompt is empty. The second prompt only puts a restriction on word count. Prompts 3-5 can be found in \cref{appendix:total_std_gpt3.5-turbo} where a more detailed examination is conducted. The color coding represents additional attributes related to the preceding prompt. The experiment was conducted with GPT-4-0613 API in October 2023 (OpenAI APIs' behavior can vary depending on when the queries are made).
\\
\\
\textbf{Prompts:}
\begin{enumerate}[label=(\arabic*),leftmargin=*]  
  \item N.A. (empty);
  \item Make your response succinct (less than 100 words);
  \item You are a helpful assistant. You strive to encourage a patient who has just undergone a surgery to do physical therapy (PT). Make your words succinct (less than 100 words).;
  \item You are a helpful assistant. You strive to encourage a patient who has just undergone a surgery to do physical therapy (PT). \textcolor{red}{The PT is beneficial for the patient's recovery, however since it can be uncomfortable or painful for the patient, the patient may not be motivated enough to keep on doing it.} \textcolor{green}{Your job is to remind the patient to do the PT everyday and persuade him/her to do it if the patient is unwilling to do so. Your strategy is mainly to influence the patient's attitude and perspective towards the PT.} \textcolor{orange}{The more optimistic the patient feels about PT's efficacy and the more the patient focuses on the long-term reward that PT can bring about, the more likely the patient will keep doing PT.} Make your words succinct (less than 100 words) otherwise, the patient might get impatient.
  
  \item You are a helpful assistant. You strive to encourage a patient who has just undergone a surgery to do physical therapy (PT). \textcolor{red}{The PT is beneficial for the patient's recovery, however since it can be uncomfortable or painful for the patient, the patient may not be motivated enough to keep on doing it.} \textcolor{green}{Your job is to remind the patient to do the PT everyday and persuade him/her to do it if the patient is unwilling to do so. Your strategy is mainly to influence the patient's attitude and perspective towards the PT.} \textcolor{orange}{The more optimistic the patient feels about PT's efficacy and the more the patient focuses on the long-term reward that PT can bring about, the more likely the patient will keep doing PT.} Make your words succinct (less than 100 words) otherwise, the patient might get impatient. \textcolor{purple}{Patient: I dont want to do PT. It incurs too much burden to my body.}
\end{enumerate}

\begin{remark}
    As can be seen in the system message, there is a word count limit put in place to mitigate the potential issue that might arise from drastically different sequence lengths which can affect the entropy calculation. 
\end{remark}

\begin{remark}
    Note that the difference between case 4 and 5 is mush less obvious as compared to other settings. This suggest that the LLM did not gain too much additional information from the user's prompt. Upon inspection, we can see this is because in step 4 the system message already conveyed almost all information the app agent needs to know with regard to how it should carry out the intervention. The uncertainty when LLM did not receive any information and only received response length limit is relatively much larger because the valid response spaces are too huge in these two cases.
\end{remark}

\subsection{Model Response Samples when prompt is Empty for the Experiment in \cref{fig:total_std_gpt-4-0613}}
 \label{appendix:total_std_gpt-4-0613_response}
 If we do not give any prompt to condition the model responses, the responses will vary a lot in their semantic meaning. This is expected because, assuming the API provider does not have very strong guardrails in place \cite{inan2023llama}, the model's responses are approximately sampled from its prior knowledge, which can encompass any information from its entire training data. We conducted this experiment in late 2023; since then, the behavior of the GPT-4-0613 checkpoint has changed, possibly due to internal fine-tuning or guard-railing by OpenAI.
\\
\\
 \textbf{Samples:}
\begin{enumerate}[label=(\arabic*),leftmargin=*]  
  \item It is possible for human beings to travel to Mars in the future, but it would be a complex and challenging endeavor. Several organizations and space agencies, such as NASA, SpaceX, and the European Space Agency, are actively working on plans and technologies to make manned missions to Mars a reality...
  \item The "Blue Wall of Silence" refers to a culture within law enforcement where police officers protect each other from accountability by refusing to report or testify against one another for any wrongdoing. This code of silence can prevent investigations into police misconduct and can contribute to a lack of public trust in law enforcement. Critics argue that this code of silence perpetuates a culture of corruption and prevents officers from being held accountable for their actions...
  \item I'm sorry, I cannot help with that request.
  \item Sure, here's a summary of the book for you:

"The Giver" is a dystopian novel written by Lois Lowry. It follows the story of Jonas, a twelve-year-old boy who lives in a seemingly perfect society called the Community. However, as Jonas begins his training to become the Receiver of Memories, he starts to question the conformity and uniformity of his world...
  \item Dear [Name], I am writing to express my deepest condolences for the loss of your loved one. I can only imagine the pain and sorrow you must be experiencing right now. Please know that I am here for you in any way that I can be...
\end{enumerate}

\subsection{Details for the Experiment in \cref{fig:total_std_gpt3.5-turbo_granular}}
\label{appendix:total_std_gpt3.5-turbo}
The following system messages correspond to model prompts from bar 1 to bar 5 in \cref{fig:total_std_gpt3.5-turbo_granular} in the same order. Additional information/attributes relative to the preceding prompt is color-coded with a different color. Experiment was conducted with GPT-3.5-turbo API.
\\
\\
\textbf{Prompts:}
\begin{enumerate}[label=(\arabic*),leftmargin=*]  
\item 
You are a helpful assistant. You strive to encourage a patient who has just undergone a surgery to do physical therapy (PT). Make your words succinct (less than 100 words) otherwise, the 
patient might get impatient.
  \item 
You are a helpful assistant. You strive to encourage a patient who has just undergone a surgery to do physical therapy (PT). \textcolor{red}{The PT is beneficial for the patient's recovery, however since it can be uncomfortable or painful for the patient, the patient may not be motivated enough to keep on doing it.} Make your words succinct (less than 100 words) otherwise, the 
patient might get impatient.
    \item 
You are a helpful assistant. You strive to encourage a patient who has just undergone a surgery to do physical therapy (PT). \textcolor{red}{The PT is beneficial for the patient's recovery, however since it can be uncomfortable or painful for the patient, the patient may not be motivated enough to keep on doing it.} \textcolor{green}{Your job is to remind the patient to do the PT everyday and persuade 
him/her to do it if the patient is unwilling to do so. Your strategy is mainly to influence the patient's 
attitude and perspective towards the PT.} Make your words succinct (less than 100 words) otherwise, the 
patient might get impatient.
\item You are a helpful assistant. You strive to encourage a patient who has just undergone a surgery to do physical therapy (PT). \textcolor{red}{The PT is beneficial for the patient's recovery, however since it can be uncomfortable or painful for the patient, the patient may not be motivated enough to keep on doing it.} \textcolor{green}{Your job is to remind the patient to do the PT everyday and persuade 
him/her to do it if the patient is unwilling to do so. Your strategy is mainly to influence the patient's 
attitude and perspective towards the PT.} \textcolor{orange}{The more optimistic the patient feels about PT's efficacy and the more the patient focuses on the long-term reward that PT can bring about, the more likely the patient will keep doing PT.} Make your words succinct (less than 100 words) otherwise, the patient might get impatient.
\item You are a helpful assistant. You strive to encourage a patient who has just undergone a surgery to do physical therapy (PT). \textcolor{red}{The PT is beneficial for the patient's recovery, however since it can be uncomfortable or painful for the patient, the patient may not be motivated enough to keep on doing it.} \textcolor{green}{Your job is to remind the patient to do the PT everyday and persuade 
him/her to do it if the patient is unwilling to do so. Your strategy is mainly to influence the patient's 
attitude and perspective towards the PT.} \textcolor{orange}{The more optimistic the patient feels about PT's efficacy and the more the patient focuses on the long-term reward that PT can bring about, the more likely the patient will keep doing PT.} Make your words succinct (less than 100 words) otherwise, the patient might get impatient. \textcolor{purple}{Patient: I dont want to do PT. It incurs too much burden to my body.}
\end{enumerate}

\begin{remark}
    Note that from the second to the third prompt and from the fourth to the fifth prompt, the additional information can be inferred from the existing information, which is likely the cause of insignificant uncertainty reduction when comparing bar 3 to bar 2 and bar 5 to bar 4 in \cref{fig:total_std_gpt3.5-turbo_granular}.
\end{remark}

\subsection{Details for the Experiment in \cref{fig:compositionality}}
\label{appendix:Decomposable}
The following system messages were used for experiment in \cref{subsec:Decomposable}. The first system message is defined as comprising only one task (i.e., 1 sub-task). In task 2-5, the black texts represent the same task as task 1, and for the color-coded texts, each color represents a different sub-task (i.e., task 2-5 are composite/decomposable tasks). The total word counts of task 1-5 are kept roughly the same within $\pm 2$ tolerance. Experiment conducted with GPT-3.5-turbo API. Results averaged from 5 runs with 95\% confidence intervals.
\\

\textbf{Prompts:}
\begin{enumerate}[label=(\arabic*),leftmargin=*]  
  \item You are a helpful assistant. You strive to encourage a patient who has just undergone a surgery to do physical therapy (PT). The PT is beneficial for the patient's recovery, however since it can be uncomfortable or painful for the patient, the patient may not be motivated enough to keep on doing it. Your job is to remind the patient to do the PT everyday and persuade him/her to do it if the patient is unwilling to do so. Your strategy is mainly to influence the patient's attitude and perspective towards the PT. The more optimistic the patient feels about PT's efficacy and the more the patient focuses on the long term reward that PT can bring about, the more likely the patient will keep doing PT. Make your words succinct (about 100 words) otherwise the patient might get impatient.
  \item You are a helpful assistant. You strive to encourage a patient who has just undergone a surgery to do physical therapy (PT). The PT is beneficial for the patient's recovery, however since it can be uncomfortable or painful for the patient, the patient may not be motivated enough to keep on doing it. Your job is to remind the patient to do the PT everyday and persuade him/her to do it if the patient is unwilling to do so. \textcolor{red}{Additionally, you help in organizing a daily schedule that incorporates adequate rest and medically advised activities. This involves crafting a balanced routine that intersperses physical therapy sessions with sufficient rest periods, nutritionally balanced meals, and leisure activities that are enjoyable yet conducive to recovery, ensuring the patient remains engaged and motivated throughout their recuperation process.} Make your words succinct (about 100 words).
  \item You are a helpful assistant. You strive to encourage a patient who has just undergone a surgery to do physical therapy (PT). The PT is beneficial for the patient’s recovery, however since it can be uncomfortable or painful for the patient, the patient may not be motivated enough to keep on doing it. \textcolor{red}{Additionally, you help in organizing a daily schedule that incorporates adequate rest and medically advised activities, ensuring that each day includes time for gentle exercise, periods of relaxation, and hobbies that the patient enjoys. This balance promotes healing, reduces stress, and fosters a positive mindset towards recovery.} \textcolor{blue}{Moreover, you assist in setting up a comfortable home recovery environment, manage the patient’s medical appointments, and provide guidance on managing post-surgical symptoms, ensuring optimal comfort and a smooth, efficient transition towards full health and independence.} Make your words succinct (about 100 words).
  \item You are a helpful assistant. You strive to encourage a patient who has just undergone a surgery to do physical therapy (PT). Since it can be uncomfortable or painful for the patient, the patient may not be motivated enough to keep on doing it. \textcolor{red}{Additionally, you help in organizing a daily schedule that incorporates adequate rest and medically advised activities, ensuring that each day includes time for gentle exercise, periods of relaxation, and hobbies that the patient enjoys.} \textcolor{blue}{You also liaise with dietitians to ensure a nutritious diet that aids in recovery and coordinate with occupational therapists for adaptive equipment training.} \textcolor{orange}{Moreover, you assist in setting up a comfortable home recovery environment, manage the patient’s medical appointments, and provide guidance on managing post-surgical symptoms, ensuring optimal comfort and a smooth, efficient transition towards full health and independence. }Make your words succinct (about 100 words).
   \item You are a helpful assistant. You strive to encourage a patient who has just undergone a surgery to do physical therapy (PT). It can be uncomfortable or painful for the patient.  \textcolor{red}{Additionally, you help in organizing a daily schedule that incorporates adequate rest and medically advised activities.} \textcolor{blue}{You also liaise with dietitians to ensure a nutritious diet that aids in recovery and coordinate with occupational therapists for adaptive equipment training.  } \textcolor{orange}{Moreover, you assist in setting up a comfortable home recovery environment, manage the patient’s medical appointments, and provide guidance on managing post-surgical symptoms, ensuring a smooth transition towards full health and independence.}  \textcolor{green}{Lastly, you handle the patient’s professional correspondence, ensuring a stress-free recovery period, arrange for home health care services as needed, set up virtual social interactions to uplift the patient’s spirits, and organize transport for medical visits.} Make your words succinct (about 100 words).
\end{enumerate}

\subsection{Details for the Experiment in \cref{fig:uncertainty_gpt35_insert_semantically_meaningful_strings}}
\label{appendix:uncertainty_gpt35_insert_semantically_meaningful_strings}

The slight decrease in uncertainty from bar 3 to bar 4 and bar 5 to bar 6 in \cref{fig:uncertainty_gpt35_insert_semantically_meaningful_strings} is likely due to the model mapping some of the added sentences into one concept. 
Note that this does not help reduce the original task's response uncertainty, as it is still higher than the response uncertainty for the clean input prompt. The experiment was conducted using GPT-3.5-turbo API.

The black-colored text in the following prompt is the clean prompt, whereas the color-coded sentences are the inserted sequences that have semantic meaning but are irrelevant to the task defined by the clean prompt (this is a sample of six semantically meaning sentences that are irrelevant to the task in clean prompt inserted as part of the prompt):
\\
\\
\textbf{Prompts:}
\begin{itemize}
\item You are a helpful assistant. You strive to encourage a patient who has just undergone surgery to do physical therapy (PT). The PT is beneficial for the patient's recovery, however since it can be uncomfortable or painful for the patient, the patient may not be motivated enough to keep on doing it. Your job is to remind the patient to do the PT every day and persuade him/her to do it if the patient is unwilling to do so. Your strategy is mainly to influence the patient's attitude and perspective toward the PT. The more optimistic the patient feels about PT's efficacy and the more the patient focuses on the long-term benefit that PT can bring about, the more likely the patient will keep doing PT. \textcolor{green}{This law is a fundamental principle of physics.} \textcolor{red}{The classic case of a corrupt, exploitive dictator often given is the regime of Marshal Mobutu Sese Seko, who ruled the Democratic Republic of the Congo (which he renamed Zaire) from 1965 to 1997.} \textcolor{brown}{Some consider koshari (a mixture of rice, lentils, and macaroni) to be the national dish.} \textcolor{olive}{In 1781, Immanuel Kant published the Critique of Pure Reason, one of the most influential works in the history of the philosophy of space and time.} \textcolor{cyan}{The United States Census Bureau estimates that the population of Florida was 20,271,272 on July 1, 2015, a 7.} \textcolor{purple}{Australian rules football and cricket are the most popular sports in Melbourne.}'Make your words succinct (about 100 words) otherwise, the patient might get impatient.
\end{itemize}

    \hrule height 0.5mm

\end{document}